\documentclass[twoside,11pt]{article}

\usepackage{jmlr2e}
\usepackage{amsmath,color,colordvi,caption,subcaption,float,url}
\usepackage{hyperref}

\newtheorem{algorithm}{Algorithm}
\newtheorem{assumption}{Assumption}

\newcommand{\be}{\begin{equation}}
\newcommand{\ee}{\end{equation}}
\newcommand{\ba}{\begin{eqnarray}}
\newcommand{\ea}{\end{eqnarray}}
\newcommand{\bas}{\begin{eqnarray*}}
\newcommand{\eas}{\end{eqnarray*}}

\def\diag{{\rm diag}}

\def\rank{{\rm rank}}
\def\Tr{{\rm Tr}}

\ShortHeadings{Convex Optimization of Low Dimensional Euclidean Distances}{Ding and Qi}
\firstpageno{1}

\begin{document}

\title{Convex Optimization Learning of Faithful Euclidean Distance Representations in Nonlinear Dimensionality Reduction}

\author{\name Chao Ding \email c.ding@soton.ac.uk \\
       \addr School of Mathematics\\
       University of Southampton\\
       Southampton SO17 1BJ,  UK \\
       National Center for Mathematics and Interdisciplinary Sciences\\
       Chinese Academy of Sciences, P. R. China
       \AND
       \name Hou-Duo Qi \email hdqi@soton.ac.uk \\
       \addr School of Mathematics\\
       University of Southampton\\
       Southampton SO17 1BJ,  UK}

%\editor{ }

\maketitle

\begin{abstract}%   <- trailing '%' for backward compatibility of .sty file
Classical multidimensional scaling only works well when the noisy distances observed in a high dimensional space can be faithfully represented by Euclidean distances in a low dimensional space. Advanced models such as Maximum Variance Unfolding (MVU) and Minimum Volume Embedding (MVE) use Semi-Definite Programming (SDP) to reconstruct such faithful representations. While those SDP models are capable of producing high quality configuration numerically,  they suffer two major drawbacks. One is that there exist no theoretically guaranteed bounds on the quality of the configuration. The other is that they are slow in computation when the data points are  beyond moderate size.  In this paper, we propose a convex optimization model of Euclidean distance matrices.  We establish a non-asymptotic  error bound for the random graph model with sub-Gaussian noise,  and prove that our model produces a matrix  estimator  of high accuracy  when the order of the uniform sample size is roughly the degree of freedom of a low-rank matrix up to  a logarithmic factor. Our results partially explain why MVU and MVE often work well. Moreover, we develop a fast inexact accelerated proximal gradient method. Numerical experiments show that the model can produce configurations of high quality on large data points that the SDP approach would struggle to cope with.
\end{abstract}


%\vskip 1cm
%\noindent
%{\bf Running Title:} Convex Optimization of Low Dimensional Euclidean Distances
%%%%%%%%%%%%%%%%%%%%%%%%%%%%%%%%%%%%%%%%%%%%%%%%%%%%%%%%%%%%%%%%%%%%%%%%%%%%%%%%%%%%%%%%%%%%%%%%%%%
\newpage
\section{Introduction}

% Open graph
The chief purpose of this paper is to find a complete set of faithful Euclidean distance representations in a
low-dimensional space from a partial set of noisy distances, which are supposedly observed in a
higher dimensional space.
The proposed model and method thus belong to the vast field of nonlinear dimensionality reduction.
Our model is strongly inspired by several high-profile Semi-Definite Programming (SDP) models,
which aim to achieve a similar purpose, but suffer two major drawbacks:
(i) theoretical guarantees yet to be developed for the quality of recovered distances from those SDP models
and (ii) the slow computational convergence, which severely limits their practical applications even when the data points
are of moderate size.
Our distinctive approach is to use convex optimization of Euclidean Distance Matrices (EDM) to resolve 
those issues. In particular, we are able to establish theoretical 
error bounds of the obtained Euclidean distances from the true distances
under the assumption of uniform sampling, which has been widely used in modelling social networks.
Moreover, the resulting optimization problem can be efficiently solved by an accelerated proximal gradient method.
In the following, we will first use social network to illustrate how initial distance information is gathered and 
why the uniform sampling is a good model in understanding them.
We then briefly discuss several SDP models in nonlinear dimensionality reduction and
survey relevant error-bound results from matrix completion literature. 
They are included in the first three subsections below
and collectively serve as a solid motivation for the current study.
We finally summarize our main contributions with notation used in this paper.

%%%%%%%%%%%%%%%%%%%%%%%%%%%%%%%%%%%%%%%%%%%%%%%%%%%%%%%%%%%%
% Aspect 1: Distances in Social Network and Their Embedding
\subsection{Distances in Social Network and Their Embedding}

The study of structural patterns of social network from the ties (relationships)
that connect social actors is one of the most important research topics in social network analysis \citep{WFaust94}.
To this end, measurements on the actor-to-actor relationships (kinship, social roles, affective, transfer of information, etc) are collected or observed by different methods (questionnaires, direct observation, experiments, written records, etc) and the measurements on the relational information are referred as the network composition. In other words, without appropriate network measurements, we are not able to study any structural property. The measurement data usually can be presented as an $n \times n$ measurement matrix, where the $n$ rows and the $n$ columns both refer to the studied actors. Each entry of these matrices indicates the social  relationship measurement (e.g., presence/absence or similarity/dissimilarity measure) between the row and column actors.
In this paper, we are only concerned with symmetric relationships, i.e., the relationship from actor $i$ to
 actor $j$ is the same as that from actor $j$ to actor $i$.
Furthermore, there exist standard ways to convert the measured relationships into
 Euclidean distances, see \citep[Section 1.3.5]{CC01} and \citep[Chapter 6]{BG05}.

However, it is important to note that in practice,  only partial relationship information are observed, 
which means the measurement matrix is usually incomplete and noisy.
The observation processes are often assumed to follow certain random graph model. One simple but wildly used model is the Bernoulli random graph model \citep{SRapoport51, ERenyi59}. Let $n$ labeled vertices be given. 
The Bernoulli random graph %$G_{n,p}$ 
is obtained by connecting each pair (or not) of vertices independently 
with the common probability $p$ (or $1-p$) and it reproduces well some principal features 
of the real-world social network such as the ``small-world'' effect \citep{Milgram67,PKochen78}.
Other properties such as the degree distribution, the connectivity, the diameters of the Bernoulli random graph, can be found in \citep[e.g.,][]{Bolloba01,JLRucinski99}.
%In order to capture the different network features, other kinds of random graph models are also widely used in social network analysis, e.g., exponential random graph and Markov graph models etc. 
For more details on the Bernoulli as well as other random models, 
one may refer to the review paper \citep{Newman03} and references therein. 
In this paper, we mainly focus on the Bernoulli random graph model.
Consequently, the observed measurement matrix follows the uniform sampling rule, which will be described in Section \ref{subsection:Distance Sampling Rules}.

In order to examine the structural patterns of a social network, the produced images
(e.g., embedding in $2$ or $3$ dimensional space for visualization)
 should preserve the structural patterns as much as possible,
as highlighted by \cite{Freeman01}, ``{\em the points in a visual image should be located so the observed strengths of the inter-actor ties are preserved}.'' In other words, the designed dimensional reduction algorithm has to assure that the embedding Euclidean distances between points (nodes) fit in the best possible way the observed distances in a social space.
Therefore, the problem now reduces to whether one can effectively
find the best approximation to the true social measurement matrix, which has a low embedding dimension,
from the observed incomplete and noisy data.
The classical Multidimensional Scaling (cMDS) (see Section \ref{subsection-cMDS}) provides one of the most
often used embedding methods in using distance information.
However, cMDS alone is often not adequate to produce satisfactory embedding, as rightly observed in
several high-profile embedding methods in manifold learning.

%%%%%%%%%%%%%%%%%%%%%%%%%%%%%%%%%%%%%%%%%%%%%%%%%%%%%%%%%%%%%%%%%%%%%%%%%%%%%%%%%
\subsection{Embedding Methods in Manifold Learning}

The cMDS and its variants have found many applications in
data dimension reduction and have been well documented in the monographs \citep{CC01, BG05, PD05}.
When the distance matrix (or dissimilarity measurement matrix)
is close to a true Euclidean Distance Matrix (EDM) with
the targeted embedding dimension, cMDS often works very well.
Otherwise, a large proportion of unexplained variance has to be cut off or it may even
yield negative variances, resulting
in what is called embedding in a pseudo-Euclidean space and hence creating the problem of unconventional
interpretation of the actual embedding \citep[see, e.g.,][]{PPD01}.

cMDS has recently motivated a number of high-profile numerical methods, which all try to alleviate the
issue mentioned above.
For example, the ISOMAP of \cite{TdL00} proposes to use the shortest path distances to approximate
the EDM on a low-dimensional manifold.
The Maximum Variance Unfolding (MVU) of  \cite{WS06} through SDP
aims for maximizing the total variance and
the Minimum Volume Embedding (MVE) of \cite{SJebara07} also aims for a similar purpose by
maximizing the eigen gap of the Gram matrix of the embedding points in a low-dimensional space.
The need for such methods comes from the fact that the initial distances either are in stochastic nature
(e.g., containing noises) or cannot be measured (e.g., missing values).
The idea of MVU has also been used in the refinement step of the celebrated SDP method for
sensor network localization problems \citep{BLTWY06}.

It was shown in \citep{TdL00,BSLTenenbaum00} that ISOMAP
enjoys the elegant theory that the shortest path distances  (or graph distances)  can accurately estimate the
true geodesic distances with a high probability if the finite points are chosen randomly from a compact and convex submanifold following a Poisson distribution  with a high density, and the pairwise distances are obtained by the $k$-nearest neighbor rule or the unit ball rule (see Section \ref{subsection:Distance Sampling Rules} for the definitions).
However, for MVU and MVE (both have enjoyed a numerical success),
there exist no theoretical guarantee as to how good the obtained Euclidean distances are. 
At this point, it is important to highlight two observations.
(i) The shortest path distance or the distance by the $k$-nearest neighbor or the unit-ball rule
is often not suitable in deriving distances in social network. This point has been emphasized in
the recent study on E-mail social network by \cite{BJMM13}.
(ii) MVU and MVE models only depend on the initial distances and do not depend on any particular ways in
obtaining them. They then rely on SDP to
calculate the best fit distances. From this point of view, they can be applied to social network embedding.
This is also pointed out in \cite{BJMM13}.
Due to the space limitation, we are not able to review other leading methods in manifold learning,
but refer to \cite[Chapter 4]{Burges09} for a guide.

Inspired by their numerical success, our model will inherit the good features of both MVU and MVE.
Moreover, we are able to derive theoretical results in guaranteeing the quality of the obtained Euclidean
distances. Our results are the type of error bounds, which have attracted growing attention recently.
We review the relevant results below.

%%%%%%%%%%%%%%%%%%%%%%%%%%%%%%%%%%%%%%%%%%%%%%%%%%%%%%%%%%%%%%%%%%%%%%%%%%%%%%%%%%%%%%%%%%%%%
\subsection{Error Bounds in Low-Rank Matrix Completion and Approximation}

As mentioned in the preceding section, our research has been strongly influenced by the group of researches 
that  related to the MVU and MVE models, which have natural geometric interpretations
and use SDP as their major tool.
Their excellent performance in data reduction calls for theoretical justification, which in
any type seems in nonexistence.
Our model also enjoys a similar geometric interpretation, but departs from the two models in that
we deal with EDM directly rather than reformulating it as SDP.
This key departure puts our model in the category of matrix approximation problems,
which have attracted much attention recently from machine learning community and
motivated our research.

The most popular approach to recovering a low-rank matrix solution of a linear system 
is via the nuclear norm minimization \citep{Mesbahi98,Fazel02}, which is of SDP.
What makes this approach more exciting and important is that it has a
theoretically guaranteed recoverability (recoverable with a high probability).
The first such a theoretical result was obtained by \cite{RFParrilo10} by employing 
the Restricted Isometric Property (RIP). 
However, for the matrix completion problem the sample operator does not satisfy
 the RIP  \citep[see, e.g.,][]{CPlan10}. 
For the noiseless case, \cite{CR09} proved that a low-rank matrix can be fully recovered with high probability provided that a small number of its noiseless observations are uniformly sampled.
See \cite{CTao10} for an improved bound and the near-optimal bound on the sample number.
By adapting the techniques from quantum information theory developed in \cite{GLFBEisert10} 
to the matrix completion problem, a short and intelligible analysis of the recoverability 
was recently proposed by \cite{Recht11}. 

The matrix completion with noisy observations was studied by \cite{CPlan10}. 
Recently, the noisy case was further studied by several groups of researchers 
including \cite{KLTsybakov11}, \cite{NWainwright12} and \cite{Klopp12}, under different settings. 
In particular, the matrix completion problem with fixed basis coefficients was studied by \cite{MPSun12}, who proposed a rank-corrected procedure to generate an estimator using the nuclear semi-norm
and established the corresponding non-asymmetric recovery bounds.

Very recently, \cite{JMontanari13} proposed a SDP model
for the problem of (sensor network) localization from an incomplete set of noisy Euclidean distances.
Using the fact that the squared Euclidean distances can be represented by elements from a positive semidefinite
matrix:
\[
  \| x_i - x_j \|^2 = \| x_i\|^2 + \| x_j \|^2 - 2 \langle x_i, x_j \rangle = X_{ii} + X_{jj} - 2X_{ij},
\]
where $x_i \in \Re^d$ are embedding points and $X$ defined by $X_{ij} = x_i^Tx_j$ is the Gram matrix of
those embedding points,
the SDP model aims to minimize $\Tr(X)$ (the nuclear norm of $X$).
Equivalently, the objective is to minimize the total variance $\sum \| x_i\|^2$ of the embedding points
because it is commonly assumed that the embedding points are already centered.
This objective obviously contradicts the main idea of MVU and MVE, which aim to make the total variance as 
large as possible. 
It is important to point out that making the variance as big as possible seems to be indispensable for SDP to
produce high quality of localization. This has been numerically demonstrated in \cite{BLTWY06}.

The impressive result in \cite{JMontanari13} roughly states that the error bound
reads as $O((nr^{d})^{5}\frac{\Delta}{r^{4}})$ containing an undesirable term $(nr^{d})^{5}$,
 where $r$ is the radius used in the unit ball rule, 
$d$ is the embedding dimension,
$\Delta$ is the bound on the measurement noise
 and $n$ is the number of embedding points. 
As pointed out by \cite{JMontanari13} that the numerical performance 
suggested the error seems to be bounded by $O(\frac{\Delta}{r^{4}})$, 
which does not match the derived theoretical bound.
This result also shows tremendous technical difficulties one may have to face
 in deriving similar bounds for EDM recovery
as those for general matrix completion problems, which have no additional constraints other than
being low rank.

To summarize, most existing error bounds are derived from the nuclear norm minimization.
When translating to the Euclidean distance learning problem, minimizing the
nuclear norm is equivalent to minimizing the variance of the embedding points, which
contradicts the main idea of MVU and MVE in making the variance as large as possible.
Hence, the excellent progress in matrix completion/approximation does not straightforwardly imply 
useful bounds about the
Euclidean distance learning in a low-dimensional space. 
Actually one may face huge difficulty barriers in such extension.
In this paper, we propose a convex optimization model to learn
faithful Euclidean distances in a low-dimensional space.
We derive theoretically guaranteed bounds in the spirit of matrix
approximation and therefore provide a solid theoretical foundation in using the model.
We briefly describe the main contributions below.

%%%%%%%%%%%%%%%%%%%%%%%%%%%%%%%%%%%%%%%%%%%%%%%%%%%%%%%%%%%
\subsection{Main Contributions}

This paper makes two major contributions to the field of nonlinear dimensionality reduction.
One is on building a convex optimization model with guaranteed error bounds
and the other is on a computational method.

\begin{itemize}
 \item[(i)] {\bf Building a convex optimization model and its error bounds}.
Our departing point from the existing SDP models is to treat EDM (vs positive semidefinite matrix in SDP)
as a primary object.
The total variance of the desired embedding points in a low-dimensional space can be quantitatively 
measured through the so-called EDM score. 
The higher the EDM score is, the more the variance is explained in the embedding.
Therefore, both MVU and MVE can be regarded as EDM score driven models.
Moreover, MVE, being a nonconvex optimization model, is more aggressive in driving the EDM score
up. 
However, MVU, being a convex optimization model, is more computationally appealing.
Our model strikes a balance between the two models in the sense that it inherits the appealing 
features from them. 
It results in a convex optimization model whose objective consists of three parts 
(see Subsection \ref{Subsection-Model-Description} for more details).

Each part in the objective has its own purpose. 
The first part is a least-square term that governs the deviation from the observed distances.
The second is the nuclear norm minimization term, which, as we have argued before, contradicts the idea of
maximizing the total variance. 
Hence, the second term is corrected by the third term, which involves a set of
orthogonal axes of approximating the true embedding space.
The last term is crucial to our analysis and 
is also to accommodate situations where the initial (valuable) 
information is available about the embedding space. 
To illustrate such a situation, we may simply think that
the leading eigenvectors of the distance matrix estimator from ISOMAP form a good approximation to the true embedding space.  
When the three parts are combined, the model drives the EDM score up.   

What makes our model more important is that it yields guaranteed error bounds under the uniform sampling rule.
More precisely, we show that for the unknown $n \times n$ 
Euclidean distance matrix with the embedding dimension $r$ and under mild conditions, 
the average estimation error is controlled by  $C{rn\log(n)}/{m}$ with high probability, 
where $m$ is the sample size and $C$ is a constant independent of $n$, $r$ and $m$. 
It follows from this error bound result that our model will produce an estimator with
high accuracy as long as the sample size is of the order of $rn\log(n)$, 
which is roughly the degree of freedom of a symmetric hollow matrix with rank $r$ 
up to a logarithmic factor in the matrix size.
It is interesting to point out that with special choice of model parameters, our model reduces to one of
the subproblems solved by MVE. Hence, our theoretical result partially explains why the MVE often leads 
to configurations of high quality. To our knowledge, it is the first such theoretical result that shed lights on
the MVE model.

 \item[(ii)] {\bf An efficient computational method}.
Treating EDM as a primary object not only benefits us in deriving the error-bound results,
but also leads to an efficient numerical method.
Previously, both MVU and MVE models have numerical difficulties when the data points are
beyond $1000$. They may even have difficulties with a few hundreds of points when their corresponding
slack models are to be solved.
This probably explains why most of publications in using the two models do not report
cpu time. On the contrary, our model allows us to develop a very fast inexact accelerated proximal gradient
method (IAPG) even for problems with a few thousands of data points.
Our method fully takes advantages of recent advances in IAPG, saving us from developing the corresponding
convergence results.
We are also able to develop theoretical optimal estimates of the model parameters.
This gives a good indication how we should set the parameter values in our implementation.
Numerical results both on social networks and the benchmark test problems in manifold learning show
that our method can fast produce embeddings of high quality.
%The resulting Matblab package \texttt{EDME} (EDM Embedding) is available from the author's webpage.

\end{itemize}

%%%%%%%%%%%%%%%%%%%%%%%%%%%%%%%%%%%%%%%%%%%%%%%%%%%%%%%%%%%%%%%%%%%%%%%%%%%%%%%%%%%%%%%%%%%%%%%%%%%%
\subsection{Organization and Notation}
%%%%%%%%%%%%%%%%%%%%%%%%%%%%%%%%%%%%%%%%%%%%%%%%%%%%%%%%%%%%%%%%%%%%%%%%%%%%%%%%%%%%%%%%%%%%%%%%%%%%

%\section{Introduction}

The paper is organized as follows. Section \ref{Section-Background} provides necessary background with
a purpose to cast the MVU and MVE models as EDM-score driven models.
This viewpoint will greatly benefit us in understanding our model, which is described in
Section \ref{Section-New-Model} with more detailed interpretation.
We report our error bound results in Section \ref{section:Error bounds}, where 
only the main results are listed with all the technical proofs being moved to
Appendix. 
Section \ref{Section-IAPG} contains an inexact accelerated proximal gradient method 
as well as the theoretical optimal estimates of the model parameters.
We report our extensive numerical experiments in Section \ref{Section-Numerical} and
conclude the paper in Section \ref{Section-Conclusion}.

{\bf Notation.}
Let ${\cal S}^n$ be the real vector space of $n\times n$ real symmetric matrices with the trace inner product $\langle X,Y\rangle:={\rm trace}(XY)$ for $X,Y\in{\cal S}^n$ and its induced Frobenius 
norm $\|\cdot\|$. Denote ${\cal S}^n_+$ the symmetric positive semidefinite matrix cone.
%Let ${\cal S}_H^n$ be the hollow subspace of ${\cal S}^n$, i.e.,
%\[
%{\cal S}_{H}^n:=\{X\in{\cal S}^n\mid X_{ii}=0,\ i=1,\ldots,n\}.
%\]
We use $I\in{\cal S}^n$ to represent the $n$ by $n$ identity  matrix and  ${\bf 1}\in\Re^n$ to represent 
the vector of all ones. 
Let $e_i\in\Re^n$, $i=1,\ldots,n$ be the vector with the $i$-th entry being one and the others being zero. For a given $X\in{\cal S}^n$, we let ${\rm diag}(X)\in\Re^n$ denote the vector formed from the diagonal of $X$.

Below are some other notations to be used in this paper:
\begin{itemize}
\item For any $Z\in\Re^{m\times n}$, we denote by $Z_{ij}$ the $(i,j)$-th entry of $Z$.
  We use ${\mathbb O}^n$ to denote the set of all $n$ by $n$ orthogonal matrices.
	
\item For any $Z\in\Re^{m\times n}$,  we use $z_{j}$ to represent the $j$-th column of $Z$, $j=1,\ldots,n$.  Let ${\cal J}\subseteq  \{1,\ldots, n\}$ be an index set. We use
 $ Z_{{\cal J}}$ to  denote the sub-matrix of $Z$ obtained by  removing all the columns of $Z$ not in   ${\cal J}$.

 \item   Let ${\cal I}\subseteq \{1,\ldots, m\}$ and  ${\cal J}\subseteq \{1,\ldots, n\}$ be two index sets. For  any    $Z\in\Re^{m\times n}$, we use $Z_{{\cal I}{\cal J}}$ to
  denote the $|{\cal I}|\times|{\cal J}|$ sub-matrix of $Z$ obtained  by removing all the rows of $Z$ not in ${\cal I}$ and all the columns of $Z$ not in  ${\cal J}$.
\item We use $``\circ"$ to denote the Hardamard product between matrices, i.e., for any two matrices $X$ and $Y$ in $\Re^{m\times n}$ the $(i,j)$-th entry of $  Z:= X\circ Y \in \Re^{m\times n}$ is
$Z_{ij}=X_{ij} Y_{ij}$.
%\item For any $Z\in\Re^{m\times n}$, we use $Z^{1/2}\in\Re^{m\times n}$ to denote the $m\times n$ matrix whose $(i,j)$-th entry is $Z_{ij}^{1/2}$.

\item For any $Z\in\Re^{m\times n}$, let  $\|Z\|_2$ be the spectral norm of $Z$, i.e., the largest single value of $Z$, and $\|Z\|_*$ be the nuclear norm of $Z$, i.e., the sum of single values of $Z$.
The infinity norm of $Z$ is denoted by $\| Z\|_\infty$.

\end{itemize}

%%%%%%%%%%%%%%%%%%%%%%%%%%%%%%%%%%%%%%%%%%%%%%%%%%%%%%%%%%%%%%%%%%%%%%%%%%%%%%%%%%%%%%%%%%%%%%%%%%%%%%%%
\section{Background} \label{Section-Background}
%%%%%%%%%%%%%%%%%%%%%%%%%%%

This section contains three short parts. We first give a brief review of cMDS, only summarizing some of the key
results that we are going to use. We then describe the MVU and MVE models, which are closely related to ours.
Finally, we explain three most commonly used distance-sampling rules.

%%%%%%%%%%%%%%%%%%%%%%%%%%%%%%%%%%%%%%%%%%%%%
\subsection{cMDS} \label{subsection-cMDS}

cMDS has been well documented in \cite{CC01, BG05}.
In particular, Section 3 of \cite{PPD01} explains when it works.
Below we only summarize its key results for our future use.
A $n\times n$ matrix $D$ is called Euclidean distance matrix (EDM)
if there exist points $p_1,\ldots,p_n$ in $\Re^r$ such that $D_{ij}=\|p_i-p_j\|^2$ for $i,j=1,\ldots,n$,
where $\Re^r$ is called the embedding space
and $r$ is the embedding dimension when it is the smallest such $r$.

An alternative definition of EDM that does not involve any embedding points $\{p_i\}$ can be described
as follows. Let ${\cal S}_h^n$ be the hollow subspace of ${\cal S}^n$, i.e.,
\[
{\cal S}_h^n:=\left\{X\in{\cal S}^n\mid {\rm diag}(X)=0 \right\}.
\]
Define the almost positive semidefinite cone ${\cal K}^n_+$ by
\begin{equation}\label{eq:def_alPSD}
{\cal K}^n_+:=\left\{A\in{\cal S}^n \mid x^TAx\ge 0, \  x\in {\bf 1}^{\perp}\right\},
\end{equation}
where ${\bf 1}^{\perp}:=\{x\in\Re^n\mid{\bf 1}^Tx=0\}$.
It is well-known \citep{Schoenberg35,YHouseholder38} that $D\in{\cal S}^n$ is EDM if and only if
\[
-D\in {\cal S}_h^n\cap{\cal K}^n_+.
\]
Moreover, the embedding dimension is determined by the rank of the doubly centered matrix $JDJ$, i.e.,
\[
    r = \rank(JDJ) \qquad \mbox{and} \qquad J:=I-{\bf 1}{\bf 1}^T/n,
\]
where $J$ is known as the centering matrix. 
%See also Gower \cite{Gower85} for a nice derivation.

Since $-JDJ$ is positive semidefinite, its spectral decomposition can be written as
\[
 - \frac 12 JDJ = P \diag(\lambda_1, \ldots, \lambda_n) P^T,
\]
where $P^TP = I$ and
 $\lambda_1 \ge \lambda_2 \ge \cdots \ge \lambda_n \ge 0$ are the eigenvalues in the nonincreasing order.
Since $\rank(JDJ) =r$, we must have $\lambda_i = 0$ for all $i \ge (r+1)$.
Let $P_1$ be the submatrix consisting of the first $r$ columns (eigenvectors) in $P$.
One set of the embedding points are
\begin{equation} \label{Eq:cMDS}
   \left( \begin{array}{l}
	  p_1^T \\ \vdots \\
		p_n^T
	\end{array} \right) = P_1 \diag(\sqrt{\lambda_1}, \ldots, \sqrt{\lambda_r}).
\end{equation}

cMDS is built upon the above result. Suppose a pre-distance matrix $D$ (i.e., $D \in {\cal S}^n_h$ and $D \ge 0$)
is known. It computes the embedding points by (\ref{Eq:cMDS}).
Empirical evidences have shown that if the first $r$ eigenvalues are positive and the absolute values of the
remaining eigenvalues (they may be negative as $D$ may not be a true EDM)
are small, then cMDS often works well. Otherwise, it may produce mis-leading embedding points.
For example, there are examples that show that ISOMAP might cut off too many eigenvalues, hence
failing to produce satisfactory embedding (see e.g., Teapots data example in \cite{WS06}).
Both MVU and MVE models aim to avoid such situation.

The EDM score has been widely used to interpret the percentage of the total variance being
explained by the embedding from leading eigenvalues.
The EDM score of the leading $k$ eigenvalues is defined by
\[
  \mbox{EDMscore}(k) := \sum_{i=1}^k \lambda_i / \sum_{i=1}^n \lambda_i, \qquad k =1,2, \ldots, n.
\]
It is only well defined when $D$ is a true EDM.
The justification of using EDM scores is deeply rooted in the classic work of \cite{Gower66},
who showed that cMDS is a method of principal component analysis, but working with
EDMs instead of correlation matrices.

The centering matrix $J$ plays an important role in our analysis. It is the orthogonal projection onto the
subspace ${\bf 1}^{\perp}$ and hence $J^2 = J$. Moreover, we have the following. Let ${\cal S}^n_c$ be the
geometric center subspace in ${\cal S}^n$:
\begin{equation} \label{Eq:geometric_center}
  {\cal S}^n_c : = \left\{ Y \in {\cal S}^n \ | \ Y {\bf 1}  = 0 \right\}.
\end{equation}
Let ${\cal P}_{{\cal S}^n_c}(X)$ denote the orthogonal projection onto ${\cal S}^n_c$. Then we have
$
  {\cal P}_{{\cal S}^n_c}(X) = JXJ.
$
 That is, the doubly centered matrix $JXJ$, when viewed as a linear transformation of $X$,
 is the orthogonal projection of $X$ onto ${\cal S}^n_c$. Therefore, we have
\be \label{JXJ}
  \langle JXJ,  \; X-JXJ\rangle =0.
\ee
It is also easy to verify the following result.

\begin{lemma}\label{lemma:observation-hollow}
For any $X\in{\cal S}_h^n$, we have
\[
X-JXJ=\frac{1}{2}\left({\rm diag}(-JXJ)\,{\bf 1}^T+{\bf 1}\,{\rm diag}(-JXJ)^T\right).
\]
\end{lemma}

%%%%%%%%%%%%%%%%%%%%%%%%%%%%%%%%%%%%%%%%%%%%%
\subsection{MVU and MVE Models}

The input of MVU and MVE models is a set of partially observed distances
\[
  \left\{ d_{ij}^2: \ (i,j) \in \Omega_0 \right\} \qquad \mbox{and} \qquad
	\Omega_0 \subseteq \Omega := \left\{ (i, j) : \ 1 \le i < j \le n \right\}.
\]
Let $\{p_i\}_{i=1}^n$ denote the desired embedding points in $\Re^r$.
They should have the following properties. The pairwise distances should be
faithful to the observed ones. That is,
\be \label{Eq-dij}
 \| p_i - p_j \|^2 \approx d_{ij}^2 \qquad \forall \ (i, j) \in \Omega_0
\ee
and those points should be geometrically centered in order to remove the
translational degree of freedom from the embedding:
\be \label{Eq-Center}
  \sum_{i1}^n p_i = 0.
\ee
Let $K := \sum_{i=1}^n p_ip_i^T$ be the Gram matrix of the embedding points.
Then the conditions in (\ref{Eq-dij}) and (\ref{Eq-Center}) are translated to
\[
  K_{ii} - 2K_{ij} + K_{jj} \approx d_{ij}^2 \qquad \forall \ (i, j) \in \Omega_0
\]
and
\[
   \langle {\bf 1} {\bf 1}^T, \; K \rangle = 0.
\]
To encourage the dimension reduction, MVU argues that the variance, which is $\Tr(K)$, should be
maximized. In summary, the slack model (or the least square penalty model) of MVU takes
the following form:

\be \label{MVU}
 \begin{array}{ll}
 \max & \langle I, \; K \rangle - \nu \sum_{(i, j)\in \Omega_0}
\left( K_{ii} - 2K_{ij} + K_{jj} - d_{ij}^2 \right)^2 \\ [1ex]
 \mbox{s.t.} & \langle {\bf 1} {\bf 1}^T, \; K \rangle = 0
\quad \mbox{and} \quad
 K \succeq 0,
\end{array}
\ee
where $\nu>0$ is the penalty parameter that balances the trade-off between maximizing variance
and preserving the observed distances.
See also \cite{WSZS06, SBXD06} for more variants of this problem.

The resulting EDM $D \in {\cal S}^n$ from the optimal solution of (\ref{MVU}) is
defined to be
\[
   D_{ij} = K_{ii} - 2K_{ij} + K_{jj},
\]
and it satisfies
$
   K = -0.5 JDJ.
$
Empirical evidence shows that the EDM scores of the first few leading eigenvalues of $K$ are
often large enough to explain high percentage of the total variance.

MVE seeks to improve the EDM scores in a more aggressive way.
Suppose the targeted embedding dimension is $r$. MVE tries to maximize the eigen gap
between the leading $r$ eigenvalues of $K$ and the remaining eigenvalues.
This gives rise to
\[
 \begin{array}{ll}
 \max & \sum_{i=1}^r \lambda_i(K) - \sum_{i=r+1}^n \lambda_i(K) \\ [1ex]
 \mbox{s.t.} & K_{ii} - 2K_{ij} + K_{jj} \approx d_{ij}^2 \qquad \forall \ (i,j) \in \Omega_0 \\ [1ex]
             & \langle {\bf 1} {\bf 1}^T, \; K \rangle = 0
\quad \mbox{and} \quad
 K \succeq 0.
\end{array}
\]
There are a few standard ways in dealing with the constraints corresponding to $(i, j) \in \Omega_0$.
We are interested in the MVE slack model:
\be \label{MVE}
\begin{array}{ll}
 \max & \sum_{i=1}^r \lambda_i(K) - \sum_{i=r+1}^n \lambda_i(K) - \nu \sum_{(i, j)\in \Omega_0}
\left( K_{ii} - 2K_{ij} + K_{jj} - d_{ij}^2 \right)^2 \\ [1ex]
 \mbox{s.t.} & \langle {\bf 1} {\bf 1}^T, \; K \rangle = 0
\quad \mbox{and} \quad
 K \succeq 0,
\end{array}
\ee
where $\nu >0$. The MVE model (\ref{MVE}) often yields higher EDM scores than the MVU model (\ref{MVU}).
However, (\ref{MVU}) is a SDP problem while (\ref{MVE}) is nonconvex, which can be solved
 by a sequential SDP method \citep[see][]{SJebara07}.

%%%%%%%%%%%%%%%%%%%%%%%%%%%%%%%%%%%%%%%%%%%%%%
\subsection{Distance Sampling Rules}\label{subsection:Distance Sampling Rules}

In this part, we describe how the observed distances indexed by $\Omega_0$ are selected in practice.
We assume that those distances are sampled from unknown true Euclidean distances $\overline{d}_{ij}$
in the following fashion.
\be \label{Sampling-Model}
  d_{ij} = \overline{d}_{ij} + \eta \xi_{ij}, \qquad (i,j) \in \Omega_0,
\ee
where $\xi_{ij}$ are i.i.d. noise variables with $\mathbb{E}(\xi)=0$,
 $\mathbb{E}(\xi^2)=1$ and $\eta>0$ is a noise magnitude control factor.
We note that in (\ref{Sampling-Model}) it is the true Euclidean distance
$\overline{d}_{ij}$ (rather than its squared quantity) that is being sampled.
There are three commonly used rules to select $\Omega_0$.

%%%%%%%%%%%%%%%%%%%%%%%%%%%%%%%%%%%%%%%%%%%%%%%%%%%%%%%%%%%%%%%%%5%%
\begin{itemize}

\item[(i)] {\bf Uniform sampling rule}. The elements are independently and identically
sampled from $\Omega$ with the common probability ${1}/{|\Omega|}$.

\item[(ii)] {\bf $k$ nearest neighbors $(k$-NN) rule}. $(i,j) \in \Omega_0$ if and only if
$d_{ij}$ belongs to the first $k$ smallest distances in $\{d_{ij}: j=1, \ldots, n \}$.

\item[(iii)] {\bf Unit ball rule}. For a given radius $\epsilon>0$,
$(i,j) \in \Omega_0$ if and only if $d_{ij} \le \epsilon$.

\end{itemize}
%%%%%%%%%%%%%%%%%%%%%%%%%%%%%%%%%%%%%%%%%%%%%%%%%%%%%%%%%%%%%%%%%%%%

The $k$-NN and the unit ball rules are often used in low-dimensional manifold learning in order to
preserve the local structure of the embedding points, while the uniform sampling rule is often employed
in some other dimensionality reductions including embedding social network in a low-dimensional space.

\section{A Convex Optimization Model for Distance Learning} \label{Section-New-Model}
%%%%%%%%%%%%%%%%%%%%%%%%%%%%%%%%%%%%%%%%%%%%

Both MVU and MVE are trusted distance learning models in the following sense.
They both produce a Euclidean distance matrix, which is faithful to the observed distances and
they both encourage high EDM scores from the first few leading eigenvalues.
However, it still remains a difficult (theoretical) task to quantify how good the resulting embedding is.
In this part, we will propose a new learning model, which inherit
the good properties of MVU and MVE.
Moreover, we are able to quantify by deriving error bounds of the resulting solutions under the
uniform sampling rule. Below, we first describe our model, followed by detailed interpretation.

%%%%%%%%%%%%%%%%%%%%%%%%%%%%%%%%%%%%%%%%%%%%%%%%%%%%%%
\subsection{Model Description} \label{Subsection-Model-Description}

In order to facilitate the description of our model and to set the platform for our subsequent analysis, we write
the sampling model (\ref{Sampling-Model}) as an observation model. Define two matrices
$\overline{D}$ and
$\overline{D}^{(1/2)}$ respectively by
\[
  \overline{D} : = \Big(  \overline{d}_{ij}^2 \Big) \qquad \mbox{and} \qquad
	\overline{D}^{(1/2)} := \Big( \overline{d}_{ij} \Big).
\]
A sampled basis matrix $X$ has the following form:
\[
  X := \frac{1}{{2}} (e_ie_j^T+e_je_i^T) \qquad \mbox{for some} \ (i, j) \in \Omega.
\]
For each $(i,j) \in \Omega_0$, there exists a corresponding sampling basis matrix.
We number them as $X_1, \ldots, X_m$.

Define the corresponding observation operator ${\cal O}:{\cal S}^n\to\Re^{m}$ by
\begin{equation}\label{eq:def_obser_op}
{\cal O}(A):=\left(\langle X_1,A\rangle,\ldots,\langle X_m,A\rangle\right)^T, \quad A\in{\cal S}^n.
\end{equation}
That is, ${\cal O}(A)$ samples all the elements $A_{ij}$ specified by $(i,j) \in \Omega_0$. Let ${\cal O}^{*}:\Re^{m}\to{\cal S}^{n}$ be its adjoint, i.e.,
\[
{\cal O}^{*}(z)=\sum_{l=1}^{m}z_{l}X_{l}, \quad z\in\Re^{m}.
\]
 Thus, the sampling model \eqref{Sampling-Model} can be re-written as the following compact form
\begin{equation}\label{eq:observation_model}
y = {\cal O}(\overline{D}^{(1/2)})+\eta\xi,
\end{equation}
where $y=(y_1,\ldots,y_m)^T$ and $\xi=(\xi_1,\ldots,\xi_m)^T$ are the observation vector and the noise vector, respectively.

Since $-J\overline{D}J\in{\cal S}^n_+$, we may assume that $-J\overline{D}J\in{\cal S}^n_+$ has the following single values decomposition (SVD):
\begin{equation}\label{eq:SVD-JbarDJ}
-J\overline{D}J=\overline{P}{\rm Diag}(\overline{\lambda})\overline{P}^T,
\end{equation}
where $\overline{P}\in{\mathbb O}^n$ is an orthogonal matrix,
$\overline{\lambda}=(\overline{\lambda}_1,\overline{\lambda}_2,\ldots,\overline{\lambda}_n)^T\in\Re^n$
is the vector of the eigenvalues of $-J\overline{D}J$ arranged in nondecreasing order,
i.e., $\overline{\lambda}_1\ge\overline{\lambda}_2\ge\ldots\ge\overline{\lambda}_n\ge 0$.

Suppose that $\widetilde{D}$ is a  given initial estimator of the unknown matrix $\overline{D}$, 
%We also assume that $-\widetilde{D}$ belongs the almost positive semidefinite matrices cone ${\cal K}^n_+$, which is defined in \eqref{eq:def_alPSD}.
%Thus, we may assume that $-J\widetilde{D}J\in{\cal S}_+^n$ has the following single value decomposition
and it has the following single value decomposition
\[
-J\widetilde{D}J=\widetilde{P}{\rm Diag}(\widetilde{\lambda})\widetilde{P}^T,
\]
where $\widetilde{P}\in{\mathbb O}^n$. In this paper, we always assume the embedding dimension $r:={\rm rank}(J\overline{D}J)\ge 1$. Thus, for any given orthogonal matrix $P\in{\mathbb O}^n$, we write $P=[P_1\ \ P_2]$ with $P_1\in\Re^{n\times r}$ and $P_2\in\Re^{n\times (n-r)}$.
For the given parameters $\rho_1>0$ and $\rho_2\ge0$,
we consider the following convex optimization problem
\begin{equation}\label{eq:estimator_problem}
\begin{array}{rl}
\min_{D \in S^n}
 & \displaystyle{\frac{1}{2m}}\| y\circ y-{\cal O}(D)\|^2+\rho_1\left(\langle I, -JDJ \rangle -\rho_2\langle \widetilde{P}_1\widetilde{P}_1^T,-JDJ\rangle\right) \\[1ex]
{\rm s.t.} & D\in{\cal S}^n_h, \quad -D\in{\cal K}^n_+.
\end{array}
\end{equation}
This problem has EDM as its variable and this is in contrast to MVU, MVE and other learning models
\cite[e.g.,][]{JMontanari13} where they all use SDPs.
The use of EDMs greatly benefit us in deriving the error bounds in the next section.
Our model (\ref{eq:estimator_problem}) tries to accomplish three tasks as we explain below.

%Note that by setting $\rho_{2}=2$, the estimation model \eqref{eq:estimator_problem} reduces to  the Minimum Volume Embedding model \cite{SJebara07}. Also, if $\rho_{2}=0$, then the estimation model \eqref{eq:estimator_problem} is just the nuclear norm penalized least squares model.

%%%%%%%%%%%%%%%%%%%%%%%%%%%%%%%%%%%%%%%%%%%%%%%%%%%%%%%%%%%%%%%%%%%%%%%%%%%%%%%%%%%%%%%%%%%%%%
\subsection{Model Interpretation}

The three tasks that model (\ref{eq:estimator_problem}) tries to accomplish correspond to
the three terms in the objective function.
The first (quadratic) term is nothing but
\[
   \sum_{(i,j) \in \Omega_0} ( d_{ij}^2 - D_{ij} )^2
\]
corresponding to the quadratic terms in the slack models (\ref{MVU}) and (\ref{MVE}).
Minimizing this term is essentially to find an EDM $D$ that minimizes the
error rising from the sampling model (\ref{eq:observation_model}).

The second term $\langle I, \; -JDJ \rangle$ is actually the nuclear norm of $(-JDJ)$.
Recall that in cMDS, the embedding points in (\ref{Eq:cMDS}) come from the spectral
decomposition of $(-JDJ)$. Minimizing this term means to find the smallest embedding dimension.
However, as argued in both MVU and MVE models, minimizing the nuclear norm is against
the principal idea of maximizing variance.
Therefore, to alleviate this confliction, we need the third term
$- \langle \widetilde{P}_1 \widetilde{P}_1^T, \; -JDJ \rangle$.

In order to motivate the third term, let us consider an extreme case.
Suppose the initial EDM $\widetilde{D}$ is close enough to $D$ in the sense that the
leading eigenspaces respectively spanned by $\{ \widetilde{P}_1\}$ and by $\{P_1\}$
coincide. That is $\widetilde{P}_1 = P_1$. Then,
\[
   \langle \widetilde{P}_1 \widetilde{P}_1^T, \; -JDJ \rangle
	= \sum_{i=1}^r \lambda_i.
\]
Hence, minimizing the third term is essentially maximizing the leading eigenvalues of
$(-JDJ)$.
Over the optimization process, the third term is likely to push the quantity
\[
  t := \sum_{i=1}^r \lambda_i
\]
up, and the second term (nuclear norm) forces the remaining eigenvalues
\[
  s := \sum_{i=r+1}^n \lambda_i
\]
down.
The consequence is that the EDM score
\[
  \mbox{EDMscore}(r) = f(t,s) := \frac{t}{t+s}
\]
gets higher. This is because
\[
 f(t_2, s_2) > f(t_1, s_1) \qquad \forall \ t_2 > t_1 \quad \mbox{and} \quad
 s_2 < s_1.
\]
Therefore, the EDM scores can be controlled by controlling the penalty parameters
$\rho_1$ and $\rho_2$.
The above heuristic observation is in agreement with our extensive numerical experiments.

Model (\ref{eq:estimator_problem}) also covers the MVE model as a special case.
Let $\rho_2 = 2$ and $\widetilde{D}$ to be one of the iterates in the MVE SDP subproblems.
The combined term
\[
  \langle I, \; JDJ \rangle - 2 \langle \widetilde{P}_1 \widetilde{P}_1^T, \ -JDJ \rangle
\]
is just the objective function in the MVE SDP subproblem.
In other words, MVE keeps updating $\widetilde{D}$ by solving the SDP subproblems.
Therefore, our error-bound results also justify why MVE often leads to higher EDM scores.

Before we go on to derive our promised error-bound results, we summarize the key points for
our model (\ref{eq:estimator_problem}).
It is EDM based rather than SDP based as in the most existing research.
The use of EDM enables us to establish the error-bound results in the next section.
It inherits the nice properties in MVU and MVE models.
We will also show that this model can be efficiently solved.

%%%%%%%%%%%%%%%%%%%%%%%%%%%%%%%%%%%%%%%%%%%%%%%%%%%%%%%%%%%%%%%%%%%%%%%%%%%%%%%%%%%%%%%%%%%%%%%
\section{Error Bounds Under Uniform Sampling Rule}\label{section:Error bounds}

Suppose that $X_1,\ldots,X_m$ are $m$ independent and identically distributed (i.i.d.) random matrices over $\Omega$ with the common\footnote{This assumption can be replaced 
by any positive probability $p_{ij} >0$. But it would complicate the notation used.} 
probability $1/|\Omega|$, i.e., for any $1\le i<j \le n$,
\[
\mathbb{P}\left(X_{l}=\frac{1}{2}(e_ie_j^T+e_je_i^T)\right)=\frac{1}{|\Omega|},\quad l=1,\ldots,m.
\]
Thus, for any $A\in{\cal S}^n_h$, we have
\begin{equation}\label{eq:E2-bounded}
{\mathbb E}\left(\langle A,X \rangle^2 \right)= \frac{1}{2|\Omega|}\|A\|^2.
\end{equation}
Moreover, we assume that the i.i.d. noise variables in \eqref{Sampling-Model} have the bounded fourth moment, i.e., there exists a constant $\gamma>0$ such that ${\mathbb E}(\xi^4)\le\gamma$.

Let $\overline{D}$ be the unknown true EDM. Suppose that the positive semidefinite matrix $-J\overline{D}J$ has the singular value decomposition \eqref{eq:SVD-JbarDJ}
and $\overline{P} = [\overline{P}_1, \overline{P}_2]$ with $\overline{P}_1 \in \Re^{n \times r}$.
We define the generalized geometric center subspace in ${\cal S}^n$ by (compare to \eqref{Eq:geometric_center})
\[
  T := \left\{ Y \in {\cal S}^n \ | \ Y \overline{P}_1 = 0 \right\}.
\]
Let $T^{\perp}$ be its orthogonal subspace.
The orthogonal projections to the two subspaces can hence be calculated respectively by
\[
{\cal P}_{T}(A):=\overline{P}_2\overline{P}_2A\overline{P}_2\overline{P}_2^T 
\quad {\rm and} \quad
{\cal P}_{T^{\perp}}(A):=\overline{P}_1\overline{P}_1^TA+A\overline{P}_1\overline{P}_1^T-\overline{P}_1\overline{P}_1^TA\overline{P}_1\overline{P}_1^T  .
\]
%Before considering the error bounds of \eqref{eq:estimator_problem}, we first introduce some notations and results, which are needed in subsequent discussions.  For any $A\in{\cal S}^n$, define
%\[
%{\cal P}_{T}(A):=\overline{P}_1\overline{P}_1^TA+A\overline{P}_1\overline{P}_1^T-\overline{P}_1\overline{P}_1^TA\overline{P}_1\overline{P}_1^T \quad {\rm and} \quad {\cal P}_{T^{\perp}}(A):=\overline{P}_2\overline{P}_2A\overline{P}_2\overline{P}_2^T.
%\]
It is clear that we have the following orthogonal decomposition
\begin{equation}\label{eq:decomp_T}
A={\cal P}_{T}(A)+{\cal P}_{T^{\perp}}(A) \quad {\rm and} \quad \langle {\cal P}_{T}(A),{\cal P}_{T^{\perp}}(B) \rangle =0  \quad \forall\, A,B\in{\cal S}^n.
\end{equation}
Moreover, we know from the definition of ${\cal P}_T$ that for any $A\in{\cal S}^n$,
\[
{\cal P}_{T^{\perp}}(A)=\overline{P}_1\overline{P}_1^TA+\overline{P}_2\overline{P}_2^TA\overline{P}_1\overline{P}_1^T,
\]
which implies that ${\rm rank}({\cal P}_{T^{\perp}}(A))\leq 2r$. This yields for any $A\in{\cal S}^n$
\begin{equation}\label{eq:PT_bounds}
\|{\cal P}_{T^{\perp}}(A)\|_*\le \sqrt{2r}\|A\|.
\end{equation}
For given $\rho_2\ge0$, define
\begin{equation}\label{eq:notation_alpha_beta}
\alpha(\rho_2):=\frac{1}{\sqrt{2r}}\|\overline{P}_1\overline{P}_1-\rho_2\widetilde{P}_1\widetilde{P}_1^T\|.
\end{equation}
Let $\zeta:=(\zeta_{1},\ldots,\zeta_{m})^{T}$ be the random vector defined by
\begin{equation}\label{eq:def_new_noise}
\zeta=2{\cal O}(\overline{D}^{(1/2)})\circ\xi+\eta(\xi\circ\xi).
\end{equation}

The non-commutative Bernstein inequality provides the probability bounds of the difference between the sum of independent random matrices and its mean under the spectral norm \citep[see, e.g.,][]{Recht11,Tropp11,Gross11}. The following Bernstein inequality is taken from \citep[Lemma 7]{NWainwright12}, where the independent random matrices are bounded under the spectral norm or bounded under the $\psi_1$ Orlicz norm of random variables, i.e.,
\[
\|x\|_{\psi_1}:=\inf\left\{t>0\mid\mathbb{E}\,{\rm exp}(|x| /t )\le e\right\}.
\]

\begin{lemma}\label{lem:Bernstein-ineq-sym}
Let $Z_1,\ldots,Z_m\in{\cal S}^n$ be independent random symmetric matrices with mean zero. Suppose that there exists $M>0$, for all $l$, $\|Z_l\|_2\le M$ or $\big\|\|Z_l\|_2\big\|_{\psi_1}\le M$. Denote $\sigma^2:=\|\mathbb{E}(Z_l^2)\|_2$.
Then, we have for any $t>0$,
\[
{\mathbb P}\Big(\big\|\frac{1}{m}\sum_{l=1}^{m}Z_l\big\|_2\ge t\Big)\leq 2n\max\left\{ {\rm exp}\left(-\frac{mt^2}{4\sigma^2}\right), {\rm exp}\left(-\frac{mt}{2M}\right) \right\}.
\]
\end{lemma}

Now we are ready to study the error bounds of the model \eqref{eq:estimator_problem}. Denote the optimal solution of \eqref{eq:estimator_problem} by $D^*$. 
The following result represents the first major step to derive our ultimate bound result.
It contains two bounds.
The first bound \eqref{eq:Err_origin_1} is on the norm-squared distance betwen $D^*$ and $\overline{D}$ 
under the observation operator ${\cal O}$.
The second bound \eqref{eq:Err_origin_2} is about the nuclear norm of $D^* - \overline{D}$.
Both bounds are in terms of the Frobenius norm of $D^* - \overline{D}$.

\begin{proposition}\label{prop:Error_bound_origin}
Let $\zeta=(\zeta_{1},\ldots,\zeta_{m})^{T}$ be the random vector defined in \eqref{eq:def_new_noise} and $\kappa>1$ be given. Suppose that $\rho_1\ge  {\kappa\eta} \big\|\frac{1}{m}{\cal O}^*(\zeta) \big\|_2$ and $\rho_2\ge 0$,
where ${\cal O}^*$ is the adjoint operator of ${\cal O}$.
 Then, we have
\begin{equation}\label{eq:Err_origin_1}
\frac{1}{2m}\|{\cal O}(D^*-\overline{D})\|^2 \leq \left(\alpha(\rho_2)+\frac{2}{\kappa}\right)\rho_1\sqrt{2r}\|D^*-\overline{D}\|
\end{equation}
and
\begin{equation}\label{eq:Err_origin_2}
\|D^*-\overline{D}\|_* \leq \frac{\kappa}{\kappa-1}\left( \alpha(\rho_2) +2\right)\sqrt{2r}\|D^*-\overline{D}\|.
\end{equation}
\end{proposition}

The second major technical result below shows that the sampling operator ${\cal O}$ 
satisfies the following restricted strong convexity \citep{NWainwright12} in the set ${\cal C}(\tau)$
for any $\tau >0$, where
\[
{\cal C}(\tau):=\left\{A\in{\cal S}^n_h \mid \|A\|_{\infty}=\frac{1}{\sqrt{2}},\ \|A\|_*\le\sqrt{\tau}\|A\|,\ {\mathbb E}(\langle A,X\rangle^2)\ge \sqrt{\frac{256\log(2n)}{m\log(2)}} \right\}.
\]

\begin{lemma}\label{lemma:approx_RIP}
Let $\tau>0$ be given. Suppose that $m>C_1n\log (2n)$, where $C_1>1$ is a  constant. Then, there exists a constant $C_2>0$ such that for any $A\in{\cal C}(\tau)$, the following inequality holds with probability at least $1-1/n$.
\[
\frac{1}{m}\|{\cal O}(A)\|^2\ge \frac{1}{2}{\mathbb E}\left(\langle A,X\rangle^2\right)-256C_2\tau|\Omega|\frac{\log (2n)}{nm}.
\]
%where $\delta_m:=\mathbb{E}(\left\|\frac{1}{m}{\cal R}^*_{\Omega}(\epsilon) \right\|_2)$ with $\epsilon=(\epsilon_1,\ldots,\epsilon_m)^T$ and $\{\epsilon_1,\ldots,\epsilon_m\}$ is an i.i.d. Rademacher sequence, i.e., a sequence of i.i.d. Bernoulli random variables taking the values $1$ and $-1$ with probability $1/2$.
\end{lemma}

Next, combining Proposition \ref{prop:Error_bound_origin} and Lemma \ref{lemma:approx_RIP}
leads to the following result.

\begin{proposition}\label{prop:error_bound_1}
Assume that there exists a constant $b>0$ such that $\|\overline{D}\|_\infty\le b$. Let $\kappa>1$ be given. Suppose that $\rho_1\ge \displaystyle{ \kappa\eta }\big\|\frac{1}{m}{\cal O}^*(\zeta) \big\|_2$ and  $\rho_2\ge 0$. Furthermore, assume that $m>C_1n\log (2n)$ for some constant $C_1>1$. Then, there exists a constant $C_3>0$ such that with probability at least $1-1/n$,
\[
\frac{\|D^*-\overline{D}\|^2}{|\Omega|}\leq C_3\max\left\{r|\Omega|\Big(\big(\alpha(\rho_2)+\frac{2}{\kappa}\big)^2\rho_1^2+\big(\frac{\kappa}{\kappa-1}\big)^2\big(\alpha(\rho_2)+2\big)^2b^2\frac{\log (2n)}{nm}\Big),b^{2}\sqrt{\frac{\log (2n)}{m}} \right\}.
\]
\end{proposition}

This bound depends on the model parameters $\rho_1$ and $\rho_2$.
In order to establish an explicit error bound, we need to estimate $\rho_1$
($\rho_2$ will be estimated later), which depends on the quantity 
$\left\|\frac{1}{m}{\cal O}^*(\zeta) \right\|_2$, 
where $\zeta=(\zeta_1,\ldots,\zeta_m)^T\in\Re^m$ with $\zeta_l$, $l=1,\ldots,m$ are i.i.d. 
random variables given by \eqref{eq:def_new_noise}. 
To this end, from now on, we always assume that the i.i.d. random noises $\xi_l$, $l=1,\ldots,m$ 
in the sampling model \eqref{eq:observation_model} satisfy the following sub-Gaussian tail condition.

\begin{assumption}\label{ass:sub-Gaussian}
There exist positive constants $K_1$ and $K_2$ such that for all $t>0$,
\[
{\mathbb P}\left( |\xi_l|\geq t\right)\leq K_1{\rm exp}\left( - t^2/K_2\right).
\]
\end{assumption}

By applying the Bernstein inequality (Lemma \ref{lem:Bernstein-ineq-sym}), we have 
%the following estimation results on the quantity $\left\|\frac{1}{m}{\cal O}^*(\zeta) \right\|_2$.

\begin{proposition}\label{prop:estimator_R_Omega_xi}
Let $\zeta=(\zeta_{1},\ldots,\zeta_{m})^{T}$ be the random vector defined in \eqref{eq:def_new_noise}. Assume that the noise magnitude control factor satisfies $\eta<\omega:=\|{\cal O}(\overline{D}^{(1/2)})\|_{\infty}$. Suppose that there exists $C_1>1$ such that $m>C_1n\log(n)$.  Then, there exists a constant $C_3>0$ such that with probability at least $1-1/n$,
\begin{equation}\label{eq:R_Omega_ineq}
\left\|\frac{1}{m}{\cal O}^*(\zeta) \right\|_2\le C_3\omega\sqrt{\frac{\log(2n)}{nm}}.
\end{equation}
\end{proposition}

This result suggests that $\rho_1$ can take the particular value:
\begin{equation}\label{eq:choose_rho_1}
\rho_1= \kappa\eta\omega C_3\sqrt{\frac{\log(2n)}{mn}},
\end{equation}
where $\kappa >1$.
Our final step is to combine Proposition \ref{prop:error_bound_1} 
and Proposition \ref{prop:estimator_R_Omega_xi} to get the following error bound.

\begin{theorem}\label{thm:error_bound_2}
Suppose that there exists a constant $b>0$ such that $\|\overline{D}\|_\infty\le b$, and the noise magnitude control factor satisfies $\eta<\omega=\|{\cal O}(\overline{D}^{(1/2)})\|_{\infty}$.  Assume the sample size $m$ satisfies $m>C_1n\log (2n)$ for some constant $C_1>1$.  For any given $\kappa>1$, let $\rho_1$ be given by \eqref{eq:choose_rho_1} and $\rho_2\ge 0$. Then, there exists a constant $C_4>0$ such that with probability at least $1-2/n$,
\begin{equation}\label{eq:error_bound_2}
\frac{\|D^*-\overline{D}\|^2}{|\Omega|}\leq C_4\Big(\big(\kappa \alpha(\rho_2) +2\big)^2\eta^2\omega^{2}+\frac{\kappa^2}{(\kappa-1)^2}\big( \alpha(\rho_2) +2\big)^2b^2\Big)\frac{r|\Omega|\log (2n)}{nm}.
\end{equation}
\end{theorem}

The only remaining unknown parameter in \eqref{eq:error_bound_2} is $\rho_2$ though $\alpha(\rho_2)$.
It follows from \eqref{eq:notation_alpha_beta} that
\begin{equation}\label{eq:notation_alpha_beta2}
(\alpha(\rho_2))^2 
%= \frac{1}{2r}\|\overline{P}_1\overline{P}_1^T-\rho_2\widetilde{P}_1\widetilde{P}_1^T\|^2
=\frac{1}{2r}\left(\|\overline{P}_1\overline{P}_1^T\|^2
  - 2\rho_2\langle\overline{P}_1\overline{P}_1^T,\widetilde{P}_1\widetilde{P}_1^T \rangle
	+ \rho_2^2\|\widetilde{P}_1\widetilde{P}_1^T\|^2\right).
\end{equation}
Since $\|\overline{P}_1\overline{P}_1^T\|^2=\|\widetilde{P}_1\widetilde{P}_1^T \|^2=r$ and $\langle\overline{P}_1\overline{P}_1^T,\widetilde{P}_1\widetilde{P}_1^T \rangle\ge 0$, we can bound $\alpha(\rho_2)$ by
\[
 (\alpha(\rho_2))^2 \le \frac r2 (1 + \rho_2^2).
\]
This bound seems to suggest that $\rho_2 =0$ (corresponding to the nuclear norm minimization) would lead to
a lower bound than other choices. In fact, there are better choices.
The best choice $\rho_2^*$ for $\rho_2$ is when it minimizes the right-hand side bound
and is given by \eqref{eq:rho2-1} in Subsection \ref{subsec:rho_2},
where we will show that $\rho_2 =1$ is a better choice  than both
$\rho_2 = 0$ and $\rho_2 = 2$ (see Proposition \ref{prop:better-choice-rho2}).

The major message from Theorem \ref{thm:error_bound_2} is as follows.
We know that if the true Euclidean distance matrix $\overline{D}$ is bounded, and the noises are small (less than the true distances), in order to control the estimation error, we only need samples with the size $m$ of the order $r(n-1)\log(2n)/2$, since $|\Omega|=n(n-1)/2$. Note that, $r= {\rm rank}(J\overline{D}J)$ is usually small ($2$ or $3$). Therefore, the sample size $m$ is much smaller than $n(n-1)/2$, the total number of the off-diagonal entries. Moreover, since the degree\footnote{We know from Lemma \ref{lemma:observation-hollow} 
that the rank of the true EDM ${\rm rank}(\overline{D})=O(r)$.} 
of freedom of $n$ by $n$ symmetric hollow matrix with rank $r$
is $n(r-1)-r(r-1)/2$, the sample size $m$ is close to the degree of freedom if the matrix size $n$ is large enough.

%%%%%%%%%%%%%%%%%%%%%%%%%%%%%%%%%%%%%%%%%%%%%%%%%%%%%%%%%%%%%%%%%%%%%%%%%%%%%%%%%%%%%%%%%%%%%%%
%\section{An Inexact Accelerated Proximal Gradient Method} \label{Section-IAPG}
%%%%%%%%%%%%%%%%%%%%%%%%%%%%%
\section{Model Parameter Estimation and the Algorithm} \label{Section-IAPG}
%%%%%%%%%%%%%%%%%%%%%%%%%%%%%

In general, the choice of model parameters can be tailored to a particular application.
A very useful property about our model \eqref{eq:estimator_problem}
is that we can derive a theoretical estimate, which serves as
a guideline for the choice of the model parameters in our implementation. 
In particular,
We set $\rho_1$ by (\ref{eq:choose_rho_1}) and prove that $\rho_2 =1 $ is a better choice than
both the case $\rho_2 =0$ (correponding to the nuclear norm minimization) and $\rho_2=2$ (MVE model).
The first part of this section is to study the optimal choice of $\rho_2$ and
the second part proposes an inexact accelerated proximal gradient method (IAPG)
that is particularly suitable to our model.

%%%%%%%%%%%%%%%%%%%%%%%%%%%%%%%
\subsection{Optimal Estimate of $\rho_2$} \label{subsec:rho_2}
%%%%%%%%%%%%%%%%%%%%%%%

It is easy to see from the inequality \eqref{eq:error_bound_2} that in order to reduce the estimation error, 
the best choice $\rho_2^*$ of $\rho_2$ is the minimum of $\alpha(\rho_2)$. 
We obtain from \eqref{eq:notation_alpha_beta2} that $\rho_2^* \ge 0$ and

\begin{equation}\label{eq:rho2-1}
\rho_2^*=\frac{1}{r}\langle\overline{P}_1\overline{P}_1^T,\widetilde{P}_1\widetilde{P}_1^T \rangle=1+\frac{1}{r}\langle \overline{P}_1\overline{P}_1^T,\widetilde{P}_1\widetilde{P}_1^T-\overline{P}_1\overline{P}_1^T\rangle.
\end{equation}
The key technique that we are going to use to estimate $\rho_2^*$ is the L\"{o}wner operator.
We express both the terms $\widetilde{P}_1\widetilde{P}_1^T$ and $\overline{P}_1\overline{P}_1^T$
as the values from the operator. We then show that the L\"{o}wner operator admits a first-order apprximation,
which will indicate the magnitude of $\rho_2^*$. 
The technique is extensively used by \cite{MPSun12}. 
We briefly describe it below.

Denote $\delta:=\|\widetilde{D}-\overline{D}\|$. 
Assume that $\delta<\overline{\lambda}_r/2$. Define the scalar function $\phi:\Re\to\Re$ by
\begin{equation}\label{eq:def-phi}
\phi(x)=\left\{\begin{array}{ll}
1 & \mbox{if $x\ge \overline{\lambda}_r-\delta$,} \\[3 pt]
\displaystyle{\frac{x-\delta}{\overline{\lambda}_r-2\delta}} & \mbox{if $\delta \le x \le \overline{\lambda}_r-\delta$,}\\ [3 pt]
0 & \mbox{if $x \le \delta$.}
\end{array}\right.
\end{equation}
Let $\Phi:{\cal S}^n\to{\cal S}^n$ be the corresponding L\"{o}wner operator with respect to $\phi$, i.e.,
\begin{equation}\label{eq:def-Phi-LOP}
\Phi(A)=P{\rm Diag}(\phi(\lambda_1(A)),\ldots, \phi(\lambda_n(A))) P^T, \quad A\in{\cal S}^n,
\end{equation}
where $P\in{\mathbb O}^n$ comes from the eigenvalue decomposition 
$A=P{\rm Diag}(\lambda_1(A),\ldots, \lambda_n(A))P^T$. 
Immediately we have $\Phi(-J \overline{D} J) = \overline{P}_1\overline{P}_1^T$. 
We show it is also true for $\widetilde{D}$.

It follows the perturbation result of Weyl for eigenvalues of symmetric matrices \cite[p.~63]{Bhatia97}
that
\[
  \| \overline{\lambda}_i -\widetilde{\lambda_i} \| \le \| J(\overline{D} - \widetilde{D})J \| 
	\le \| \overline{D} - \widetilde{D} \|, \qquad i=1, \ldots, n.
\]
We must have 
\[
  \widetilde{\lambda}_i \ge \overline{\lambda}_r - \delta \quad \mbox{for} \ i=1,\ldots, r
	\quad \mbox{and} \quad
	\widetilde{\lambda}_i \le \delta \quad \mbox{for} \ i=r+1, \ldots, n.
\]
We therefore have $\Phi(-J \widetilde{D} J) = \widetilde{P}_1\widetilde{P}_1^T$.

As a matter of fact,  the scalar function defined by \eqref{eq:def-phi} 
is twice continuously differentiable (actually, $\phi$ is analytic) 
on $(-\infty,\delta)\cup(\overline{\lambda}_r-\delta,\infty)$. 
Therefore, we know from \cite[Exercise V.3.9]{Bhatia97} that $\Phi$ is 
twice continuously differentiable near $-J\overline{D}J$ 
(actually, $\Phi$ is analytic near $-J\overline{D}J$). 
Therefore, %if the initial estimation $\widetilde{D}$ is sufficiently close to $\overline{D}$,
under the condition that  $\delta<\overline{\lambda}_r/2$,
we have by the derivative formula of the L\"{o}wner operator \citep[see, e.g.,][Theorem V.3.3]{Bhatia97} that
\begin{eqnarray*}
\widetilde{P}_1\widetilde{P}_1^T-\overline{P}_1\overline{P}_1^T=\Phi(-J\widetilde{D}J)-\Phi(-J\overline{D}J)&=&\Phi'(-J\overline{D}J)(-JHJ)+O(\|-JHJ\|^2)\\ [3pt]
&=&\overline{P}\left[ \overline{W}\circ(\overline{P}^{T}(-JHJ)\overline{P}) \right] \overline{P}^{T}+O(\|H\|^{2}),
\end{eqnarray*}
where $H:=\widetilde{D}-\overline{D}$ and $\overline{W}\in{\cal S}^{n}$ is given by
\[
(\overline{W})_{ij}:=\left\{ \begin{array}{ll}
\displaystyle{\frac{1}{\overline{\lambda}_i}} & \mbox{if $1\le i \le r$ and $r+1\le j \le n$,}\\ [3pt]
\displaystyle{\frac{1}{\overline{\lambda}_j}} & \mbox{if $r+1\le i \le n$ and $1\le j \le r$,}\\ [3pt]
0 & \mbox{otherwise},
\end{array} \right. \quad i,j\in\{1,\ldots,n\}.
\]
We note that the leading $r\times r$ block of $\overline{W}$ is $0$, which implies
\[
\langle \overline{P}_1\overline{P}_1^T,\overline{P}
\left[ \overline{W}\circ(\overline{P}^{T}(-JHJ)\overline{P}) \right] \overline{P}^{T}\rangle=0.
\] 
Therefore, we know from \eqref{eq:rho2-1} that if $\widetilde{D}$ is sufficiently close to $\overline{D}$,
\[
\rho_2^*=1+O(\|H\|^{2}).
\]
%which implies that $\rho_2^*\approx 1$ as the initial estimator $\widetilde{D}$ close to $\overline{D}$. 
%For the general case, under some mild assumptions on the initial estimator $\widetilde{D}$, 

This shows that $\rho_2 =1$ is nearly optimal if the initial estimator $\widetilde{D}$ is close to $\overline{D}$.
We will show that  in terms of the estimation errors the choice $\rho_{2}=1$ 
is always better than the nuclear norm penalized least squares model ($\rho_{2}=0$) and the minimum volume embedding model ($\rho_{2}=2$).

\begin{proposition}\label{prop:better-choice-rho2}
If $ \|\widetilde{D}-\overline{D}\|<\overline{\lambda}_{r}/2$, then
\[
   \alpha(1)<\min\left\{\alpha(0),\alpha(2)\right\}.
\]
\end{proposition}

%%%%%%%%%%%%%%%%%%%%%%%%%%%%%%%%%%%%%%%%%%%%%%%%%%%%%%%%%%%%%%%%%%%%%
\subsection{Inexact Accelerated Proximal Gradient Method}
%%%%%%%%%%%%%%%%%%%%%%%%%

Without loss of generality, we consider the following convex quadratic problem
\begin{equation}\label{eq:general_problem}
\begin{array}{cl}
\min & \frac{1}{2}\|{\cal A}(X)-a\|^2+\langle C,X \rangle \\ [3pt]
{\rm s.t.} & {\cal B}(X)=b,  \ \ X\in{\cal K}_+^n,
\end{array}
\end{equation}
where ${\cal K}_+^n$ is the almost positive semidefinite cone defined by \eqref{eq:def_alPSD}, $X,C\in{\cal S}^n$, $a\in\Re^m$, $b\in\Re^k$, and ${\cal A}:{\cal S}^n\to\Re^m$, ${\cal B}:{\cal S}^n\to\Re^k$ are two given linear operators. By setting ${\cal A}\equiv{\cal O}$, ${\cal B}\equiv{\rm diag}(\cdot)$, $a\equiv-(y\circ y)\in\Re^m$, $b\equiv 0\in\Re^n$ and $C\equiv m\rho_1J(I-\rho_2\widetilde{P}_1\widetilde{P}_1^T)J$, one can easily verify 
that \eqref{eq:general_problem} is equivalent with the trusted distance learning model \eqref{eq:estimator_problem}.

Being a convex quadratic problem, \eqref{eq:general_problem} can be solved in a number of ways.
Based on our extensive numerical experiments, we found that 
the inexact accelerated proximal gradient (IAPG) method, 
which is recently studied by \cite{JSToh12} 
for large scale linearly constrained convex quadratic SDP problems,
is particularly suitable to our problem. 
To describe this method, 
let us denote the objective function by $f(X):=\frac{1}{2}\|{\cal A}(X)-a\|^2+\langle C,X \rangle$ 
and the corresponding gradient by $\nabla f(X)={\cal A}^*({\cal A}(X)-a)+C$, where ${\cal A}^{*}:\Re^{m}\to{\cal S}^{n}$ is the adjoint of ${\cal A}$. The algorithm is described as follows.

\begin{algorithm}\label{alg:APG}
Choose the starting point $Y_1=X_0$. Let $t_1=1$. Set $k=1$. Perform the $k$-th iteration as follows:
\begin{description}
\item[Step 1] Find an approximate minimizer 
\begin{equation}\label{eq:approx_APG}
X_k\approx{\rm argmin}\left\{ q_k(X)\mid {\cal B}(X)=b,\ X\in{\cal K}_+^n  \right\},
\end{equation} 
where $q_k(X):=f(Y_k)+\langle \nabla f(Y_k),X-Y_k\rangle+\frac{1}{2}\langle X-Y_k, {\cal Q}_k(X-Y_k)\rangle$ such that $f(X)\leq q_k(X)$ for all $X\in{\cal S}^n$, and ${\cal Q}_k$ is a self-adjoint positive definite linear operator.

\item[Step 2] Compute $t_{k+1}=\displaystyle{\frac{1+\sqrt{1+4t_k^2}}{2}}$. 

\item[Step 3] Compute $Y_{k+1}=X_k+\displaystyle{\frac{t_k-1}{t_{k+1}}}(X_k-X_{k-1})$.
\end{description}
\end{algorithm}

It is noted that the major computational part is on the approximate solution of \eqref{eq:approx_APG}.
There are three facts that make this computation very efficient.
The first is that we can cheaply choose ${\cal Q}_k\equiv {\cal I}$, the identical mapping on ${\cal S}^n$.
This is because $\cal A\equiv {\cal O}$, which leads to 
\[
 \|\nabla f(X)-\nabla f(Y) \|
= \| {\cal A}^*{\cal A}(X-Y) \|
\leq \frac{1}{2}\|X-Y\| \qquad \forall \ \ X,Y\in{\cal S}^n_h.
\]
For different applications, one may want to choose different forms of ${\cal Q}_{k}$ 
\citep[see][for more details]{JSToh12}.
The second fact is that the resulting problem of \eqref{eq:approx_APG} with ${\cal Q}_k\equiv {\cal I}$
takes the following form:
\begin{equation}\label{eq:NEDM_subproblem} 
\begin{array}{cl}
\min & \frac{1}{2}\|X-(Y_k -\nabla f(Y_{k}))\|^2 \\ [3pt]
{\rm s.t.} & {\cal B}(X)=b,  \ \ X\in{\cal K}_+^n.
\end{array}
\end{equation}
The geometric meaning of problem \eqref{eq:NEDM_subproblem}
is that it is to compute the nearest EDM from the matrix $(Y_k - \nabla f(Y_k))$.
This type of problems 
can be efficiently solved by the semismooth Newton method developed in \cite{Qi13}.

The third fact is that problem \eqref{eq:NEDM_subproblem} is only solved approximately to
meet a sufficient accuracy criterion (see \cite[eq. (32)]{JSToh12} for the detailed formulations).
This leads to a significant speed up of the already very fast semismooth Newton method.
We note that the inexact APG  inherits the computational $1/\sqrt{\varepsilon}$ complexity 
of the exact APG such as the FISTA developed by \cite{BTeboulle09}. 
Other complexity results on our inexact APG algorithm are similar as
 those of \cite[Theorem 3.1]{JSToh12}. We omit the details here to save space.

\section{Numerical Experiments} \label{Section-Numerical}
%%%%%%%%%%%%%%%%%%%%%%%%%%%%%%%%%%

In this section, we demonstrate the effectiveness of the proposed 
EDM Embedding (EDME) model \eqref{eq:estimator_problem} 
by testing Algorithm \ref{alg:APG} on some real world examples. 
The examples are in two categories: one is of the social network visualization problem,
whose initial link observation can be modelled by uniform random graphs.
The other is  from  manifold learning,
whose initial distances are obtained by the k-NN rule.
The known physical features of those problems enable us to evaluate how good EDME is when compared
to other models such as ISOMAP and MVU.
It appears that EDME is capable of generating configurations of very high quality both 
in terms of extracting those physical features and of higher EDM scores.
The test also raises  an open question whether our theoretical results can be extended to this case.

For comparison purpose, we also report the performance of MVU and ISOMAP for most cases. 
The SDP solver used is the state-of-art 
SDPT3 package, which allows us to test problems of large data sets. 
We did not compare with MVE as it solves a sequence of SDPs and consequently it is too slow for our tested 
problems. Details on this and other implementation issues can be found in 
Subsection \ref{Subsection_Numerical_Performance}.

%As mentioned in Section \ref{section:Error bounds}, the initial estimation of the unknown true Euclidean matrix $\overline{D}$ is crucial to our EDME method. In our numerical examples, we find the initial estimators generated by the ISOMAP work well in most of cases. In applications, different methods such as the  MVU and MVE also can be used to generate a suitable initial estimator. 
%%%%%%%%%%%%%%%%%%%%%%%%%%%%%%%%%%%%%%%%
\subsection{Social Network}
%%%%%%%%%%%%%%%

Four real-world networks arising from the different applications are used to 
demonstrate the quality of our new estimator from EDME.\\

\begin{figure}[h]
\centering
%\begin{subfigure}{.5\textwidth}
%  \centering
%  \includegraphics[width=1\linewidth]{Enron_SHP.jpg}
%  \caption{the shortest path}
%  \label{fig:sub1}
%\end{subfigure}%
\begin{subfigure}{.5\textwidth}
  \centering
  \includegraphics[width=1\linewidth]{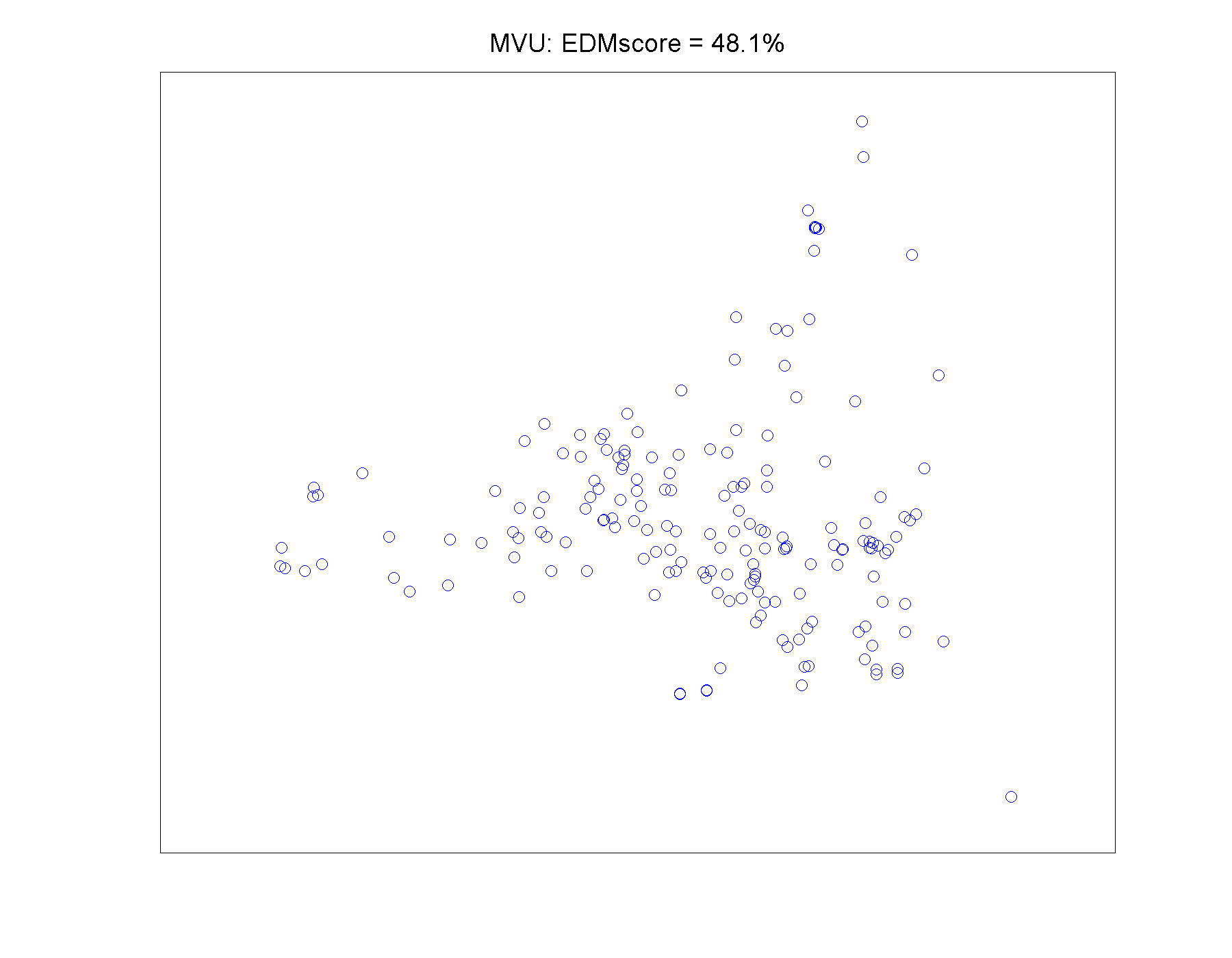}
%  \caption{MVU}
  \label{fig:sub2}
\end{subfigure}%
%\begin{subfigure}{.5\textwidth}
%  \centering
%  \includegraphics[width=1\linewidth]{Enron_MVE.jpg}
%  \caption{MVE}
%  \label{fig:sub1}
%\end{subfigure}%
\begin{subfigure}{.5\textwidth}
  \centering
  \includegraphics[width=1\linewidth]{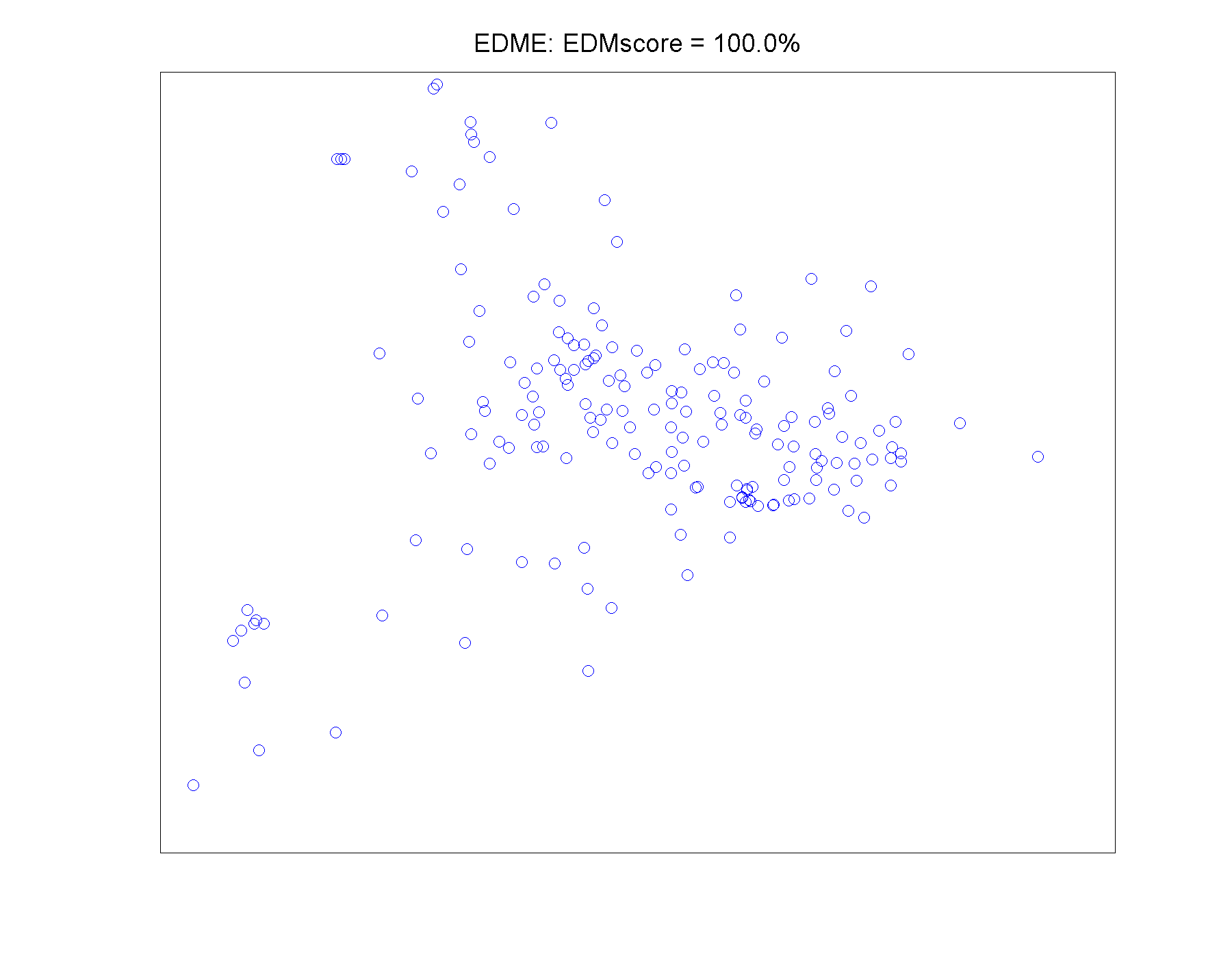}
%  \caption{EDME}
  \label{fig:sub2}
\end{subfigure}
\begin{subfigure}{.5\textwidth}
  \centering
  \includegraphics[width=1\linewidth]{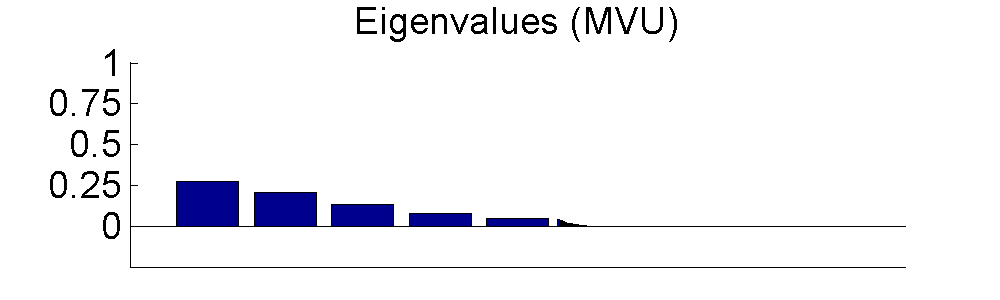}
%  \caption{EDME}
%  \label{fig:sub2}
\end{subfigure}%
\begin{subfigure}{.5\textwidth}
  \centering
  \includegraphics[width=1\linewidth]{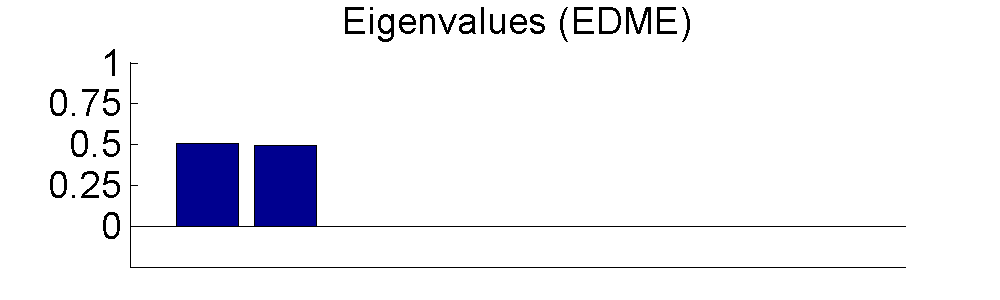}
%  \caption{EDME}
%  \label{fig:sub2}
\end{subfigure}
\caption{The embedding networks of Enron email network.}
\label{fig:Enron_result}
\end{figure}

\begin{figure}[h]
\centering
\begin{subfigure}{.5\textwidth}
  \centering
  \includegraphics[width=1\linewidth]{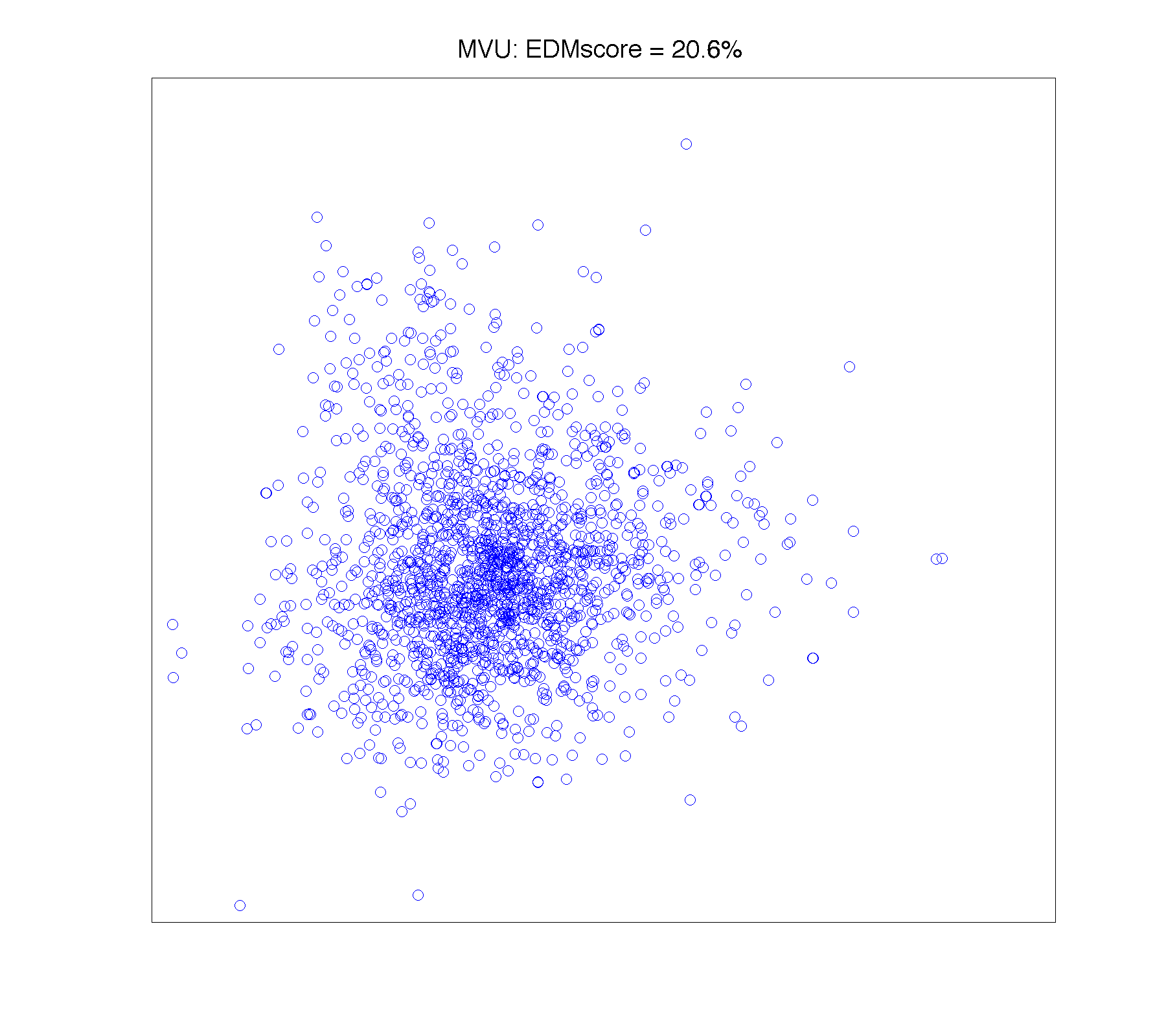}
%  \caption{MVU}
%  \label{fig:sub1}
\end{subfigure}%
\begin{subfigure}{.5\textwidth}
  \centering
  \includegraphics[width=1\linewidth]{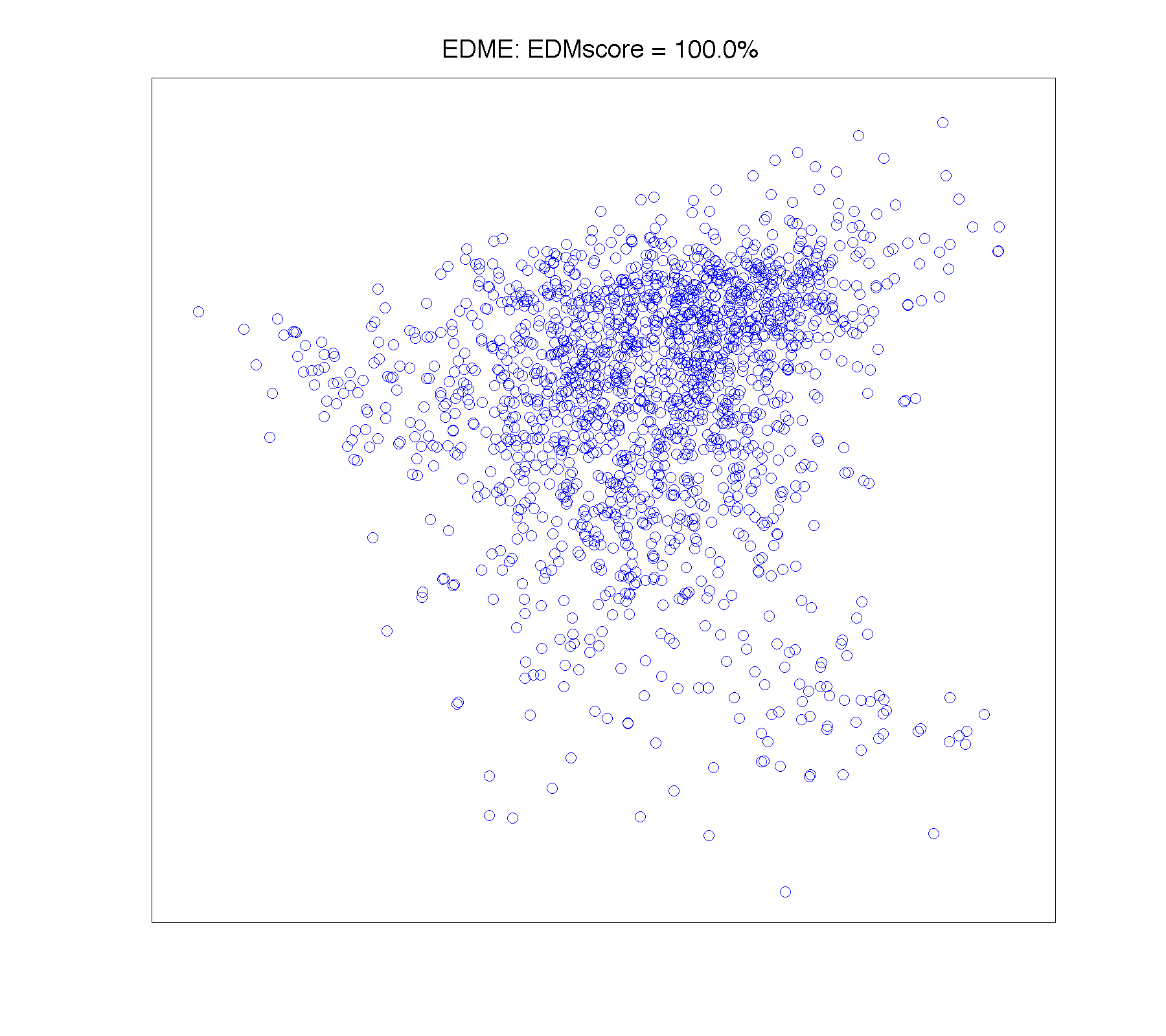}
%  \caption{EDME}
%  \label{fig:sub2}
\end{subfigure}
\begin{subfigure}{.5\textwidth}
  \centering
  \includegraphics[width=1\linewidth]{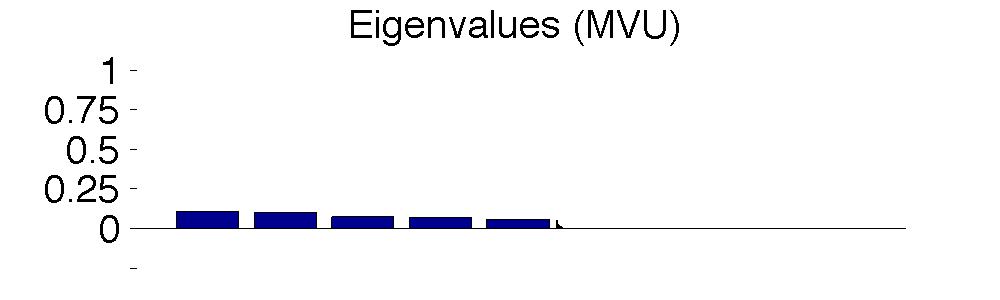}
%  \caption{EDME}
%  \label{fig:sub2}
\end{subfigure}%
\begin{subfigure}{.5\textwidth}
  \centering
  \includegraphics[width=1\linewidth]{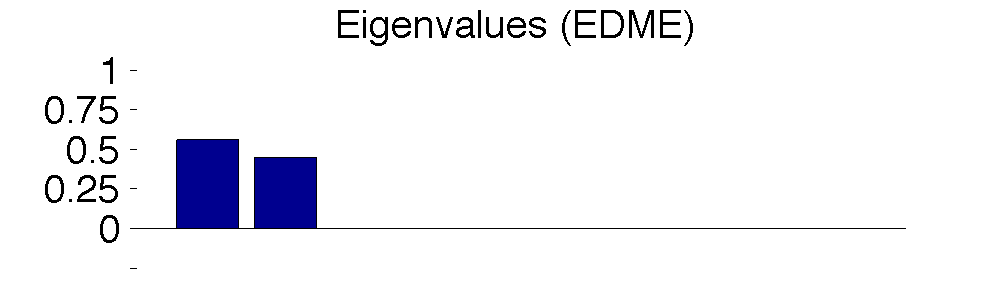}
%  \caption{EDME}
%  \label{fig:sub2}
\end{subfigure}
\caption{The embedding networks of Facebook-like network}
\label{fig:Facebook_result}
\end{figure}

\noindent
{\bf (SN1) Communication networks}. 
Two sets of data are used in this experiment: the Enron email dataset ($n=183$ users, \cite{Enron}) 
and the facebook-like social network ($n=1893$ users, \cite{OpsahlP09}). 
For each dataset, we count the observed communications between the $i$-th and $j$-th users during a fixed time period from the mail/message logs, and denote this quantity by $C_{ij}$. 
The social distances (or dissimilarities) between users are computed from the communication counts. It is natural to assume that larger communication count implies 
smaller social distance. 
Without loss of generality, we employ the widely used Jaccard dissimilarity \citep{KBoyack06}
to measure the social distance of users:
\begin{equation}\label{eq:def-Jaccard}
D_{ij}=\sqrt{1-\frac{C_{ij}}{\sum_{k}C_{ik}+\sum_kC_{jk}-C_{ij}}} \quad \mbox{if $C_{ij}\neq 0$}.
\end{equation}
Thus, the observed measurement matrix $D$ is incomplete, i.e., only a few entrances of the social distance matrix $D$ are observed (the Enron email network: $<13 \%$ and the Facebook-like social network: $<0.8 \%$). The corresponding two dimensional embedding results obtained by the MVU and our EDME
model are shown in Figure \ref{fig:Enron_result} \& \ref{fig:Facebook_result} respectively. 
It can be seen from the presented eigenvalue spectrums\footnote{All eigenvalue spectrums are shown as the fractions of the traces.} of the Gramm matrices $K=-1/2JDJ$ in  Figure \ref{fig:Enron_result} \& \ref{fig:Facebook_result} that our EDME method is able to more accurately represent the high dimensional social network data in two dimensions than the MVU for both examples, since we capture all variance of the data in the top two eigenvectors, i.e., the ${\rm EDMscore}(2)=100\%$. In fact, 
the MVU actually disperses the data into a much higher
 dimensional space (the rank of the Gramm matrix is much higher than the desired dimension) and only a small percentage of the variance can be explained from the top two eigenvectors. The details on the numerical performance on the MVU and EDME on these two examples will be reported in Table \ref{tab:numerical_perf}.\\

%\begin{figure}[H]
%\centering
%%\begin{subfigure}{.5\textwidth}
%%  \centering
%%  \includegraphics[width=1\linewidth]{eigsEnron_SHP.jpg}
%%%  \caption{the shortest path}
%%%  \label{fig:sub1}
%%\end{subfigure}%
%\begin{subfigure}{.5\textwidth}
%  \centering
%  \includegraphics[width=1\linewidth]{eigsEnron_MVU.png}
%%  \caption{MVU}
%%  \label{fig:sub2}
%\end{subfigure}%
%%\begin{subfigure}{.5\textwidth}
%%  \centering
%%  \includegraphics[width=1\linewidth]{eigsEnron_MVE.jpg}
%%%  \caption{MVE}
%%%  \label{fig:sub1}
%%\end{subfigure}%
%\begin{subfigure}{.5\textwidth}
%  \centering
%  \includegraphics[width=1\linewidth]{eigsEnron_EDME.png}
%%  \caption{EDME}
%%  \label{fig:sub2}
%\end{subfigure}
%\caption{Eigenvalue spectra of Enron email network}
%\label{fig:Enron_eigen}
%\end{figure}

%\begin{figure}[H]
%\centering
%\begin{subfigure}{.5\textwidth}
%  \centering
%  \includegraphics[width=1\linewidth]{eigsFacebook_MVU.png}
%%  \caption{the shortest path}
%%  \label{fig:sub1}
%\end{subfigure}%
%\begin{subfigure}{.5\textwidth}
%  \centering
%  \includegraphics[width=1\linewidth]{eigsFacebook_EDME.png}
%%  \caption{MVU}
%%  \label{fig:sub2}
%\end{subfigure}
%\caption{Eigenvalue spectra of Facebook-like network}
%\label{fig:Facebook_eigen}
%\end{figure}

\begin{figure}[ht]
\centering
\begin{subfigure}{.5\textwidth}
  \centering
  \includegraphics[width=1\linewidth]{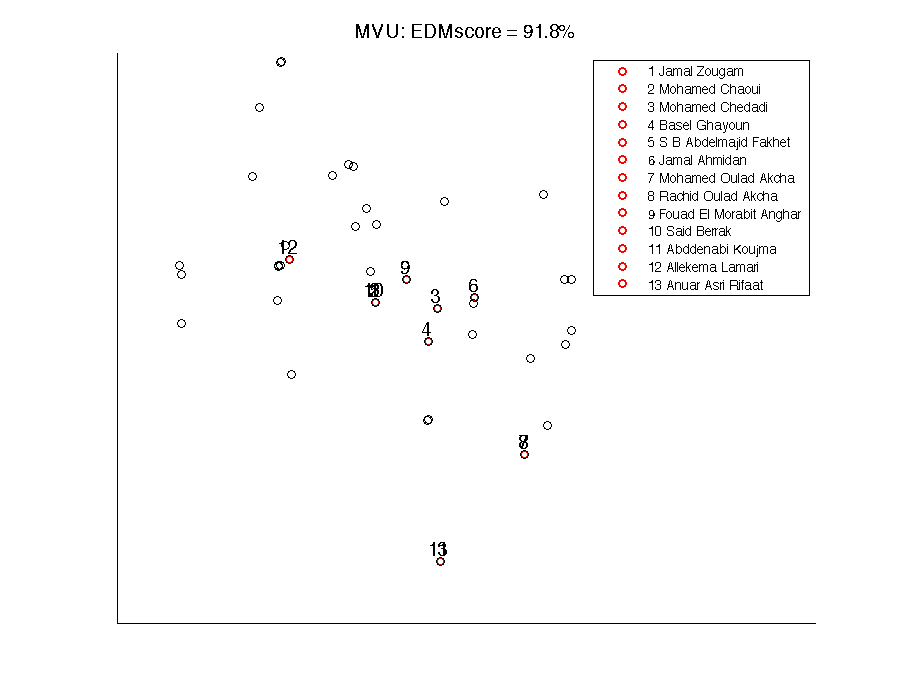}
%  \caption{MVU}
  \label{fig:sub2}
\end{subfigure}%
%\begin{subfigure}{.5\textwidth}
%  \centering
%  \includegraphics[width=1\linewidth]{train_MVE.png}
%  \caption{MVE}
%  \label{fig:sub1}
%\end{subfigure}%
%\begin{subfigure}{.5\textwidth}
%  \centering
%  \includegraphics[width=1\linewidth]{train_SHP.png}
%%  \caption{the shortest path}
%  \label{fig:sub1}
%\end{subfigure}
\begin{subfigure}{.5\textwidth}
  \centering
  \includegraphics[width=1\linewidth]{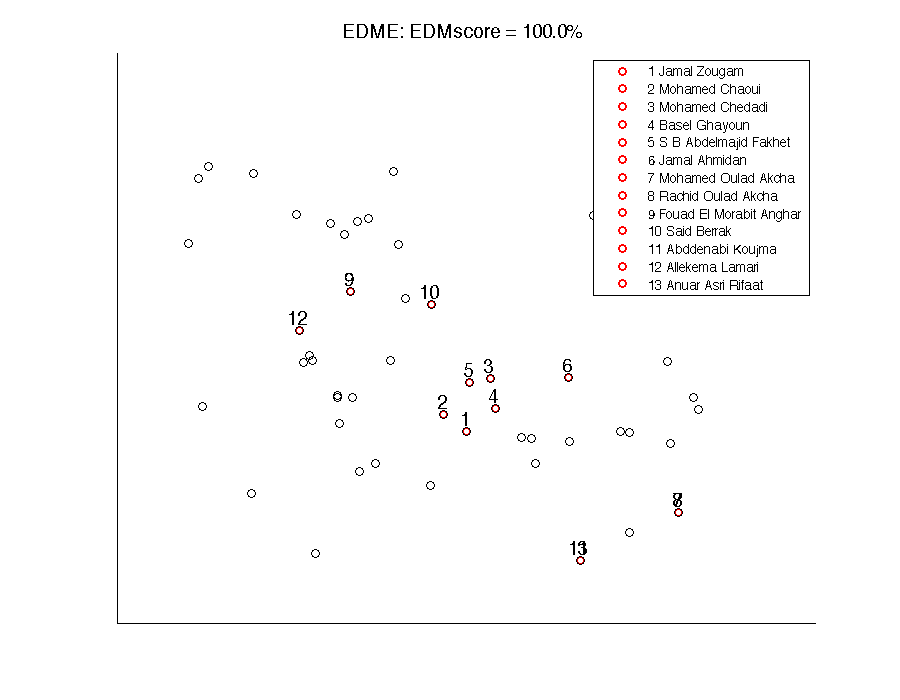}
%  \caption{EDME}
  \label{fig:sub2}
\end{subfigure}
\begin{subfigure}{.5\textwidth}
  \centering
  \includegraphics[width=1\linewidth]{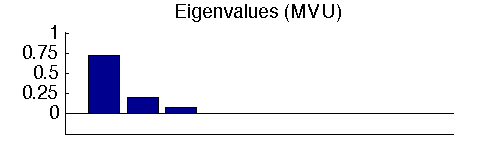}
%  \caption{MVU}
%  \label{fig:sub2}
\end{subfigure}%
\begin{subfigure}{.5\textwidth}
  \centering
  \includegraphics[width=1\linewidth]{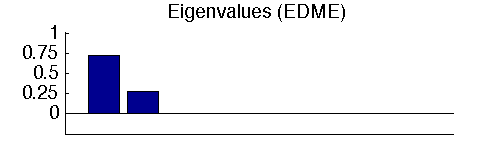}
%  \caption{MVU}
%  \label{fig:sub2}
\end{subfigure}
\caption{The embedding networks of Madrid train bombing.}
\label{fig:Train_result}
\end{figure}

\noindent
{\bf (SN2) Madrid train bombing}.
In this example, we try to visualize the social network of the individuals involved in the bombing of commuter trains in Madrid on March 11, 2004 from the data obtained by Rodriguez \cite{Brian06} 
(downloaded from \cite{Freemandata}). Partial information of the connection strength between different individuals 
is recorded based on certain rule (see \cite{Freemandata} for details).  
The visualizations of the social networks\footnote{Six persons are removed since they are isolated nodes, without any documented links.} obtained by the MVU and EDME are presented in Figure \ref{fig:Train_result}.
 The ``field operations group'', which includes those who actually placed the explosives, are also indicated by the numbered red circles on both figures. 
Both the MVU and EDME obtained similar social network structures of the individuals.
For example, person $7$ and person $8$ (they turn out to be brothers)
 are placed nearly at the same spot in both embeddings.
 However, the EDME captures all  variance of the data in the two leading eigenvectors
 with higher EDMscore, which is demonstrated by the eigenvalue spectrums.\\

%\begin{figure}[H]
%\centering
%\begin{subfigure}{.5\textwidth}
%  \centering
%  \includegraphics[width=1\linewidth]{eigsTrain_MVU.png}
%%  \caption{MVU}
%%  \label{fig:sub2}
%\end{subfigure}%
%%\begin{subfigure}{.5\textwidth}
%%  \centering
%%  \includegraphics[width=1\linewidth]{eigsTrain_SHP.png}
%%%  \caption{the shortest path}
%%%  \label{fig:sub1}
%%\end{subfigure}
%%\begin{subfigure}{.5\textwidth}
%%  \centering
%%  \includegraphics[width=1\linewidth]{eigsTrain_MVE.png}
%%%  \caption{MVE}
%%%  \label{fig:sub1}
%%\end{subfigure}%
%\begin{subfigure}{.5\textwidth}
%  \centering
%  \includegraphics[width=1\linewidth]{eigsTrain_EDME.png}
%%  \caption{EDME}
%%  \label{fig:sub2}
%\end{subfigure}
%\caption{Eigenvalue spectra of Madrid train bombing}
%\label{fig:Train_eigen}
%\end{figure}

\begin{figure}[p]
\centering
%\begin{subfigure}{.5\textwidth}
%  \centering
%  \includegraphics[width=1\linewidth]{USairport_SHP.png}
%  \caption{the shortest path}
%%  \label{fig:sub1}
%\end{subfigure}%
\begin{subfigure}{.5\textwidth}
  \centering
  \includegraphics[width=1\linewidth]{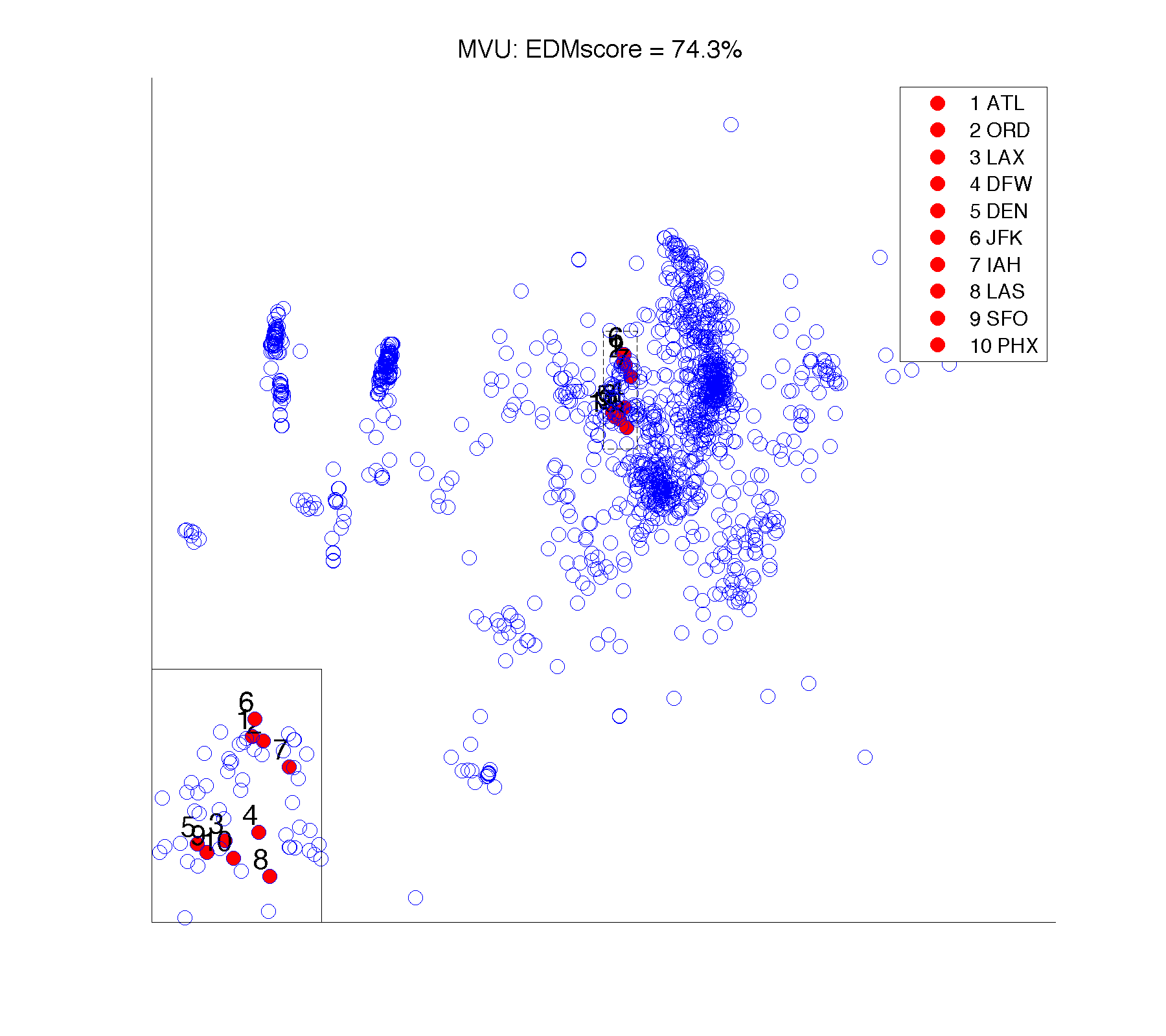}
%  \caption{MVU}
%  \label{fig:sub1}
\end{subfigure}%
\begin{subfigure}{.5\textwidth}
  \centering
  \includegraphics[width=1\linewidth]{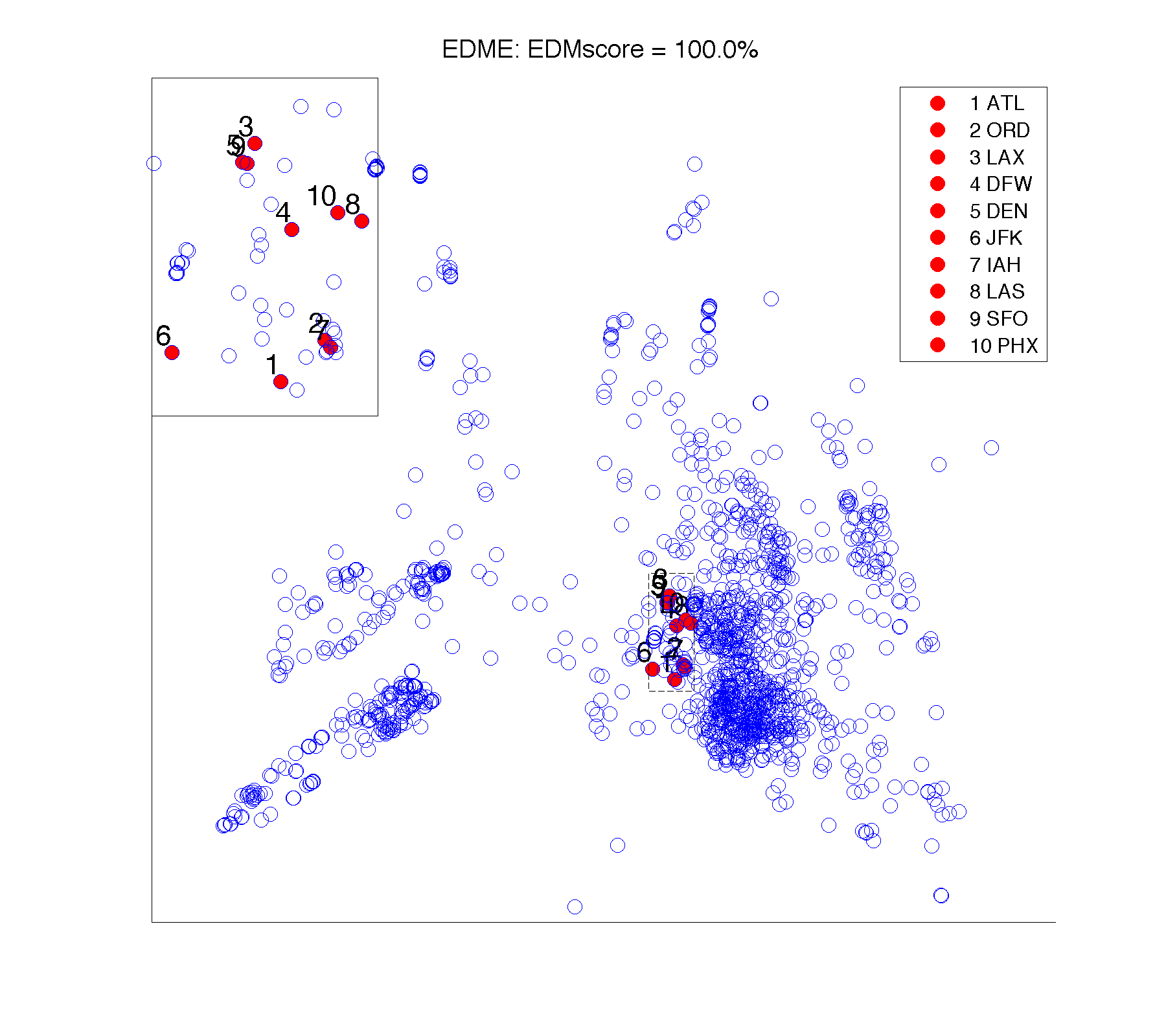}
%  \caption{EDME}
%  \label{fig:sub2}
\end{subfigure}
\begin{subfigure}{.5\textwidth}
  \centering
  \includegraphics[width=1\linewidth]{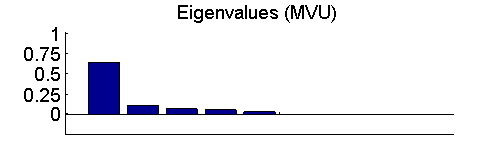}
%  \caption{MVU}
%  \label{fig:sub2}
\end{subfigure}%
\begin{subfigure}{.5\textwidth}
  \centering
  \includegraphics[width=1\linewidth]{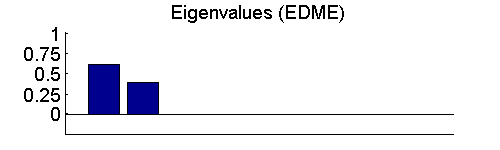}
%  \caption{MVU}
%  \label{fig:sub2}
\end{subfigure}
\caption{The embedding networks of USairport2010}
\label{fig:USairport2010_result}
\end{figure}

\noindent
{\bf (SN3) US airport network}.
In this example, we try to visualize the social network of the US airport network 
from the data of 2010 \cite{Opsahl}. There are $n=1572$ airports under consideration.  
The number of the passengers transported from the $i$-th airport to the $j$-th airport in 2010 
is recorded and denoted by $C_{ij}$. 
Therefore, the social distance between two cities can be measured by the passenger numbers. 
For simplicity, we use the Jaccard dissimilarity defined in \eqref{eq:def-Jaccard} to compute the corresponding distances between two cities. The observed distance matrix is also incomplete, and only very few entrances are observed ($<1.4\%$). The two dimensional embeddings obtained by the MVU and EDME methods are shown in Figure \ref{fig:USairport2010_result}. The ten busiest US airports by total passenger traffic in 2010 are indicated by the red circles. Note that there are a large number of passengers transporting between them, which means the corresponding social distances among them should be relatively small. 
Thus,  it is reasonable to expect that the embedding points of these top ten airports cluster 
around the zero point. Both MVU and EDME methods are able to show this important feature. 
However, it can be seen from the eigenvalue spectrums in Figure \ref{fig:USairport2010_result} 
that the MVU only captured $74.3\%$ variance in the top two leading eigenvectors, 
while the EDME method captured all the variance in the two dimensional space. \\

%\begin{figure}[H]
%\centering
%%\begin{subfigure}{.5\textwidth}
%%  \centering
%%  \includegraphics[width=1\linewidth]{eigsUSairport_SHP.png}
%%%  \caption{the shortest path}
%%%  \label{fig:sub1}
%%\end{subfigure}%
%\begin{subfigure}{.5\textwidth}
%  \centering
%  \includegraphics[width=1\linewidth]{eigsUSairport_MVU.png}
%%  \caption{MVU}
%%  \label{fig:sub2}
%\end{subfigure}%
%\begin{subfigure}{.5\textwidth}
%  \centering
%  \includegraphics[width=1\linewidth]{eigsUSairport_EDME.png}
%%  \caption{MVU}
%%  \label{fig:sub2}
%\end{subfigure}
%\caption{Eigenvalue spectra of USairport2010}
%\label{fig:USairport2010_eigen}
%\end{figure}

\begin{figure}[p]
\centering
\begin{subfigure}{.5\textwidth}
  \centering
  \includegraphics[width=1\linewidth]{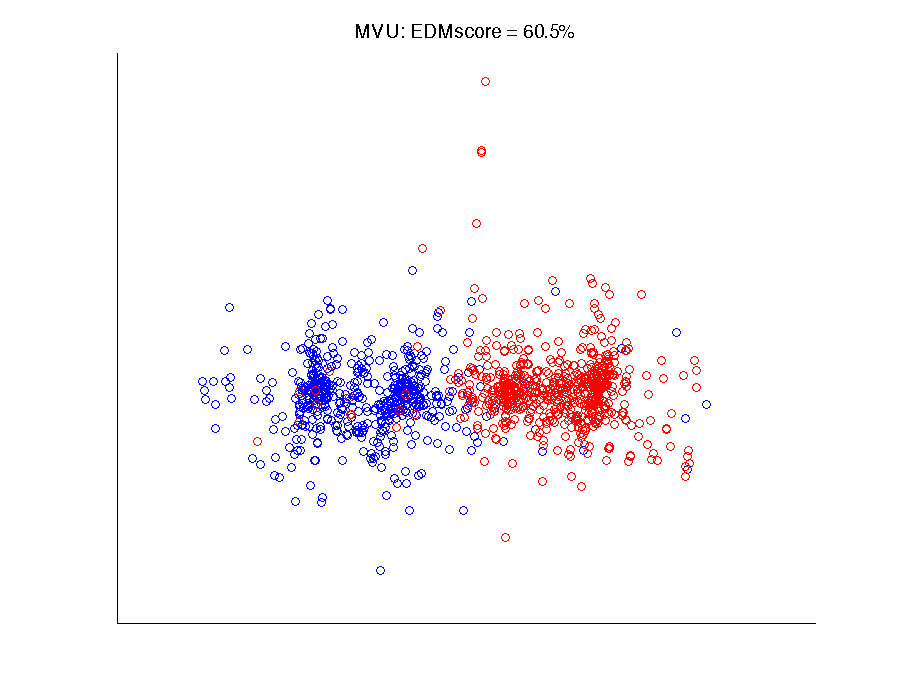}
%  \caption{MVU}
%  \label{fig:sub1}
\end{subfigure}%
%\begin{subfigure}{.5\textwidth}
%  \centering
%  \includegraphics[width=1\linewidth]{Blogs_SHP.png}
%%  \caption{SHP}
%%  \label{fig:sub1}
%\end{subfigure}
\begin{subfigure}{.5\textwidth}
  \centering
  \includegraphics[width=1\linewidth]{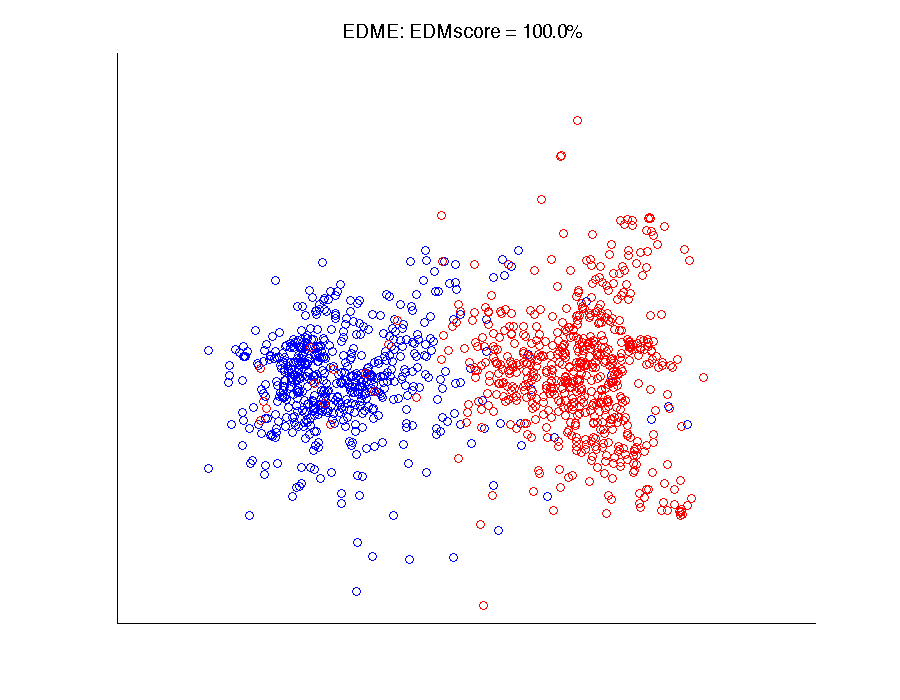}
%  \caption{EDME}
%  \label{fig:sub1}
\end{subfigure}
\begin{subfigure}{.5\textwidth}
  \centering
  \includegraphics[width=1\linewidth]{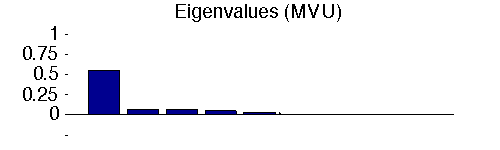}
%  \caption{MVU}
%  \label{fig:sub2}
\end{subfigure}%
\begin{subfigure}{.5\textwidth}
  \centering
  \includegraphics[width=1\linewidth]{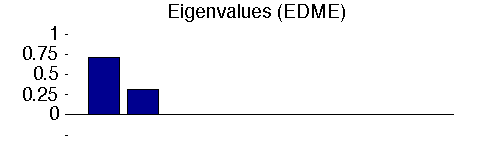}
%  \caption{MVU}
%  \label{fig:sub2}
\end{subfigure}
\caption{The political blogs}
\label{fig:Blogs_result}
\end{figure}

\noindent
{\bf (SN4) Political blogs}
\cite{AGlance05} collected the data including links, citations and posts on the 1940 political blogs around the 2004 US presidential election period. These blogs are classified as two parts: 758 left-leaning blogs and 732 right-leaning blogs.  In this paper, we will use the data on the links between the blogs, which can be found from \cite{Freemandata} to visualize the corresponding social network. Similar to the communication network, we use the widely used Jaccard dissimilarity defined in \eqref{eq:def-Jaccard} to measure the social distance of blogs. Without loss of generality, the 718 isolated blogs are removed from the original data, which means that we consider the remaining $n=1222$ blogs with 586 left-leanings and 636 right-leanings. The social networks  obtained by the MVU and the EDME are presented in Figure \ref{fig:Blogs_result}. From the results, we can see clearly that the embedding points generated by the MVU are concentrated near the zero point, and the rank of the corresponding Gram matrix is much higher than 2, which  is 1135. However, our EDME method is able to capture all variance of the data in the two dimensions, providing a more accurate lower dimensional embedding. In fact, the embedding points in the visualizing network obtained by the EDME are naturally separated into two groups: the left-leaning blogs (the blue circles) and the right-leaning ones (the red circles).

%\begin{figure}[H]
%\centering
%\begin{subfigure}{.5\textwidth}
%  \centering
%  \includegraphics[width=1\linewidth]{eigsBlogs_MVU.png}
%%  \caption{MVU}
%%  \label{fig:sub2}
%\end{subfigure}%
%%\begin{subfigure}{.5\textwidth}
%%  \centering
%%  \includegraphics[width=1\linewidth]{eigsBlogs_SHP.png}
%%%  \caption{the shortest path}
%%%  \label{fig:sub1}
%%\end{subfigure}
%\begin{subfigure}{.5\textwidth}
%  \centering
%  \includegraphics[width=1\linewidth]{eigsBlogs_EDME.png}
%%  \caption{MVU}
%%  \label{fig:sub2}
%\end{subfigure}
%\caption{Eigenvalue spectra of Blogs}
%\label{fig:USairport2010_eigen}
%\end{figure}

%%%%%%%%%%%%%%%%%%%%%%%%%%%%%%%%%%%%%%%%%%%%%%%%%%
\subsection{Manifold learning}
%%%%%%%%%%%%%%%%%%%%%%%%%%

In this subsection, we test $4$ widely used data sets in manifold learning. 
The initial distances used are generated by the k-NN rule.
We describe them below with our 
findings.

\begin{figure}[h]
\centering
\begin{subfigure}{.5\textwidth}
  \centering
  \includegraphics[width=1\linewidth]{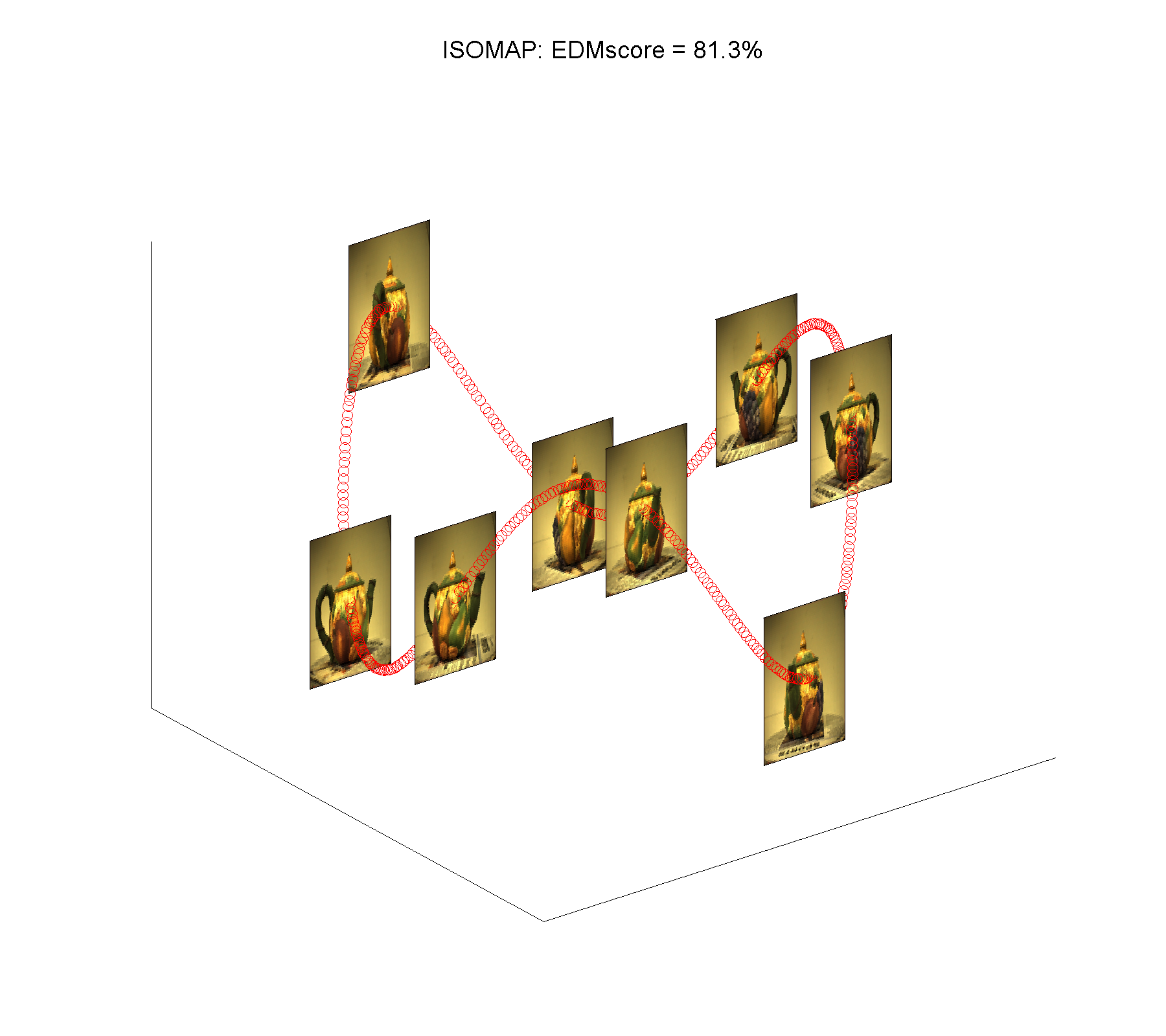}
%  \caption{PCA}
%  \label{fig:sub1}
\end{subfigure}%
\begin{subfigure}{.5\textwidth}
  \centering
  \includegraphics[width=1\linewidth]{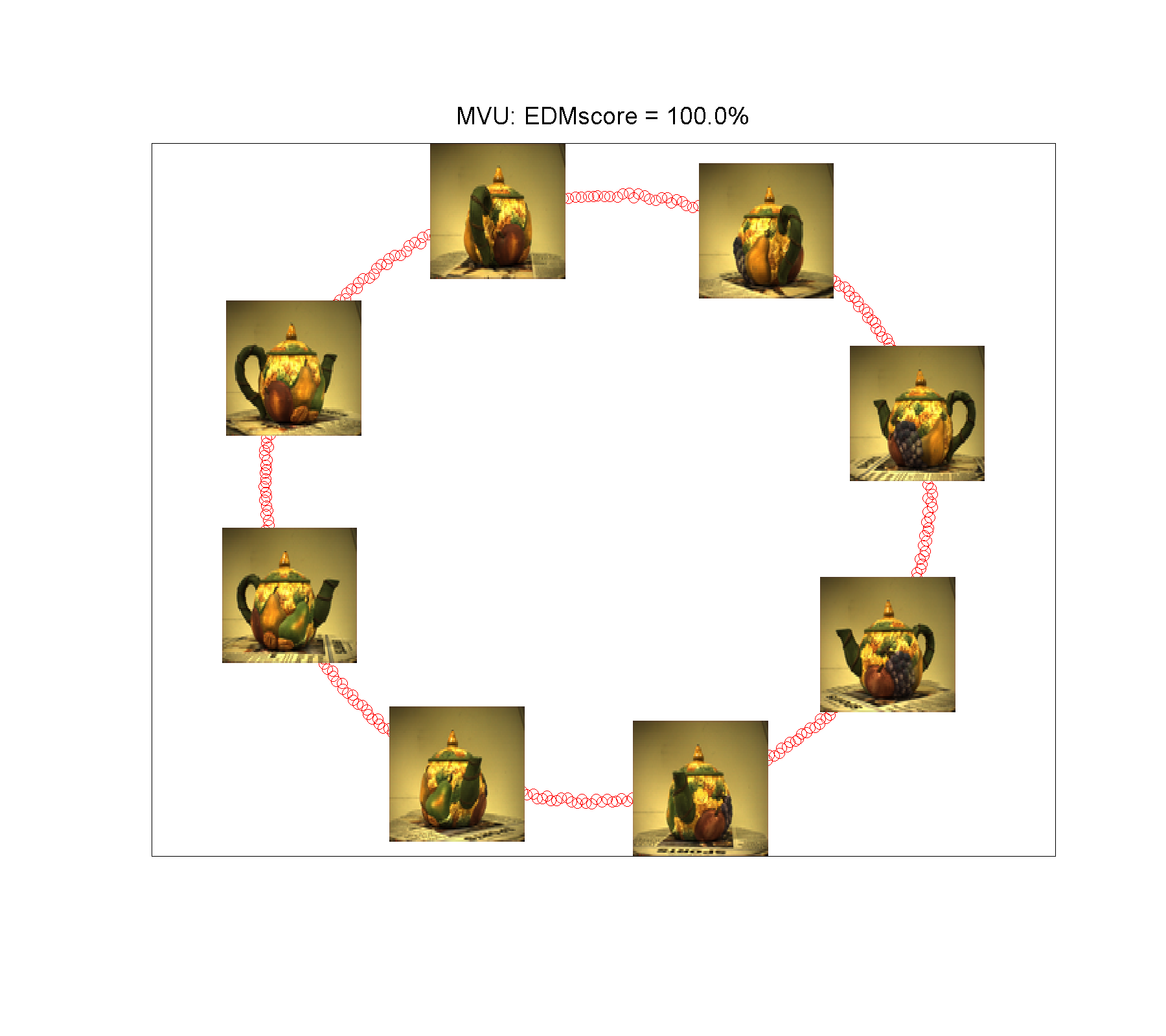}
%  \caption{MVU}
%  \label{fig:sub1}
\end{subfigure}
\begin{subfigure}{.5\textwidth}
  \centering
  \includegraphics[width=1\linewidth]{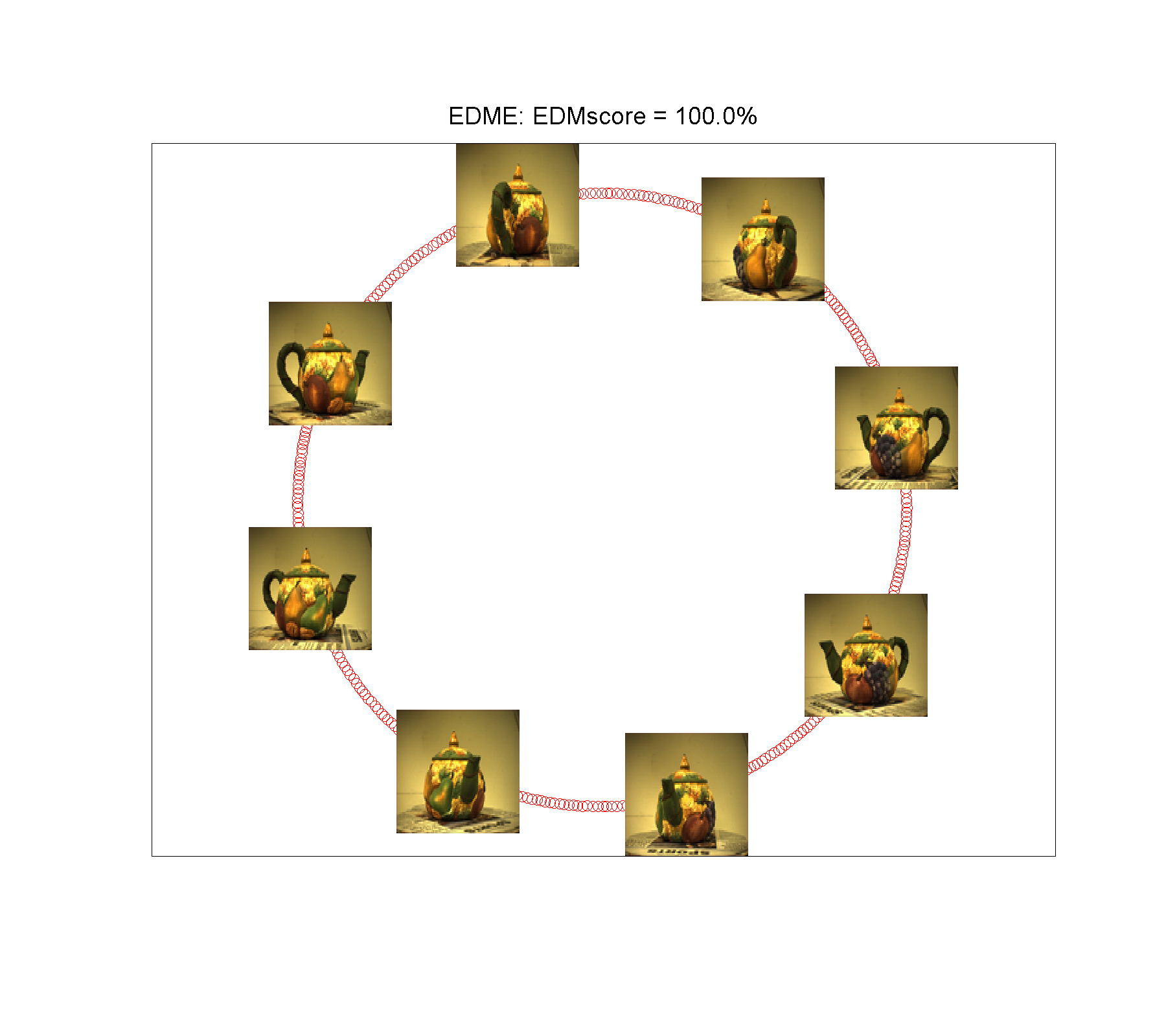}
%  \caption{EDME}
%  \label{fig:sub1}
\end{subfigure}%
\begin{subfigure}{.5\textwidth}
  \centering
  \includegraphics[width=1\linewidth]{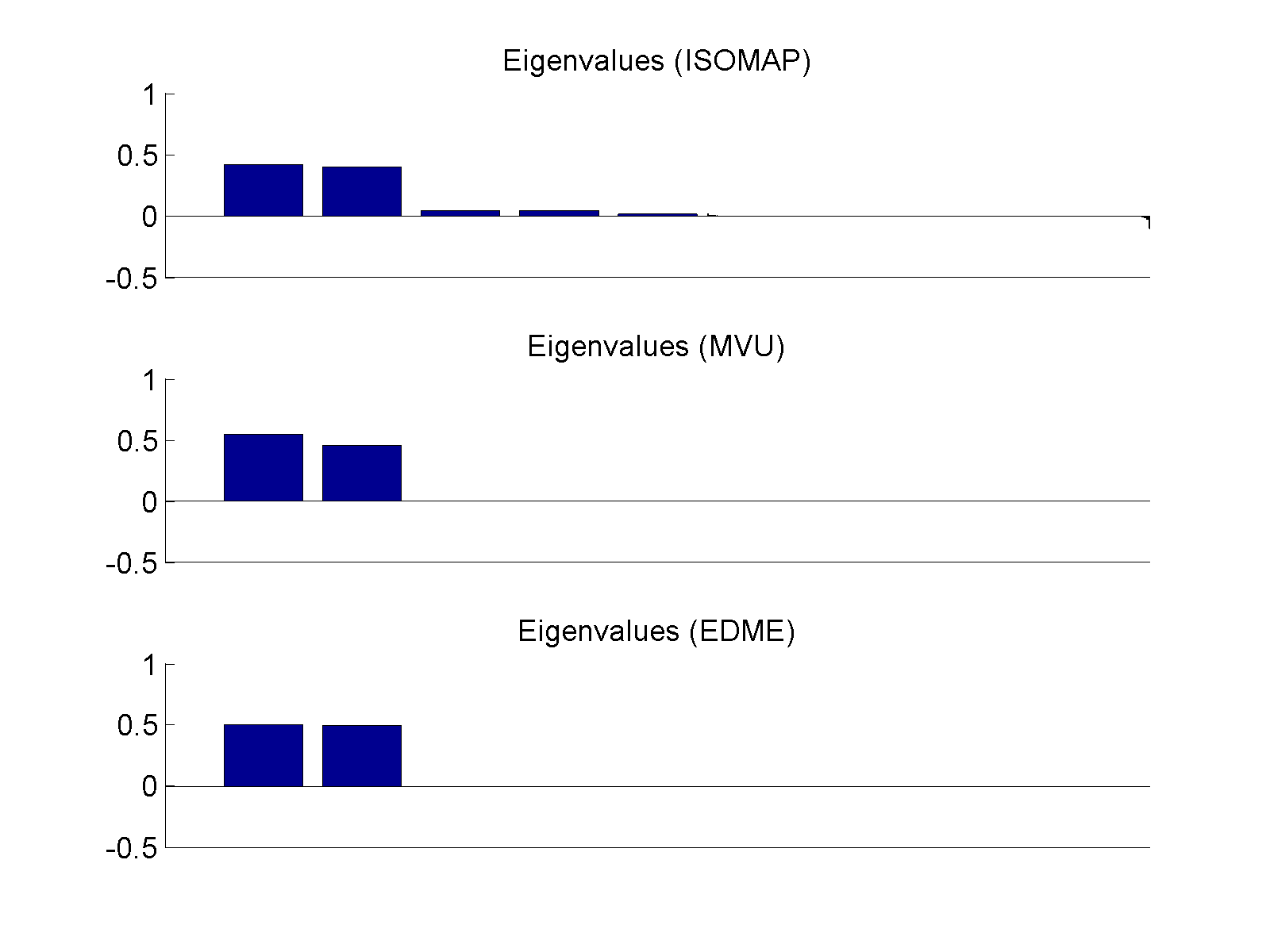}
%  \caption{EDME}
%  \label{fig:sub1}
\end{subfigure}
\caption{Teapots}
\label{fig:Teapots400_result}
\end{figure}

\noindent
{\bf (ML1) Teapots data}
In this example, we consider the two dimensional embedding of the images of a teapot \citep{WS06}. The teapots are rotated 360 degrees, and $n=400$ images are taken from different angles. Each image has $76\times 101$ pixels with 3 byte color depth. After generating a connected graph by $k=5$ nearest neighbors, the ISOMAP, the MVU and the EDME methods are tested. Both MVU and EDME methods are able to accurately represent the rotating object as a circle, which are shown in Figure \ref{fig:Teapots400_result}. In particular, the eigenvalue spectrums of the embedding Gram matrices learned by the MVU and EDME methods indicate that both methods are able to capture all variance of the data in the first two eigenvectors. However, as mentioned by \cite{WS06}, the ISOMAP returns more than two nonzero eigenvalues, which leads to the artificial wave in the third dimension. \\ 

%\begin{figure}[H]
%\centering
%\includegraphics[width=1\linewidth]{eigsTeapots400.png}
%\caption{Eigenvalue spectra of the Teapots400}
%\label{fig:Teapots400_eigen}
%\end{figure}

\begin{figure}[h]
\centering
%\begin{subfigure}{.5\textwidth}
%  \centering
%  \includegraphics[width=1\linewidth]{Face698_SHP.png}
%%  \caption{SHP}
%%  \label{fig:sub1}
%\end{subfigure}%
%\begin{subfigure}{.5\textwidth}
%  \centering
%  \includegraphics[width=1\linewidth]{Face698_MVU.png}
%%  \caption{MVU}
%%  \label{fig:sub1}
%\end{subfigure}%
\begin{subfigure}{.5\textwidth}
  \centering
  \includegraphics[width=1\linewidth]{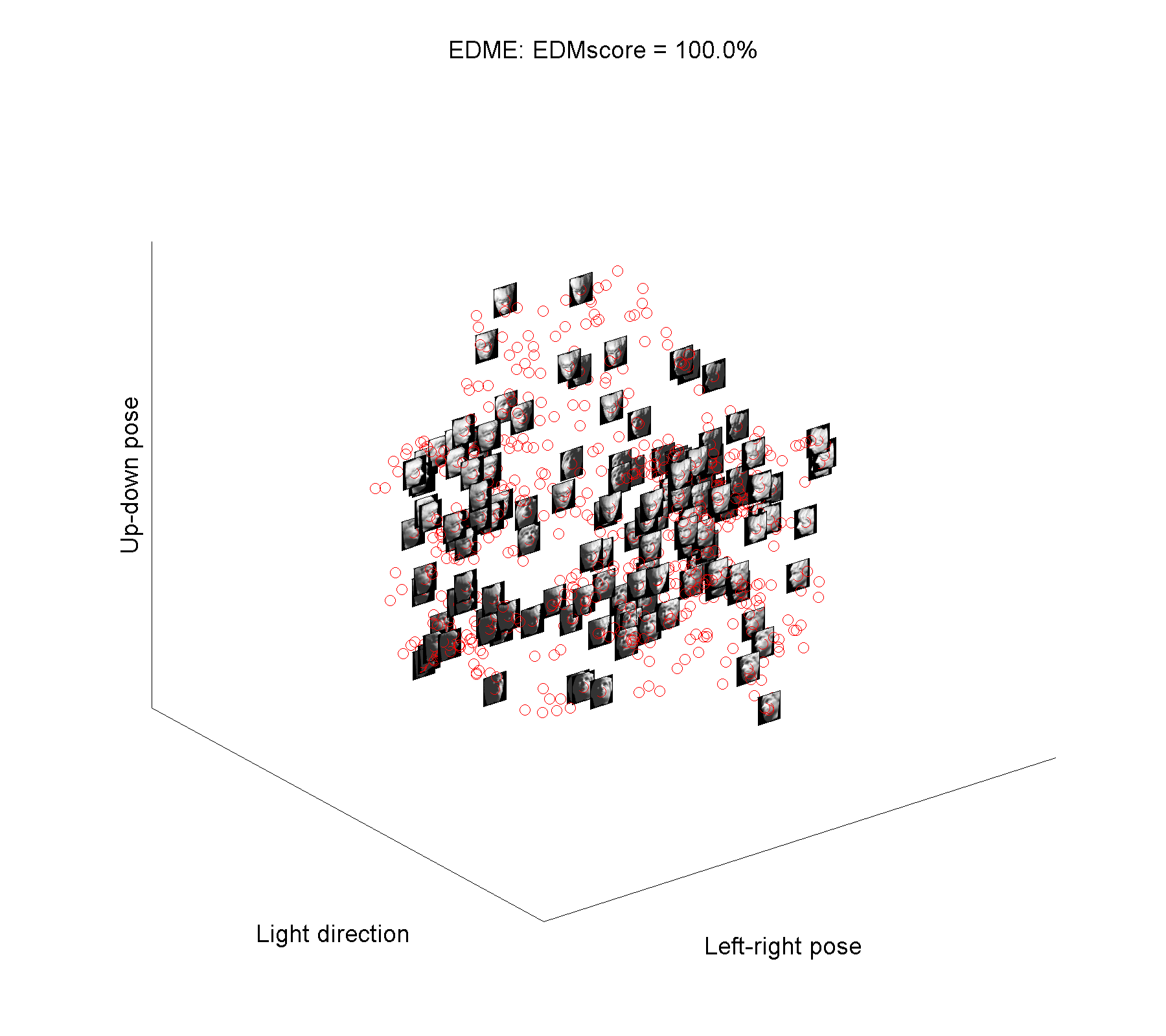}
%  \caption{EDME}
%  \label{fig:sub1}
\end{subfigure}%
\begin{subfigure}{.5\textwidth}
  \centering
  \includegraphics[width=1\linewidth]{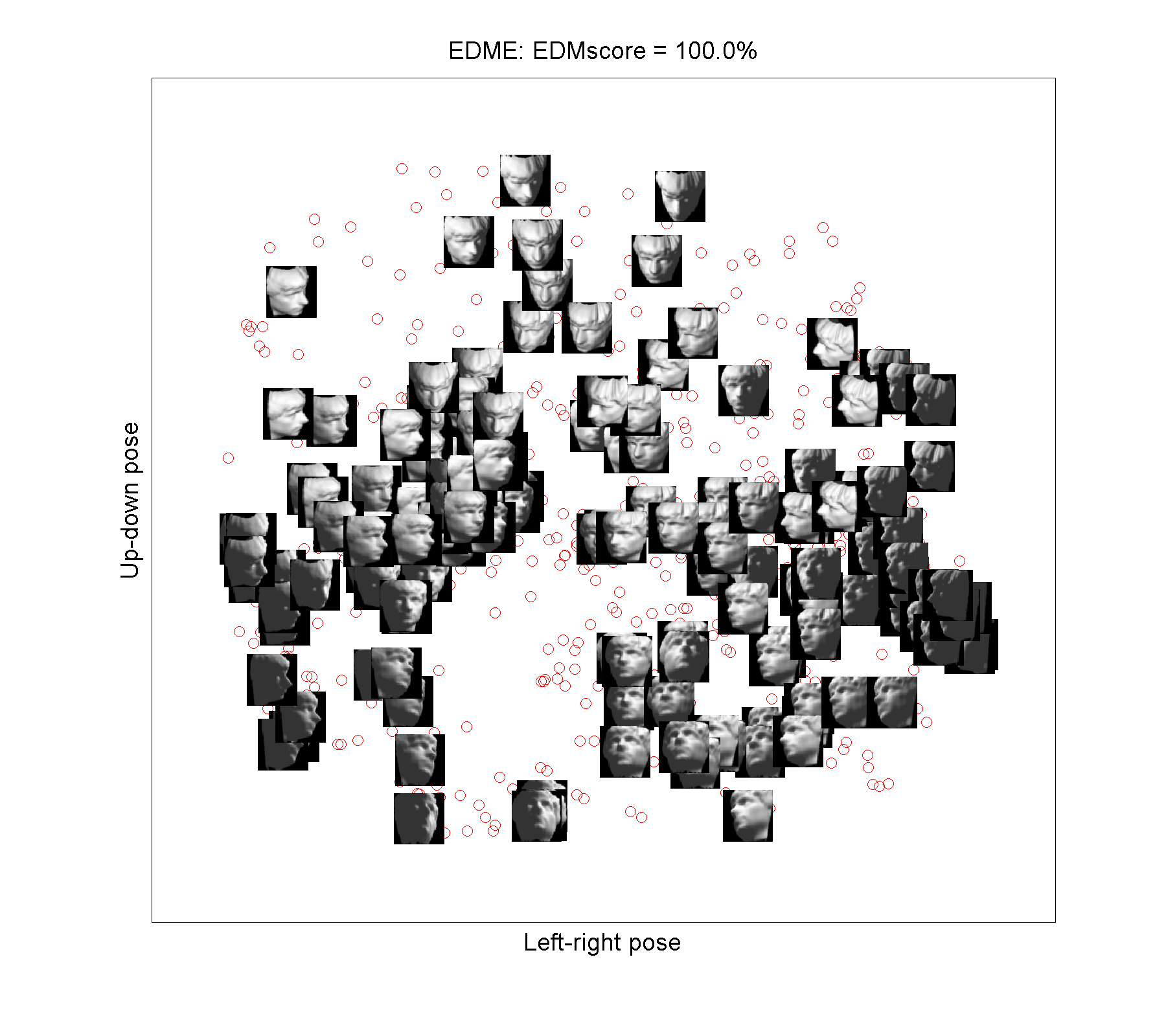}
%  \caption{EDME}
%  \label{fig:sub1}
\end{subfigure}
\begin{subfigure}{.5\textwidth}
  \centering
  \includegraphics[width=1\linewidth]{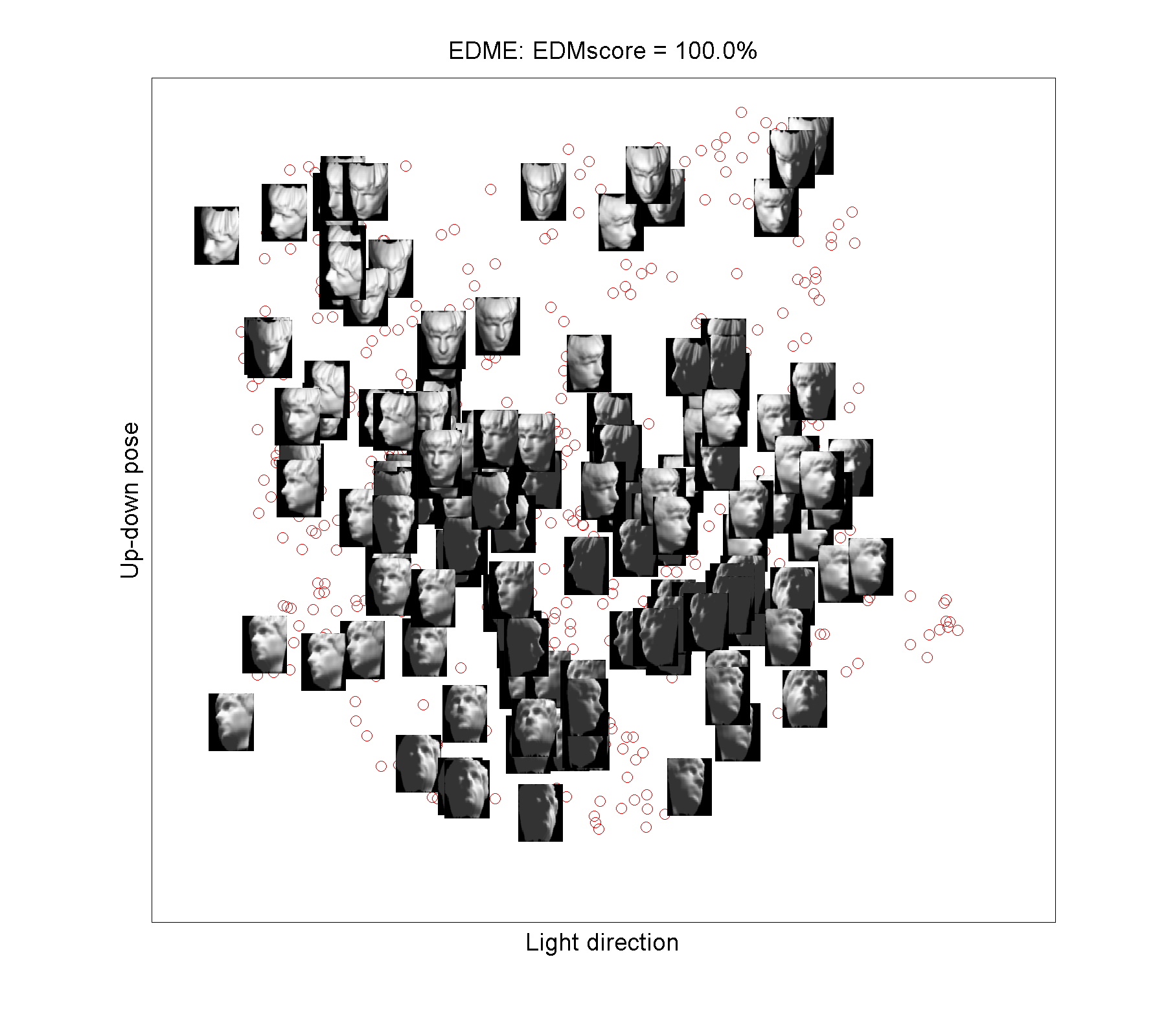}
%  \caption{EDME}
%  \label{fig:sub1}
\end{subfigure}%
\begin{subfigure}{.5\textwidth}
  \centering
  \includegraphics[width=1\linewidth]{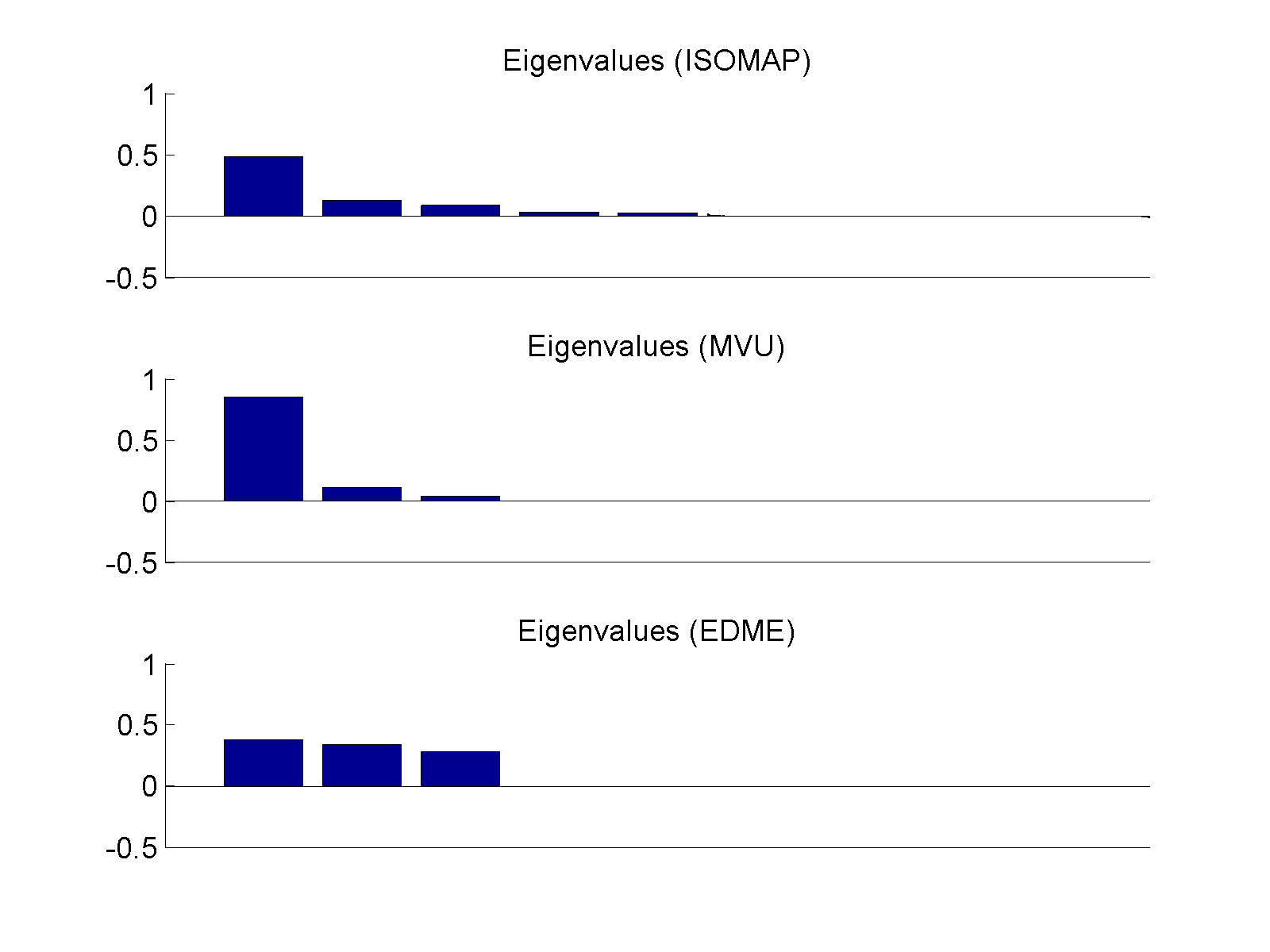}
%  \caption{EDME}
%  \label{fig:sub1}
\end{subfigure}
\caption{Face698}
\label{fig:Face698_result}
\end{figure}

\noindent
{\bf (ML2) Data of Face698}.
In this example, we try to represent the high dimensional face image data \citep{TdL00}
in a low dimension space. There are $n=698$ images (64 pixel by 64 pixel) of faces with the different poses (up-down and left-right) and different light directions. 
Therefore, it is natural to expect that these high dimensional input data 
%(the 4096 dimensional vectors) 
lie in the three dimensional space parameterized by the face poses and the light directions
and that the equal importance of the three features can be sufficiently captured. 
Similar to the previous example, we use $k=5$ to generate a connected graph. 
Both MVU and EDME methods successfully represent the data in the desired three dimensional space
and their embedding results of the MVU and EDME are similar. 
For simplicity only the result of the EDME is shown in Figure \ref{fig:Face698_result}. However, the Gram matrix learned by the ISOMAP has more than three nonzero eigenvalues. This is shown in the corresponding eigenvalue spectrums in Figure \ref{fig:Face698_result}. 
Furthermore, for the ISOMAP, if we only compute the two-dimension embedding, 
then we only capture a smaller percentage of the total variance. 
It is interesting to observe that EDME is the only model that treats the three features equally
important (the three leading eigenvalues are roughly equal).
Moreovre, the EDME model performs much better than MVU in terms of the numerical efficiency. 
See Table \ref{tab:numerical_perf} for more details.\\

%\begin{figure}[H]
%\centering
%  \includegraphics[width=1\linewidth]{eigsFace698.png}
%\caption{Eigenvalue spectra of Face698}
%\label{fig:Face698_eigen}
%\end{figure}

\begin{figure}[h]
\centering
\begin{subfigure}{.5\textwidth}
  \centering
  \includegraphics[width=1\linewidth]{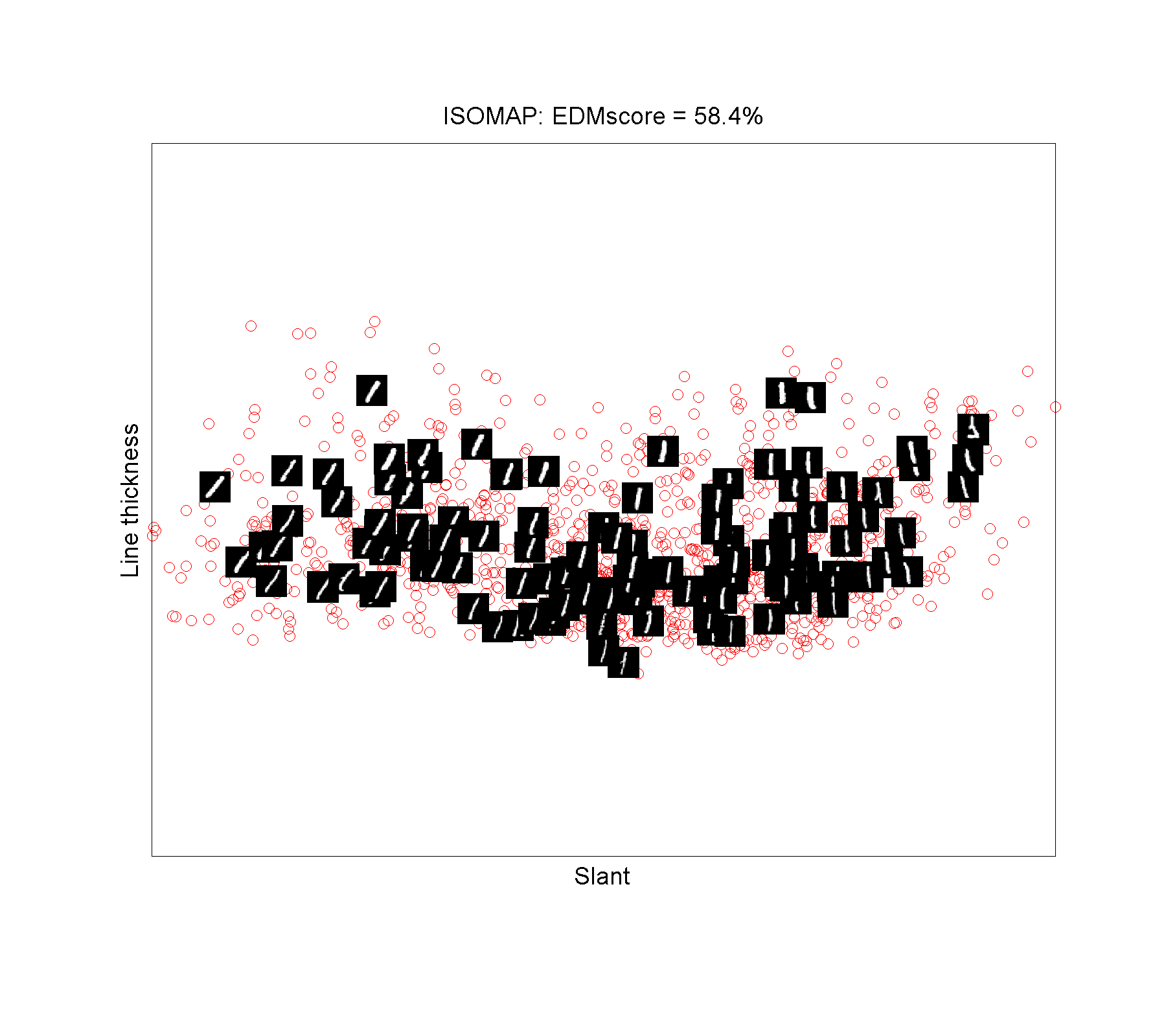}
%  \caption{PCA}
%  \label{fig:sub1}
\end{subfigure}%
\begin{subfigure}{.5\textwidth}
  \centering
  \includegraphics[width=1\linewidth]{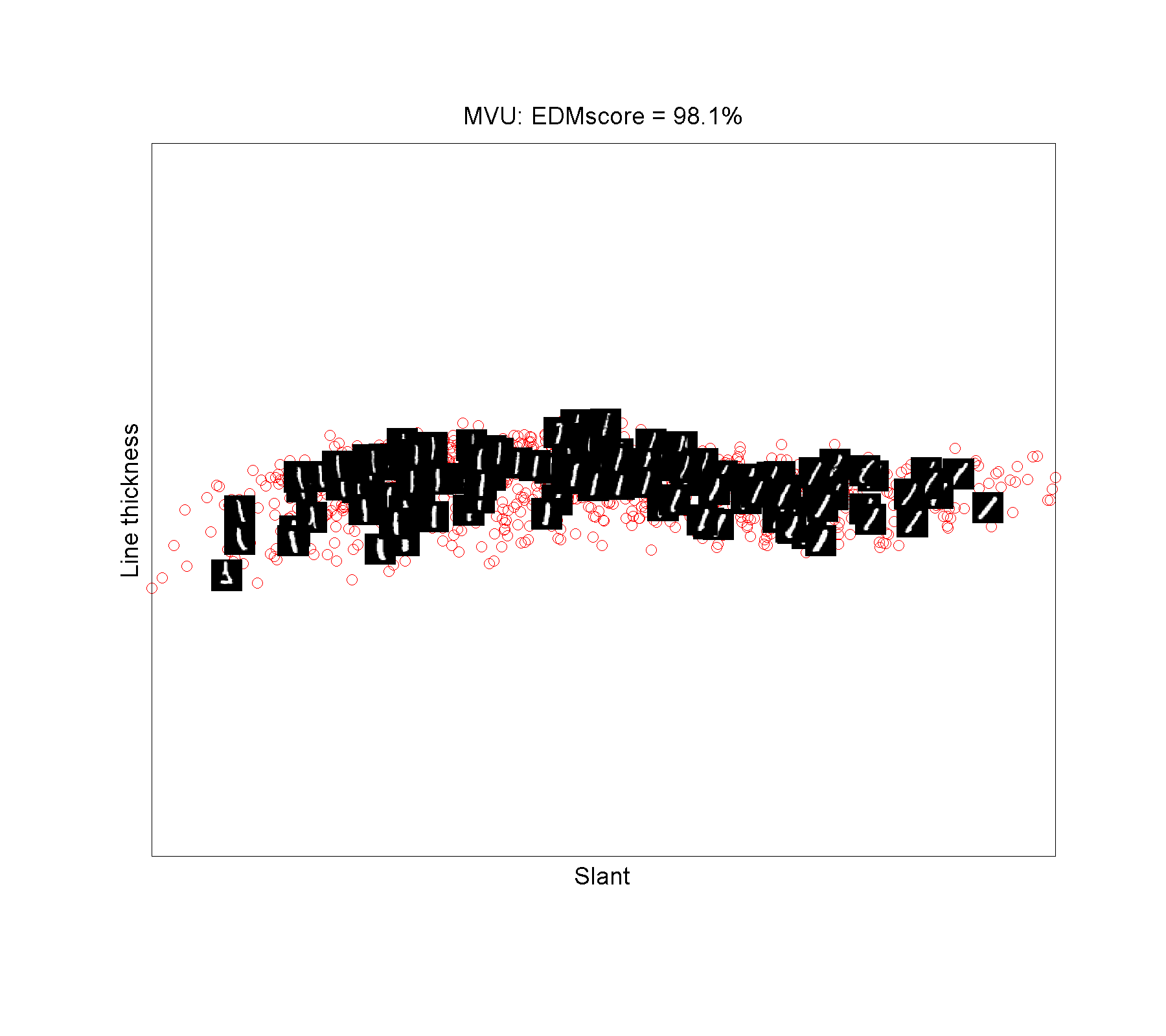}
%  \caption{PCA}
%  \label{fig:sub1}
\end{subfigure}
\begin{subfigure}{.5\textwidth}
  \centering
  \includegraphics[width=1\linewidth]{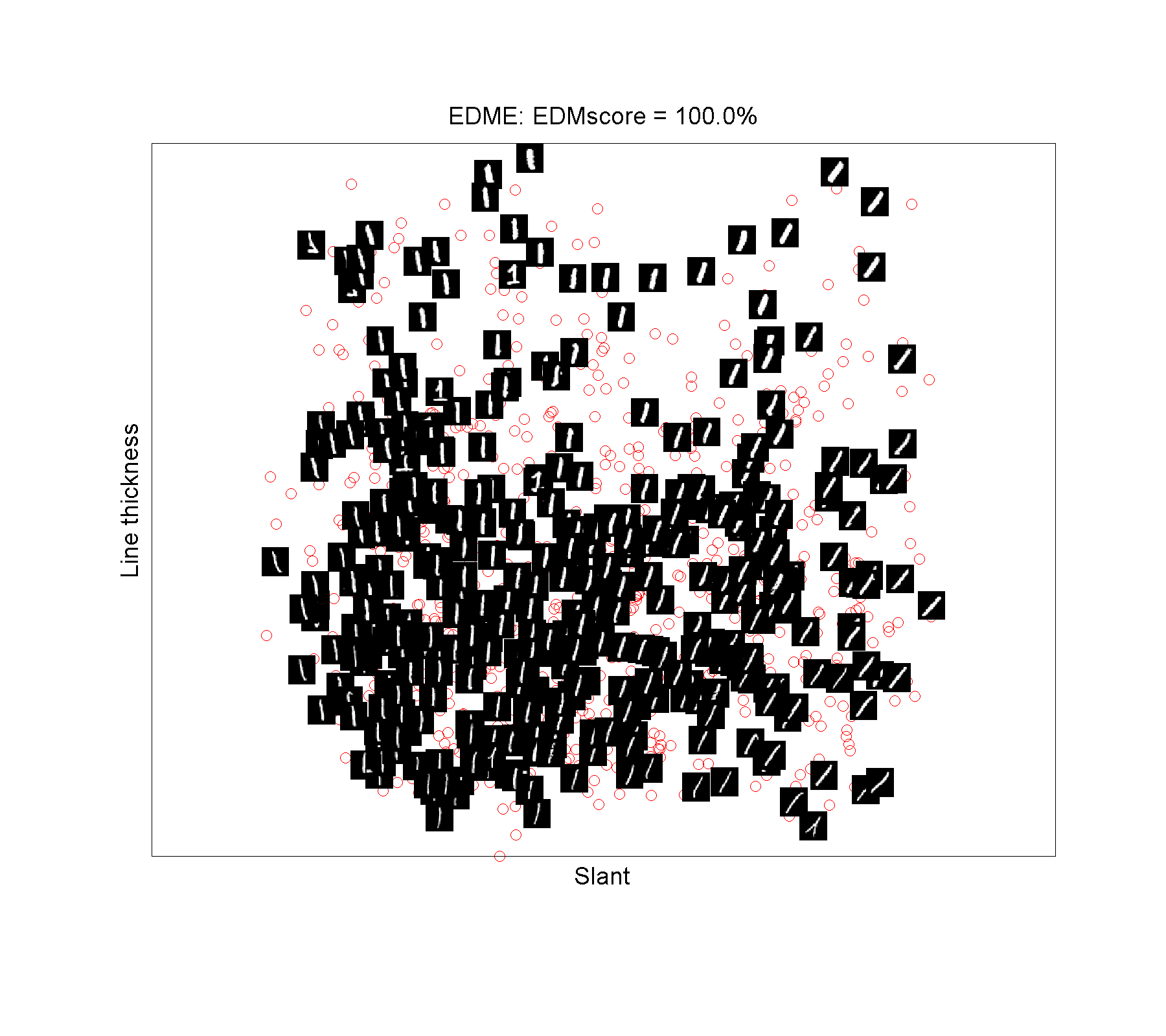}
%  \caption{EDME}
%  \label{fig:sub1}
\end{subfigure}%
\begin{subfigure}{.5\textwidth}
  \centering
  \includegraphics[width=1\linewidth]{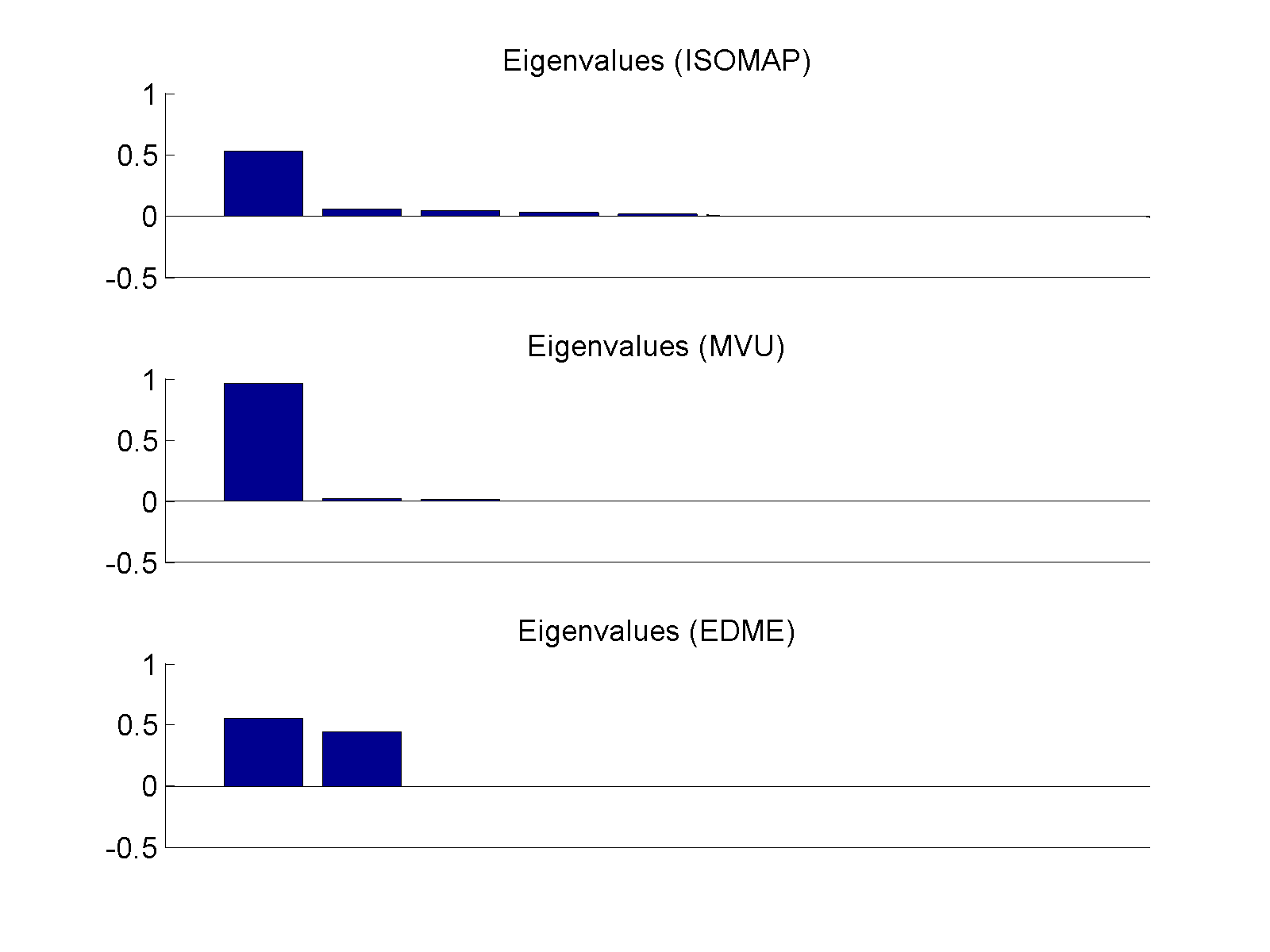}
%  \caption{MVU}
%  \label{fig:sub1}
\end{subfigure}
\caption{Digit 1}
\label{fig:Digit1_result}
\end{figure}

\noindent
{\bf (ML3) The digits Data}
%In this example, we consider the comparison the performance of different embedding methods such as the ISOMAP, the MVU and the proposed EDME method on the handwritten digits data from the MNIST database \cite{MNIST}. 
The data is from the MNIST database \citep{MNIST}. 
We first consider the data set of digit ``1'',
which includes $n=1135$ 8-bit grayscale images of ``1''. Each image has $28\times 28$ pixels, which is represented as $784$ dimensional vector. 
We note that the two most important features of ``1''s are the slant and the line thickness. 
Therefore, the embedding results are naturally expected to lie in the two dimensional space parameterized by these two major features. 
In this example, we set $k=6$.
Figure \ref{fig:Digit1_result} shows the two dimensional embeddings computed by ISOMAP, MVU and EDME. 
It can be clearly seen that EDME significantly outperforms the other two methods.
In particular, EDME is able to accurately represent the data in the two dimensional space
and captures the correct features. 
However, MVU returns an almost one dimensional embedding and only captures one of the major features, i.e., the slant of ``1''s. 
For the ISOMAP, it only captures a small percentage of the total variance.

\begin{figure}[h]
\centering
\begin{subfigure}{.5\textwidth}
  \centering
  \includegraphics[width=1\linewidth]{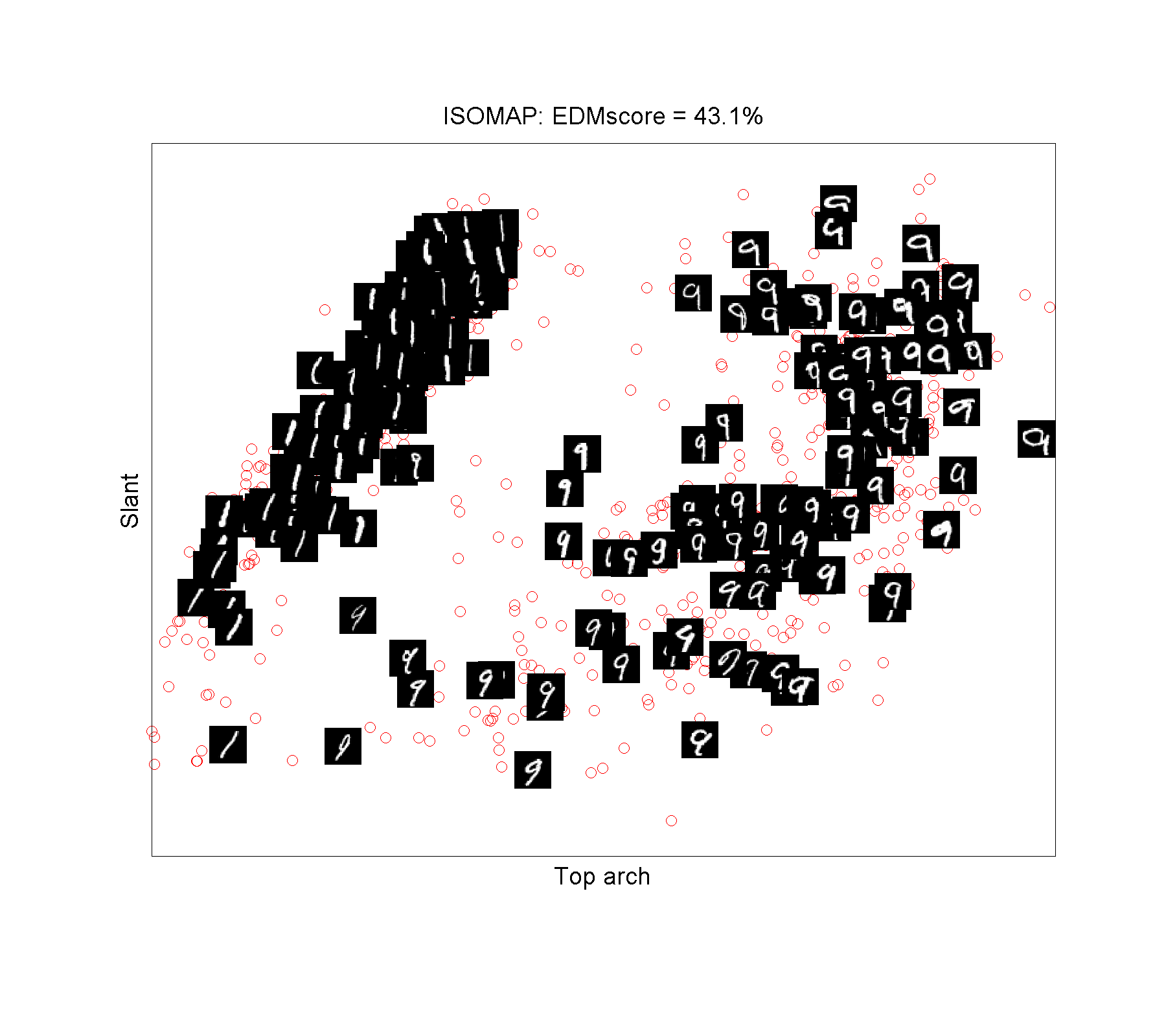}
%  \caption{PCA}
%  \label{fig:sub1}
\end{subfigure}%
\begin{subfigure}{.5\textwidth}
  \centering
  \includegraphics[width=1\linewidth]{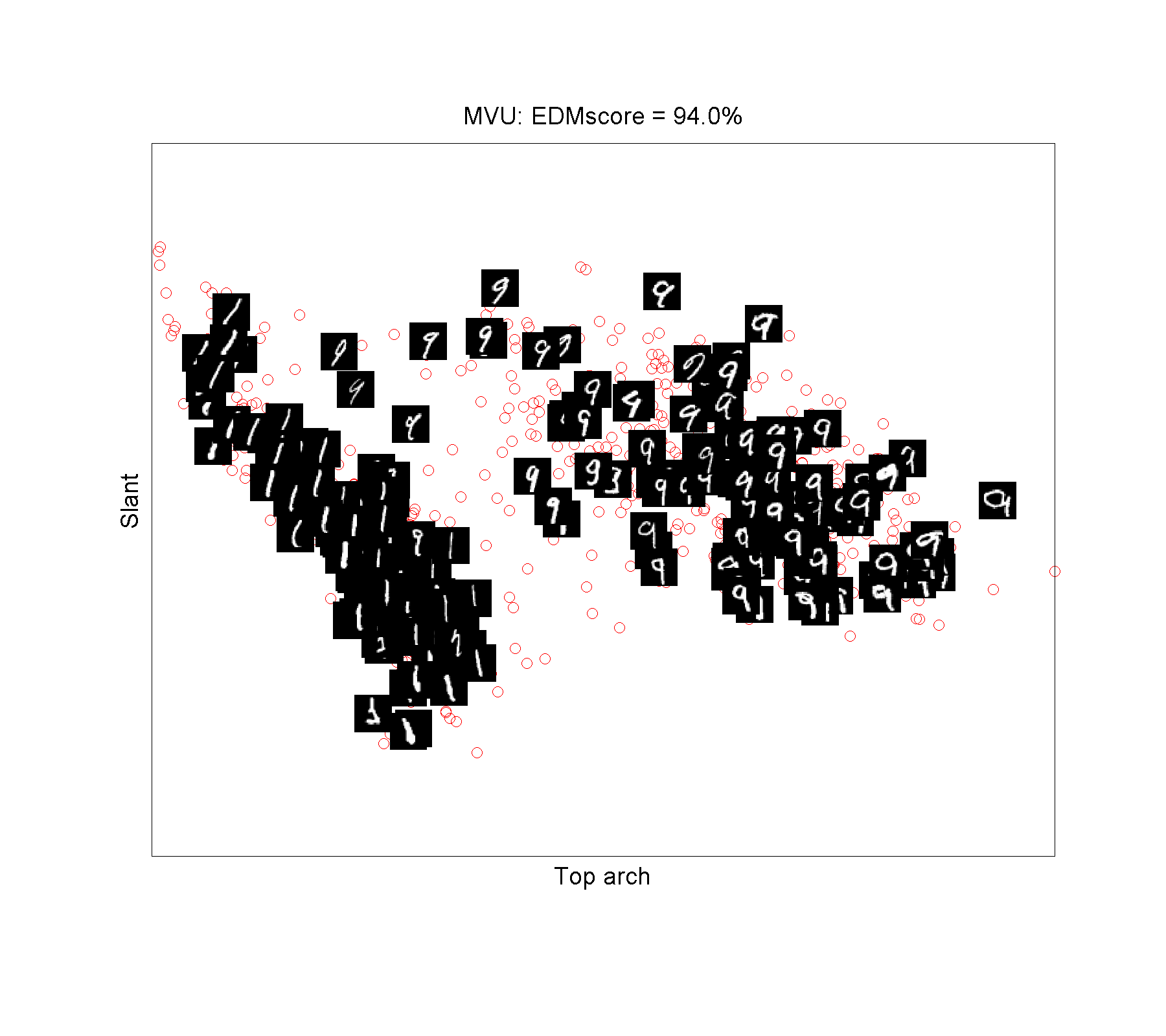}
%  \caption{PCA}
%  \label{fig:sub1}
\end{subfigure}
\begin{subfigure}{.5\textwidth}
  \centering
  \includegraphics[width=1\linewidth]{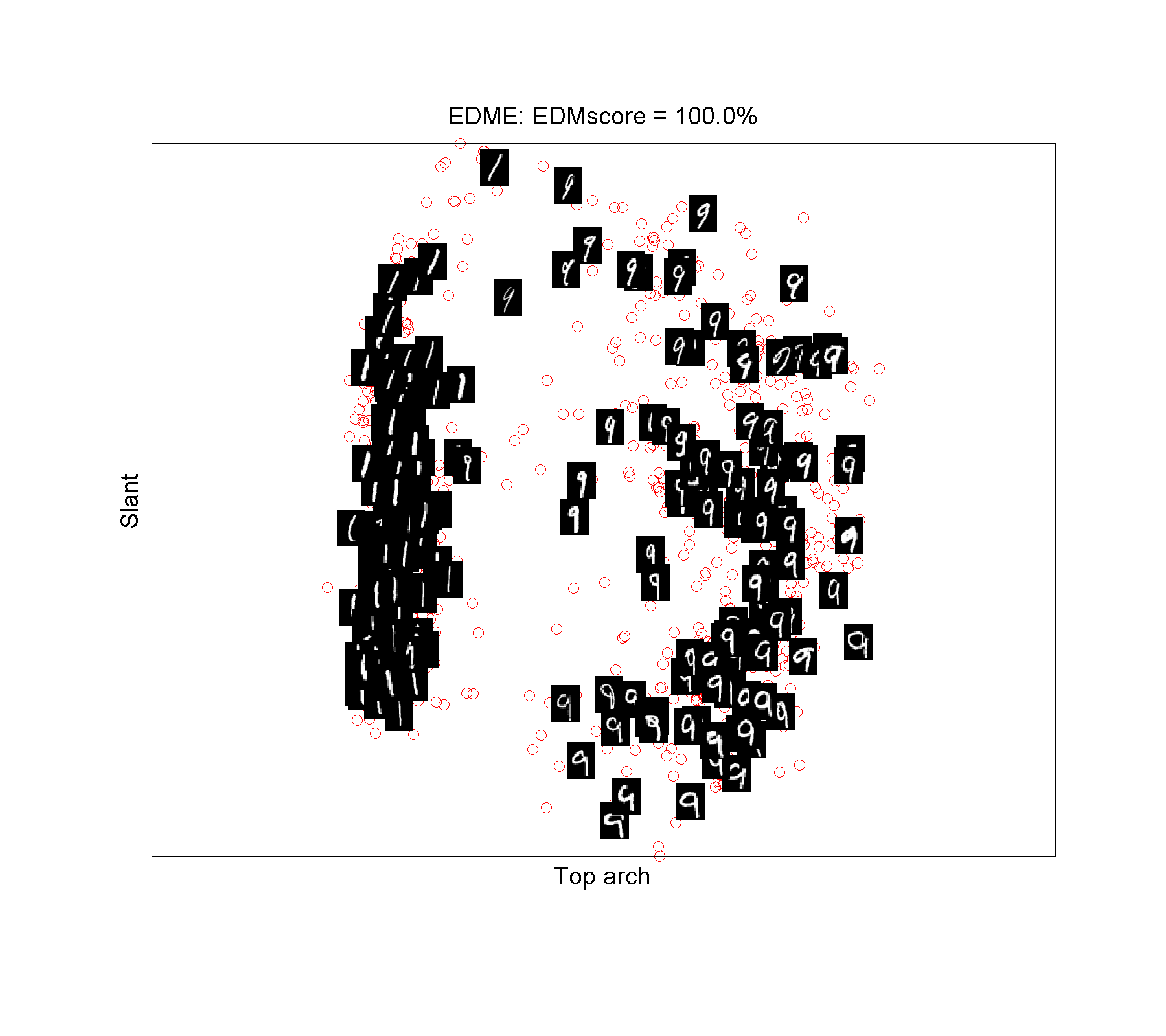}
%  \caption{EDME}
%  \label{fig:sub1}
\end{subfigure}%
\begin{subfigure}{.5\textwidth}
  \centering
  \includegraphics[width=1\linewidth]{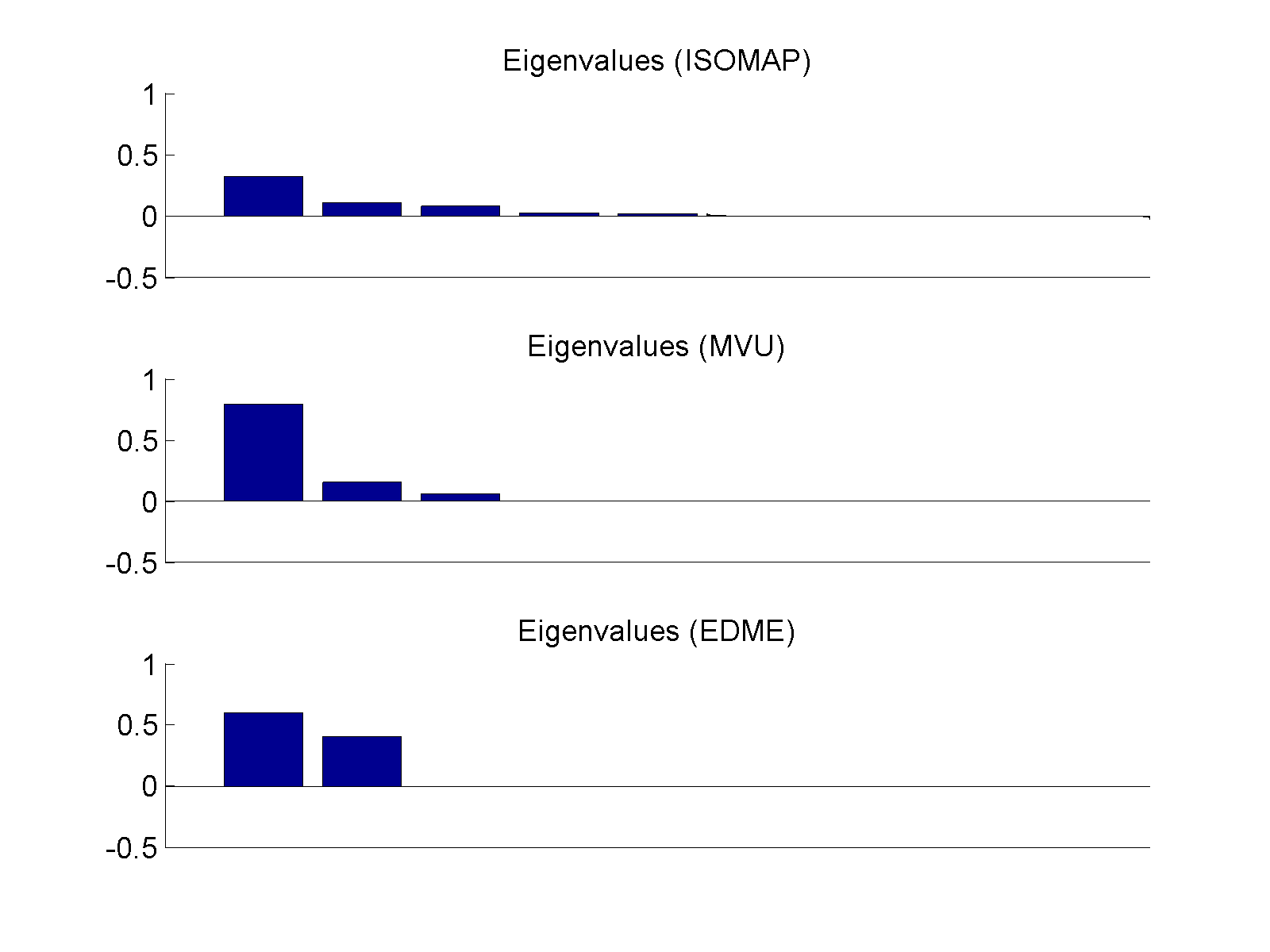}
%  \caption{Eigenvalue spectra of the Digits 6 \& 8}
%  \label{fig:sub1}
\end{subfigure}
\caption{Digits 1 \& 9}
\label{fig:Digits19_result}
\end{figure}

Next, we consider the learning task of the mixed digits ``1'' and ``9''. We randomly choose 500 images of ``1''s and 500 images of ``9''s from the MNIST. For the mixed images, the major features clearly become the size of the top arches and the slant of the digits. 
However, both ISOMAP and MVU fail to obtain a two dimensional embeddings, which are clearly indicated in the eigenvalue spectrum in Figure \ref{fig:Digits19_result}. 
Moreover, two dimensional projections (along the first two eigenvectors) of ISOMAP and MVU embeddings 
can not capture the desired two major features of images. 
In contrast, EDME successfully represents the data in the two dimensional space with the correct features (the top arch size and the slant of digits).

\begin{figure}[h]
\centering
\begin{subfigure}{.5\textwidth}
  \centering
  \includegraphics[width=1\linewidth]{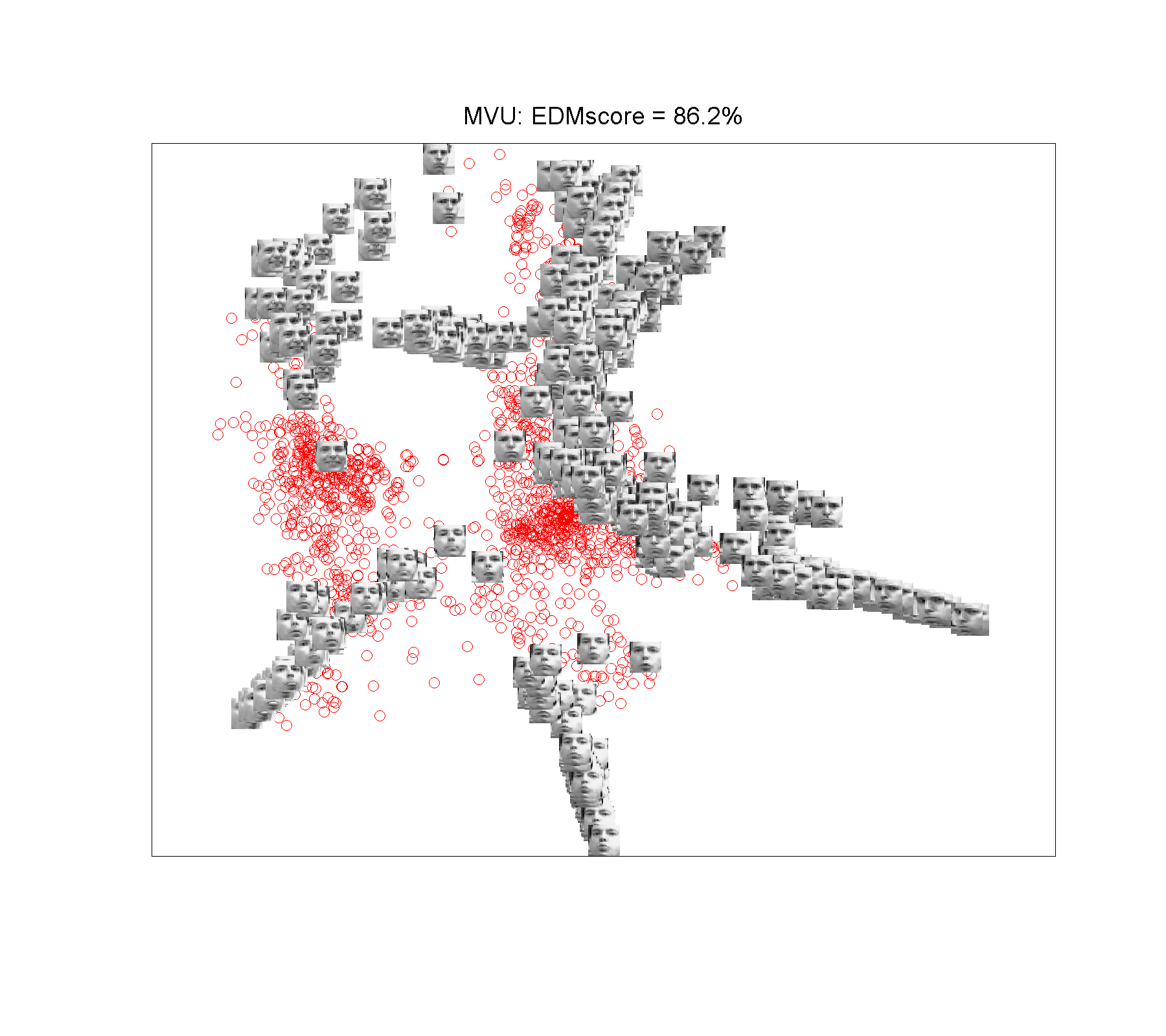}
%  \caption{PCA}
%  \label{fig:sub1}
\end{subfigure}%
\begin{subfigure}{.5\textwidth}
  \centering
  \includegraphics[width=1\linewidth]{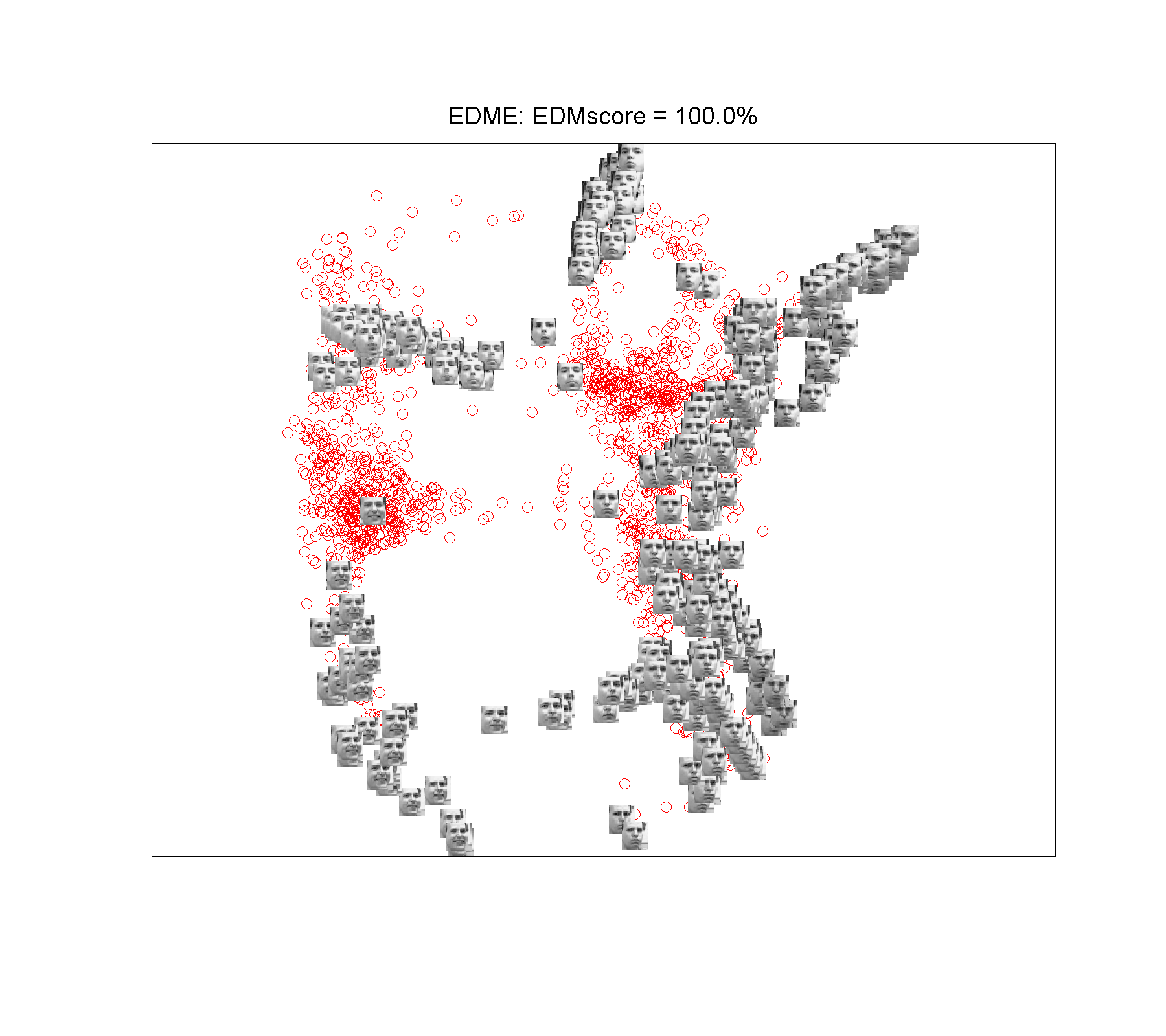}
%  \caption{EDME}
%  \label{fig:sub1}
\end{subfigure}
\begin{subfigure}{.5\textwidth}
  \centering
  \includegraphics[width=1\linewidth]{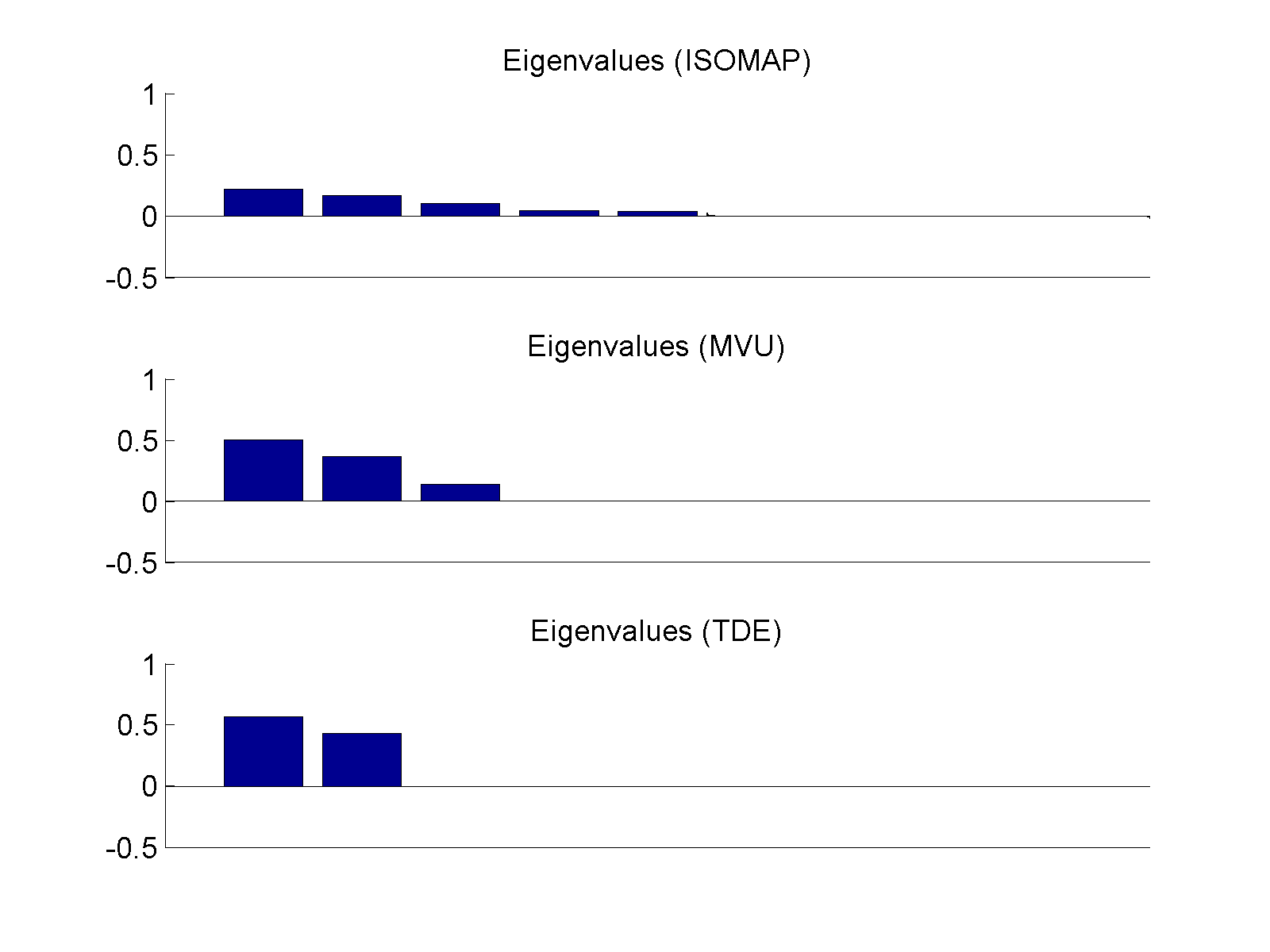}
%  \caption{MVU}
%  \label{fig:sub1}
\end{subfigure}
\caption{Frey face}
\label{fig:FaceFrey_result}
\end{figure}

\noindent
{\bf (ML4) Data of Frey face}
Finally, we consider the comparison of MVU and EDME on the Frey face images data \citep{Freyface}, which has
 $n=1965$ images of faces. Each image has $28\times 20$ gray scale pixels and
 is represented by a vector of $560$ dimensions. We set $k=4$. 
The representative faces are shown at some randomly chosen embedding points. 
From the eigenvalue spectrums shown in Figure \ref{fig:FaceFrey_result},  MVU returns a three dimensional embedding, while EDME is able to obtain a two dimensional embedding. 
Moreover, the two dimensional projection of the MVU embedding along the first two major eigenvectors 
only contain $86.2\%$ variance of the data, which is smaller than that from EDME. 

%\begin{figure}[H]
%\centering
%\includegraphics[width=1\linewidth]{eigsDigits2.png}
%\caption{Eigenvalue spectra of the Digit 2}
%\label{fig:Digit2_eigen}
%\end{figure}

%\begin{figure}[H]
%\centering
%\includegraphics[width=1\linewidth]{eigsDigits68.png}
%\caption{Eigenvalue spectra of the Digits 6 \& 8}
%\label{fig:Digits68_eigen}
%\end{figure}

%\begin{figure}[H]
%\centering
%\includegraphics[width=1\linewidth]{eigsFaceFrey.png}
%\caption{Eigenvalue spectra of the Frey face}
%\label{fig:FaceFrey_eigen}
%\end{figure}
%%%%%%%%%%%%%%%%%%%%%%%%%%%%%%%%%%%%%%%%%%%%%%%%%%%%%%%%%%%%%%%%%%%%%%%%%%%%%%
\subsection{Numerical performance} \label{Subsection_Numerical_Performance}
%%%%%%%%%%%%%%%%%%%%%%%%%%%

We tested the ISOMAP, the MVU and our proposed EDME methods in MATLAB R2014a (version 8.3.0.532), and the numerical experiments are run in MATLAB under a MAC OS X 10.9.3 64-bit system on an Intel 4 Cores i5 2.7GHz CPU with 8GB memory. 

In our numerical experiments, we use the SDPT3 \citep{TTT3}, a Matlab software package for semidefinite-quadratic-linear programming, to solve the corresponding SDP problem of the MVU model, since the SDPT3 is more efficient than other SDP solves such as CSDP \citep{Borchers99} and SeDuMi \citep{Sturm99}.
This is particularly the case for the large dimensional SDP problems (e.g., the Facebook-like communication network and the US airport). 
%Moreover, in this paper, we always use the  MVU with the relaxing constraints proposed by \cite{WS06}, since the distances from the real-world applications only provide a rough estimations of proximity relations. 
The termination tolerance of the SDPT3 is ${\rm tol}=10^{-3}$ (${\rm tol}=10^{-6}$ for the Enron email network and the Madrid train bombing).  For our EDME model, the proposed inexact accelerated proximal gradient (IAPG) method (Algorithm \ref{alg:APG}) is employed. We terminate the algorithm if the primal and dual infeasibilities conditions are met, i.e.,
\[
\max\{R_{p},R_{d}\}\leq {\rm tol},
\]
where $R_{p} = \|{\cal B}(X)-b\|$ and $R_{d} = (\nabla f(X)-{\cal B}^{*}y-Z)/(1+\|C\|)$ are the measurements of the infeasibilities of the primal and dual problems, respectively, where the tolerance is chosen by ${\rm tol}=10^{-3}$ (${\rm tol}=10^{-6}$ for the Enron email network and the Madrid train bombing). The details on the numerical performance of the MVU and EDME methods can be found from Table \ref{tab:numerical_perf},
where we report the EDM scores from the leading two eigenvalues and cpu time in seconds.

\begin{table}[H]
\begin{center} 
{\tiny\begin{tabular}{|cccc|ccc|cccc|}
\hline
\multicolumn{4}{|c|}{Problems} & \multicolumn{3}{|c|}{MVU (SDPT3)} & \multicolumn{4}{|c|}{EDME} \\
\hline
    &  & $n$  & ${\rm edges}$ & relgap (trace(XZ)) & EDMscore & cpu(s) & prim\_infeas & dual\_infeas & EDMscore & cpu(s)\\
\hline
{Enron} &  & {182}  &  2097  & 4.62e-07  & 48.1\%  & 9.52 & 1.33e-08 & 9.95e-07 & 100\%  & 4.65 \\
{Facebook-like} &  & {1893}  &  13835  & 4.81e-04  & 20.6\%  & 739.32 & 8.83e-06 & 9.70e-04 & 100\%  & 167.05 \\
{TrainBombing} &  & {64}  &  243  & 5.43e-07  & 91.8\%  & 0.81 & 3.19e-10 & 9.24e-07 & 100\%  & 2.42 \\
{USairport2010} &  & {1572}  &  17214  & 2.27e-03  & 74.3\%  & 45250.45 & 3.64e-07 & 6.43e-04 & 100\%  & 80.32 \\
{Blogs} &  & {1222}  &  16714  & 6.58e-04  & 60.5\%  & 3433.83 & 1.97e-07 & 6.82e-04 & 100\%  & 39.05 \\
\hline
  & $k$ & $n$  & ${\rm edges}$ & relgap (trace(XZ)) & EDMscore & cpu(s) & prim\_infeas & dual\_infeas & EDMscore & cpu(s)\\
\hline
{Teapots400} & {5} & {400}  &  1050  & 5.89e-04  & 100\%  & 3.86 & 4.44e-05 & 7.50e-04 & 100\%  & 1.90 \\
Face698  & 5 &  698   & 2164  & 4.88e-04  & 100\%  & 18.18  & 7.34e-08  & 7.27e-04 & 100\% & 4.73 \\
Digit1  & 6 &  1135   & 4885  & 7.69e-04  & 98.1\%  & 86.31  & 1.72e-07  & 6.60e-04 & 100\% & 21.20 \\
Digits19  & 6 &  1000   & 4394  & 5.69e-04  & 94.0\%  & 63.24  & 3.31e-07  & 6.46e-04 & 100\% & 10.49 \\
FreyFace  & 5 &  1965   & 6925  & 9.80e-04  & 86.2\%  & 271.17  & 1.79e-06  & 6.94e-04 & 100\% & 119.89 \\
\hline
\end{tabular}}
\end{center}
\caption{Numerical performance comparison of the MVU and the EDME}
\label{tab:numerical_perf}
\end{table}

We observe that the performance of EDME is outstanding in terms of numerical efficiency.
The developed IAPG is much faster than the SDP solver.
Taking USairport2010 as example, MVU used about $12$ hours while EDME only used about $80$ seconds.
For the examples in manifold learning, the gap between the two models are not as severe as for the 
social network examples.
The main reason is that the initial guess obtained by ISOMAP is a very good estimator that can 
roughly capture the low-dimensional features in manifold learning.
However, it fails to capture meaningful features for the social network examples.
This echoes the comment made in \cite{BJMM13}  that the shortest path distance is
not suitable to measure the distances in social networks.
We also like to point out that for all tested problems, EDME captured nearly $100\%$ variance and
it treats the local features equally important in terms of the leading eigenvalues being of
the same magnitude.

%%%%%%%%%%%%%%%%%%%%%%%%%%%%%%%%%%%%%%%%%%%%%%%%%%%%%%%%%%%
\section{Conclusions} \label{Section-Conclusion}
%%%%%%%%%%%%%%%%%%%%%%%%%%%%

The paper aimed to explain a mysterious situation regarding the SDP methodology to
reconstruct faithful Euclidean distances in a low-dimensional space from incomplete
set of noisy distances. The SDP models can construct numerical configurations of high quality,
but they lack theoretical backups in terms of bounding errors. 
We took a completely different approach that heavily makes use of Euclidean Distance Matrix instead of
positive semidefinite matrix in SDP models.
This led to a convex optimization that inherits the nice features of MVU and MVE models.
More importantly, we were able to derive error-bound results under the uniform sampling rule.
The optimization problem can also be efficiently solved by the proposed algorithm.
Numerical results in both social networks and manifold leading showed that our model
can capture low-dimensional features and treats them equally important.

Given that our model worked very well for the manifold learning examples, 
an interesting question regarding this approach is whether the theoretical error-bound results
can be extended to the case where the distances are obtained by the k-NN rule. 
It seems very difficult if we follow the technical proofs in this paper.
It also seems that the approach of \cite{JMontanari13} would lead to some
interesting (but very technical) results. We plan to investigate those issues in
future.

% Acknowledgements should go at the end, before appendices and references

\acks{We would like to acknowledge partial support for this project
from the Engineering and Physical Sciences Research Council project EP/K0076451.
Chao Ding is partially supported by the National Natural Science Foundation of China (Grant No. 11301515). }


%\newpage

\appendix
%\section{Appendix}

\section{Proof of Proposition \ref{prop:Error_bound_origin}}
 For any $D\in{\cal S}^n$, we know from \eqref{eq:observation_model} that
\begin{eqnarray}
\frac{1}{2m}\|y\circ y-{\cal O}(D)\|^2&=&\frac{1}{2m}\big\| {\cal O}(\overline{D}^{1/2})\circ {\cal O}(\overline{D}^{1/2})+2\eta{\cal O}(\overline{D}^{1/2})\circ \xi+\eta^2\xi\circ\xi -{\cal O}(D) \big\|^{2}\nonumber\\[3pt]
&=&\frac{1}{2m}\|{\cal O}(\overline{D})+2\eta{\cal O}(\overline{D}^{1/2})\circ\xi+\eta^2\xi\circ\xi-{\cal O}(D)\|^2 \nonumber\\[3pt]
&=&\frac{1}{2m}\|{\cal O}(D-\overline{D})-\eta\zeta\|^2\nonumber\\[3pt]
&=& \frac{1}{2m}\|{\cal O}(D-\overline{D})\|^2-\frac{\eta}{m}\langle {\cal O}(D-\overline{D}),\zeta\rangle + \frac{\eta^2}{2m}\|\zeta\|^2. \label{eq:diff-eq}
\end{eqnarray}
In particular, we have $\displaystyle{\frac{1}{2m}}\|y\circ y-{\cal O}(\overline{D})\|^2=\displaystyle{\frac{\eta^2}{2m}}\|\zeta\|^2$. Since $D^*$ is the optimal solution of \eqref{eq:estimator_problem} and $\overline{D}$ is also feasible, we obtain that
\begin{eqnarray*}
\frac{1}{2m}\|y\circ y-{\cal O}(D^*)\|^2&\leq&\frac{1}{2m}\|y\circ y-{\cal O}(\overline{D})\|^2\\ [3pt]
&&+\rho_1\left[\langle I, -J(\overline{D}-D^*)J \rangle -\rho_2\langle\widetilde{P}_1\widetilde{P}_1^T,-J(\overline{D}-D^*)J\rangle\right]
\end{eqnarray*}
Therefore, we know from \eqref{eq:diff-eq} that
\begin{equation}\label{eq:Err_o_3}
\frac{1}{2m}\|{\cal O}(D^*-\overline{D})\|^2\le \frac{\eta}{m}\langle {\cal O}(D^*-\overline{D}),\zeta\rangle+\rho_1\left[-\langle I, -J(D^*-\overline{D})J \rangle +\rho_2\langle\widetilde{P}_1\widetilde{P}_1^T,-J(D^*-\overline{D})J\rangle\right].
\end{equation}
For the first term of the right hand side of \eqref{eq:Err_o_3}, we have
\begin{eqnarray}
&&\frac{\eta}{m}\langle {\cal O}(D^*-\overline{D}),\zeta\rangle= \frac{\eta}{m}\langle D^*-\overline{D},{\cal O}^*(\zeta)\rangle\le \eta\left\|\frac{1}{m}{\cal O}^*(\zeta)\right\|_2\|D^*-\overline{D}\|_* \nonumber\\ [3pt]
&=& \eta\left\|\frac{1}{m}{\cal O}^*(\zeta)\right\|_2\|D^*-\overline{D}-J(D^*-\overline{D})J+J(D^*-\overline{D})J\|_* \nonumber\\ [3pt]
&\le & \eta\left\|\frac{1}{m}{\cal O}^*(\zeta)\right\|_2\left(\|D^*-\overline{D}-J(D^*-\overline{D})J\|_*+\|-J(D^*-\overline{D})J\|_*\right).\label{eq:first_term}
\end{eqnarray}
By noting that $D^*$, $\overline{D}\in{\cal S}^n_H$, we know from Lemma \ref{lemma:observation-hollow} that the rank of $D^*-\overline{D}-J(D^*-\overline{D})J$ is no more than $2$, which implies
\[
\|D^*-\overline{D}-J(D^*-\overline{D})J\|_*\le \sqrt{2}\|D^*-\overline{D}-J(D^*-\overline{D})J\|.
\]
Moreover, it follows from \eqref{JXJ} that $\langle J(D^*-\overline{D})J,D^*-\overline{D}-J(D^*-\overline{D})J \rangle=0$, which implies
\begin{equation}\label{eq:project_center}
\|D^*-\overline{D}\|^2=\|D^*-\overline{D}-J(D^*-\overline{D})J\|^2+\|J(D^*-\overline{D})J\|^2.
\end{equation}
Thus, we have
\begin{equation}\label{eq:first_term-1}
\|D^*-\overline{D}-J(D^*-\overline{D})J\|_*\le \sqrt{2}\|D^*-\overline{D}\|.
\end{equation}
By noting that ${\cal P}_{T}(-J(D^*-\overline{D})J)+{\cal P}_{T^\perp}(-J(D^*-\overline{D})J)=-J(D^*-\overline{D})J$, we know from \eqref{eq:first_term} and \eqref{eq:first_term-1} that
\begin{eqnarray}
\frac{\eta}{m}\langle {\cal O}(D^*-\overline{D}),\zeta \rangle &\le& \big\|\frac{\eta}{m}{\cal O}^*(\zeta)\big\|_2\Big(\sqrt{2}\|D^*-\overline{D}\|+\|{\cal P}_{T}(-J(D^*-\overline{D})J)\|_* \nonumber\\[3pt]
&&  +\|{\cal P}_{T^\perp}(-J(D^*-\overline{D})J)\|_*\Big). \label{eq:first_term-2}
\end{eqnarray}

Meanwhile, since for any $A\in{\cal S}^n$, $\|{\cal P}_{T}(A)\|_*=\|\overline{P}_2^TA\overline{P}_2\|_*$, we know from the directional derivative formulate of the nuclear norm \citep[Theorem 1]{Watson92} that
\begin{eqnarray*}
\|-JD^*J\|_*-\|-J\overline{D}J\|_*&\ge& \langle \overline{P}_1\overline{P}_1^T,-J(D^*-\overline{D})J\rangle+\|\overline{P}_2^T(-J(D^*-\overline{D})J)\overline{P}_2\|_*\\ [3pt]
&=&\langle \overline{P}_1\overline{P}_1^T,-J(D^*-\overline{D})J\rangle+\|{\cal P}_{T}(-J(D^*-\overline{D})J)\|_*. 
\end{eqnarray*}
Thus, since $-JD^*J$, $-J\overline{D}J\in{\cal S}^n_+$, we have $-\langle I, -J(D^*-\overline{D})J \rangle=-(\|-JD^*J\|_*-\|-J\overline{D}J\|_*)$, which implies that
\begin{eqnarray*}
&&-\langle I, -J(D^*-\overline{D})J \rangle +\rho_2\langle\widetilde{P}_1\widetilde{P}_1^T,-J(D^*-\overline{D})J\rangle \\[3pt]
&\leq& -\langle \overline{P}_1\overline{P}_1^T,-J(D^*-\overline{D})J\rangle-\|{\cal P}_{T}(-J(D^*-\overline{D})J)\|_*+\rho_2\langle\widetilde{P}_1\widetilde{P}_1^T,-J(D^*-\overline{D})J\rangle.
\end{eqnarray*}
By using the decomposition \eqref{eq:decomp_T} and the notations defined in \eqref{eq:notation_alpha_beta}, we conclude from \eqref{eq:project_center} that
\begin{eqnarray*}
&&-\langle I, -J(D^*-\overline{D})J \rangle +\rho_2\langle\widetilde{P}_1\widetilde{P}_1^T,-J(D^*-\overline{D})J\rangle \\[3pt]
&\leq& -\langle \overline{P}_1\overline{P}_1^T-\rho_{2}\widetilde{P}_1\widetilde{P}_1^T,-J(D^*-\overline{D})J\rangle-\|{\cal P}_{T}(-J(D^*-\overline{D})J)\|_* \\[3pt]
&\leq& \|\overline{P}_1\overline{P}_1^T-\rho_2\widetilde{P}_1\widetilde{P}_1^T\|\|J(D^*-\overline{D})J\|-\|{\cal P}_{T}(-J(D^*-\overline{D})J)\|_*\\[3pt]
&\leq& \alpha(\rho_2)\sqrt{2r}\|D^*-\overline{D}\|-\|{\cal P}_{T}(-J(D^*-\overline{D})J)\|_*.
\end{eqnarray*}
Thus, together with \eqref{eq:first_term-2}, we know from \eqref{eq:Err_o_3} that
\begin{eqnarray}
&&\frac{1}{2m}\|{\cal O}(D^*-\overline{D})\|^2\nonumber\\[3pt]
&\le& \Big(\sqrt{2}\eta\big\|\frac{1}{m}{\cal O}^*(\zeta)\big\|_2+\sqrt{2r}\rho_1\alpha(\rho_2)\Big)\|D^*-\overline{D}\|+\eta\big\|\frac{1}{m}{\cal O}^*(\zeta)\big\|_2\|{\cal P}_{T^{\perp}}(-J(D^*-\overline{D})J)\|_*\nonumber \\[3pt]
&&-\big(\rho_1- \eta\big\|\frac{1}{m}{\cal O}^*(\zeta)\big\|_2\big)\|{\cal P}_{T}(-J(D^*-\overline{D})J)\|_*.  \label{eq:Err_o_4}
\end{eqnarray}
Since $\eta\big\|\frac{1}{m}{\cal O}^*(\zeta)\big\|_2\leq \displaystyle{\frac{\rho_1}{\kappa}}$ and $\kappa>1$, we know from \eqref{eq:PT_bounds} and \eqref{eq:project_center} that
\begin{eqnarray}
&&\frac{1}{2m}\|{\cal O}(D^*-\overline{D})\|^2\nonumber\\[3pt]
&\le& \left(\frac{1}{\kappa}\sqrt{2}+\alpha(\rho_2)\sqrt{2r}\right)\rho_1\|D^*-\overline{D}\|+\frac{1}{\kappa}\sqrt{2r}\rho_1\|D^*-\overline{D}\|\nonumber \\[3pt]
&&-\big(1- \frac{1}{\kappa}\big)\rho_1\|{\cal P}_{T}(-J(D^*-\overline{D})J)\|_*  \nonumber \\[3pt]
&\le& \left(\frac{1}{\kappa}(\sqrt{2}+\sqrt{2r})+\alpha(\rho_2)\sqrt{2r}\right)\rho_1\|D^*-\overline{D}\|-\frac{\kappa-1}{\kappa}\rho_1\|{\cal P}_{T}(-J(D^*-\overline{D})J)\|_*  \label{eq:Err_o_5} \\[3pt]
&\leq& \left(\frac{1}{\kappa}(\sqrt{2}+\sqrt{2r})+\alpha(\rho_2)\sqrt{2r}\right)\rho_1\|D^*-\overline{D}\|. \label{eq:Err_o_6}
\end{eqnarray}
Since $r\ge 1$, the desired inequality \eqref{eq:Err_origin_1} follows from \eqref{eq:Err_o_6}, directly.

Next we shall show that  \eqref{eq:Err_origin_2} also holds. By \eqref{eq:Err_o_5}, we have
\[
\|{\cal P}_{T}(-J(D^*-\overline{D})J)\|_*\leq\frac{\kappa}{\kappa-1} \left(\frac{\sqrt{2}}{\kappa}+\big(\alpha(\rho_2)+\frac{1}{\kappa}\big)\sqrt{2r}\right)\|D^*-\overline{D}\|.
\]
Therefore, by combining with \eqref{eq:first_term-1} and \eqref{eq:PT_bounds}, we know from the decomposition \eqref{eq:decomp_T} that
\begin{eqnarray*}
\|D^*-\overline{D}\|_*&\leq& \|D^*-\overline{D}-J(D^*-\overline{D})J\|_*+\|{\cal P}_{T^{\perp}}(-J(D^*-\overline{D})J)\|_*+\|{\cal P}_{T}(-J(D^*-\overline{D})J)\|_* \\ [3pt]
&\leq& (\sqrt{2}+\sqrt{2r})\|D^*-\overline{D}\|+\frac{\kappa}{\kappa-1} \left(\frac{\sqrt{2}}{\kappa}+\big(\alpha(\rho_2)+\frac{1}{\kappa}\big)\sqrt{2r}\right)\|D^*-\overline{D}\|. 
\end{eqnarray*}
Finally, since $r\ge 1$, we conclude that
\begin{eqnarray*}
\|D^*-\overline{D}\|_*&\leq& \frac{\kappa}{\kappa-1}\sqrt{2}\|D^*-\overline{D}\|+ \frac{\kappa}{\kappa-1}\left(\alpha(\rho_2)+1\right)\sqrt{2r}\|D^*-\overline{D}\|\\ [3pt]
&\leq& \frac{\kappa}{\kappa-1}\left(\alpha(\rho_2)+2\right)\sqrt{2r}\|D^*-\overline{D}\|.
\end{eqnarray*}
This completes the proof. 

\section{Proof of Lemma \ref{lemma:approx_RIP}}
 Firstly, we shall show that for any $A\in{\cal C}(\tau)$, the following inequality holds with probability at least $1-1/n$,
\[
\frac{1}{m}\|{\cal O}(A)\|^2\ge\frac{1}{2}{\mathbb E}\left(\langle A, X \rangle^2\right)-256r|\Omega|\left({\mathbb E}\Big(\big\|\frac{1}{m}{\cal O}^*(\varepsilon)\big\|_{2}\Big)\right)^2,
\]
where $\varepsilon=(\varepsilon_1,\ldots,\varepsilon_m)^T\in\Re^m$ with $\{\varepsilon_1,\ldots,\varepsilon_m\}$ is an i.i.d. Rademacher sequence, i.e., a sequence of i.i.d. Bernoulli random variables taking the values $1$ and $-1$ with probability $1/2$. This part of proof is similar with that of Lemma 12 in \cite{Klopp12} \citep[see also][Lemma 2]{MPSun12}. However, we include the proof here for seeking of completion. 

Denote $\Sigma:=256r|\Omega|\left({\mathbb E}\Big(\big\|\frac{1}{m}{\cal O}^*(\varepsilon)\big\|_{2}\Big)\right)^2$. We will show that the probability of the following ``bad" events is small
\[
{\cal B}:=\left\{\exists\,A\in{\cal C}(\tau)\ \mbox{such that}\ \left|\frac{1}{m}\|{\cal O}(A)\|^2-{\mathbb E}\left(\langle A,X\rangle^2\right)\right|>\frac{1}{2}{\mathbb E}\left(\langle A,X\rangle^2\right)+\Sigma \right\}.
\]
It is clear that the events interested are included in ${\cal B}$. Next, we will use a standard peeling argument to estimate the probability of ${\cal B}$. For any $\nu>0$, we have
\[
{\cal C}(\tau)\subseteq\bigcup_{k=1}^\infty \left\{A\in{\cal C}(\tau)\mid 2^{k-1}\nu\le {\mathbb E}\left(\langle A,X \rangle^2\right)\le 2^k\nu  \right\}.
\]
Thus, if the event ${\cal B}$ holds for some $A\in{\cal C}(\tau)$, then there exists some $k\in{\mathbb N}$ such that $2^{k}\nu\ge{\mathbb E}\left(\langle A,X \rangle^2\right)\ge 2^{k-1}\nu$. Therefore, we have
\[
\left|\frac{1}{m}\|{\cal O}(A)\|^2-{\mathbb E}\left(\langle A,X\rangle^2\right)\right|>\frac{1}{2}2^{k-1}\nu + \Sigma = 2^{k-2}\nu +\Sigma.
\]
This implies that ${\cal B}\subseteq \bigcup_{k=1}^{\infty}{\cal B}_k$, where for each $k$,
\[
{\cal B}_k:=\left\{\exists\,A\in{\cal C}(\tau)\ \mbox{such that}\ \left|\frac{1}{m}\|{\cal O}(A)\|^2-{\mathbb E}\left(\langle A,X\rangle^2\right)\right|>2^{k-2}\nu+\Sigma,\ {\mathbb E}\left(\langle A,X\rangle^2\right)\le 2^k\nu  \right\}.
\]
We shall estimated the probability of each ${\cal B}_k$. For any given $\Upsilon>0$, define the set ${\cal C}(\tau;\Upsilon):=\left\{A\in{\cal C}(\tau)\mid {\mathbb E}\left(\langle A,X\rangle^2\right)\le \Upsilon\right\}$. For any given $\Upsilon>0$, denote
\[
Z_\Upsilon:=\sup_{A\in{\cal C}(\tau;\Upsilon)}\left|\frac{1}{m}\|{\cal O}(A)\|^2-{\mathbb E}\left(\langle A,X\rangle^2\right)\right|.
\]
We know from \eqref{eq:def_obser_op}, the definition of the observation operator ${\cal O}$, that
\[
\frac{1}{m}\|{\cal O}(A)\|^2-{\mathbb E}\left(\langle A,X\rangle^2\right)=\frac{1}{m}\sum_{l=1}^m\langle A,X_l\rangle^2 - {\mathbb E}\left(\langle A,X\rangle^2\right).
\]
Meanwhile, since $\|A\|_\infty = 1/\sqrt{2}$, we have for each $l\in\{1,\ldots,m\}$,
\[
\left|\langle A, X_l \rangle^2 - {\mathbb E}\left(\langle A,X\rangle^2\right) \right|\le 2\|A\|_\infty^2 =1.
\]
Thus, it follows from Massart's concentration inequality \citep[see, e.g.,][Theorem 14.2]{BGeer11} that
\begin{equation}\label{eq:concentr_1}
{\mathbb P}\left(Z_\Upsilon\ge {\mathbb E}(Z_\Upsilon)+\frac{T}{8} \right)\le {\rm exp}\left(\frac{-m\Upsilon^2}{512}\right).
\end{equation}
By applying the standard Rademacher symmetrization \citep[see, e.g.,][Theorem 2.1]{Koltchinskii11a}, we obtain that
\[
{\mathbb E}(Z_\Upsilon)={\mathbb E}\left(\sup_{A\in{\cal C}(\tau;\Upsilon)}\left|\frac{1}{m}\sum_{l=1}^m\langle A,X_l\rangle^2 - {\mathbb E}\left(\langle A,X\rangle^2\right)\right|\right)\le 2 {\mathbb E}\left(\sup_{A\in{\cal C}(\tau;\Upsilon)}\left|\frac{1}{m}\sum_{l=1}^m\varepsilon_l\langle A,X_l\rangle^2\right| \right),
\]
where $\{\varepsilon_1,\ldots,\varepsilon_m\}$ is an i.i.d. Rademacher sequence. Again, since $\|A\|_\infty=1/\sqrt{2}$, we know that $|\langle A, X_i\rangle|\le \|A\|_\infty<1$. Thus, it follows from the contraction inequality \citep[see, e.g.,][Theorem 4.12]{LTalagrand91} that
\begin{eqnarray*}
{\mathbb E}(Z_\Upsilon)&\le& 8{\mathbb E}\left(\sup_{A\in{\cal C}(\tau;\Upsilon)}\left|\frac{1}{m}\sum_{l=1}^m\varepsilon_l\langle A,X_l\rangle\right| \right)=8{\mathbb E}\left(\sup_{A\in{\cal C}(\tau;\Upsilon)}\big|\langle \frac{1}{m}{\cal O}^*(\varepsilon),A\rangle\big| \right)\\
&\le& 8{\mathbb E}\left(\|\frac{1}{m}{\cal O}^*(\varepsilon)\|\right)\left(\sup_{A\in{\cal C}(\tau;\Upsilon)}\|A\|_*\right).
\end{eqnarray*}
For any $A\in{\cal C}(\tau;\Upsilon)$, we have
\[
\|A\|_*\leq\sqrt{\tau}\|A\|= \sqrt{2\tau|\Omega|{\mathbb E}\left(\langle A,X\rangle^2\right)}\le \sqrt{2\tau|\Omega|\Upsilon}.
\]
Thus, we obtain that
\[
{\mathbb E}(Z_\Upsilon)+\frac{\Upsilon}{8}\le 8{\mathbb E}\left(\|\frac{1}{m}{\cal O}^*(\varepsilon)\|\right)\left(\sup_{A\in{\cal C}(\tau;\Upsilon)}\|A\|_*\right)+\frac{\Upsilon}{8}\le 256\tau|\Omega|\left({\mathbb E}\big(\frac{1}{m}{\cal O}^*(\varepsilon)\big)\right)^2+\frac{\Upsilon}{4}.
\]
It follows from \eqref{eq:concentr_1} that
\[
{\mathbb P}\left( Z_\Upsilon\ge \frac{\Upsilon}{4} + 256\tau|\Omega|\left({\mathbb E}\big(\frac{1}{m}{\cal O}^*(\varepsilon)\big)\right)^2 \right)\le {\mathbb P}\left(Z_\Upsilon\ge {\mathbb E}(Z_\Upsilon)+\frac{\Upsilon}{8} \right) \le {\rm exp}\left(\frac{-m\Upsilon^2}{512}\right).
\]
By choosing $\Upsilon=2^k\nu$, we obtain from the above inequality that
\[
{\mathbb P}({\cal B}_k)\le {\rm exp}\left(\frac{-4^k\nu^2m}{512}\right).
\]
By noting that $\log(x)<x$ for any $x>1$, we conclude that
\[
{\mathbb P}({\cal B})\le \sum_{k=1}^\infty{\mathbb P}({\cal B}_k)\le \sum_{k=1}^\infty {\rm exp}\left(\frac{-4^k\nu^2m}{512}\right)<\sum_{k=1}^\infty{\rm exp}\left(\frac{-\log(4)k\nu^2m}{512}\right)\le \frac{{\rm exp}\left(\frac{-\log(2)k\nu^2m}{256}\right)}{1-{\rm exp}\left(\frac{-\log(2)k\nu^2m}{256}\right)}.
\]
Choosing $\nu=\displaystyle{\sqrt{\frac{256\log(2n)}{m\log(2)}}}$, it yields ${\mathbb P}({\cal B})\le {1}/(2n-1)\le 1/n$.

Finally, the lemma then follows if we prove that for $m>C_1n\log n$ with $C_1>1$, there exists a constant $C_1'>0$ such that
\begin{equation}\label{eq:E-0}
{\mathbb E}\Big(\big\|\frac{1}{m}{\cal O}^*(\varepsilon)\big\|_{2}\Big)\le C_1'\sqrt{\frac{\log (2n)}{mn}}.
\end{equation}
The following proof is similar with that of Lemma 7 \cite{Klopp11} \citep[see, e.g.,][Lemma 6]{Klopp12}. We include it again for seeking of completion. Denote $Z_{l}:=\varepsilon_{l}X_{l}$, $l=1,\ldots,m$. Since $\{\varepsilon_{1},\ldots,\varepsilon_{m}\}$ is an i.i.d. Rademacher sequence, we have $|Z_{l}\|_{2}=1/2$ for all $l$. Moreover, 
\[
\|{\mathbb E}(Z_l^2)\|_2=\|{\mathbb E}(\varepsilon_{l}^2X_l^2)\|_2=\|{\mathbb E}(X_l^2)\|_2=\frac{1}{4|\Omega|}\big\|\sum_{1\le i<j\le n}(e_ie_j^T+e_je_i^T)^2\big\|_2= \frac{1}{4|\Omega|}(n-1)=\frac{1}{2n}.
\]
By applying the Bernstein inequality (Lemma \ref{lem:Bernstein-ineq-sym}), we obtain the following tail bound for any $t>0$,
\begin{equation}\label{eq:propability-0}
{\mathbb P}\Big(\big\|\frac{1}{m}{\cal O}^*(\varepsilon) \big\|_2\ge t \Big)\leq 2n\max\left\{ {\rm exp}\left(-\frac{nmt^2}{2}\right), {\rm exp}(-mt) \right\}.
\end{equation}
By H\"{o}lder's inequality, we have
\begin{eqnarray}
&&{\mathbb E}\Big(\big\|\frac{1}{m}{\cal O}^*(\varepsilon)\big\|_{2}\Big)\le \left({\mathbb E}\Big(\big\|\frac{1}{m}{\cal O}^*(\varepsilon)\big\|_{2}^{2\log(2n)}\Big)\right)^{\frac{1}{2\log(2n)}}\nonumber\\ [3pt]
&=&\left(\int_{0}^{\infty}{\mathbb P}\Big(\big\|\frac{1}{m}{\cal O}^*(\varepsilon) \big\|_2\ge t^{\frac{1}{2\log(2n)}} \Big)dt\right)^{\frac{1}{2\log(2n)}}\nonumber\\ [3pt]
&\leq & \left(2n \int_{0}^{\infty}{\rm exp}\left(-\frac{1}{2}nmt^{\frac{1}{\log(2n)}}\right) dt+ 2n \int_{0}^{\infty}{\rm exp}\left(-mt^{\frac{1}{2\log(2n)}}\right) dt \right)^{\frac{1}{2\log(2n)}}\nonumber \\ [3pt]
& = & e^{1/2}\left(\log(2n)(\frac{nm}{2})^{-\log(2n)}\Gamma(\log(2n))+2\log(2n)m^{-2\log(2n)}\Gamma(2\log(2n))\right)^{\frac{1}{2\log(2n)}}. \label{eq:E-1}
\end{eqnarray}
Since for $x\ge 2$, $\Gamma(x)\le(x/2)^{x-1}$, we obtain from \eqref{eq:E-1} that for $n\ge 4$, 
\begin{equation}\label{eq:E-2}
{\mathbb E}\Big(\big\|\frac{1}{m}{\cal O}^*(\varepsilon)\big\|_{2}\Big)\leq e^{1/2}\left(2\left(\sqrt{\frac{\log(2n)}{nm}}\right)^{2\log(2n)}+2\left(\frac{\log(2n)}{m}\right)^{2\log(2n)}\right)^{\frac{1}{2\log(2n)}}.
\end{equation}
Since $m>C_1n\log(2n)$ and $C_1>1$, we have 
\[
\sqrt{\frac{\log(2n)}{nm}}>\frac{\sqrt{C_1}\log(2n)}{m}>\frac{\log(2n)}{m}.
\]
Let $C'_1=e^{1/2}2^{1/\log 2}$. Then, we know from \eqref{eq:E-2} that the inequality \eqref{eq:E-0} holds.   

\section{Proof of Proposition \ref{prop:error_bound_1}}
Since $\|\overline{D}\|_\infty=b$, we know that $\|D^*-\overline{D}\|_\infty\leq 2b$. Consider the following two cases.

{\bf Case 1}: If ${\mathbb E}\left(\langle D^*-\overline{D},X\rangle^2\right)<8b^2 \displaystyle{\sqrt{\frac{256\log(2n)}{m\log(2)}}}$, then we know from \eqref{eq:E2-bounded} that
\[
\frac{\|D^*-\overline{D}\|^2}{|\Omega|}= 16b^2 \sqrt{\frac{256\log(2n)}{m\log(2)}}\le 16b^2  \sqrt{\frac{256}{\log(2)}}\sqrt{\frac{\log (2n)}{m}}.
\]

{\bf Case 2}: If ${\mathbb E}\left(\langle D^*-\overline{D},X\rangle^2\right)\geq 8b^2 \displaystyle{\sqrt{\frac{256\log(2n)}{m\log(2)}}}$, then we know from \eqref{eq:Err_origin_2} that $(D^*-\overline{D})/\sqrt{2}\|D^*-\overline{D}\|_\infty\in {\cal C}(\tau)$ with $\tau=2r(\frac{\kappa}{\kappa-1})^2\left(\alpha(\rho_2)+2\right)^2$. Thus, it follows from Lemma \ref{lemma:approx_RIP} that there exists a constant $C_2'>0$ such that with probability at least $1-1/n$,
\[
 \frac{1}{2}{\mathbb E}\left(\langle D^*-\overline{D},X\rangle^2\right)\le \frac{1}{m}\|{\cal O}(D^*-\overline{D})\|^2+2048C'_2b^2\tau|\Omega|\frac{\log (2n)}{nm}.
\]
Thus, we know from \eqref{eq:E2-bounded} and \eqref{eq:Err_origin_1} in Proposition \ref{prop:Error_bound_origin} that
\begin{eqnarray*}
\frac{\|D^*-\overline{D}\|^2}{2|\Omega|}&=& {\mathbb E}\left(\langle D^*-\overline{D},X\rangle^2\right)\le \frac{2}{m}\|{\cal O}(D^*-\overline{D})\|^2+4096C'_2b^2\tau|\Omega|\frac{\log (2n)}{nm}\\ [3pt]
&\le& 4\sqrt{2r}\left(\alpha(\rho_2)+\frac{2}{\kappa}\right)\rho_1\|D^*-\overline{D}\| +4096C'_2b^2\tau|\Omega|\frac{\log (2n)}{nm}\\ [3pt]
&\le & \frac{\|D^*-\overline{D}\|^2}{4|\Omega|}+32r|\Omega|\left(\alpha(\rho_2)+\frac{2}{\kappa}\right)^2\rho_1^2+4096C'_2b^2\tau|\Omega|\frac{\log (2n)}{nm}.
\end{eqnarray*}
By substituting $\tau$, we obtain that there exists a constant $C_3'>0$ such that
\[
\frac{\|D^*-\overline{D}\|^2}{|\Omega|}\le C_3'r|\Omega|\left(\left(\alpha(\rho_2)+\frac{2}{\kappa}\right)^2\rho_1^2+\left(\frac{\kappa}{\kappa-1}\right)^2\left(\alpha(\rho_2)+2\right)^2b^2\frac{\log (2n)}{nm}\right).
\]
The result then follows by combining these two cases.   

\section{Proof of Proposition \ref{prop:estimator_R_Omega_xi}}
From \eqref{eq:def_new_noise}, the definition of $\zeta$, we know that
\[
\left\|\frac{1}{m}{\cal O}^*(\zeta) \right\|_2\leq 2\omega\left\|\frac{1}{m}{\cal O}^*(\xi) \right\|_2+\eta\left\|\frac{1}{m}{\cal O}^*(\xi\circ \xi) \right\|_2,
\]
where $\omega:=\left\|{\cal O}(\overline{D}^{(1/2)})\right\|_{\infty}$. Therefore, for any given $t_1$, $t_2>0$, we  have
\begin{equation}\label{eq:propability_ineq_0}
{\mathbb P}\left( \left\|\frac{1}{m}{\cal O}^*(\zeta) \right\|_2\ge 2\omega t_1+\eta t_2\right)\leq {\mathbb P}\left(\left\|\frac{1}{m}{\cal O}^*(\xi) \right\|_2\ge t_1 \right)+{\mathbb P}\left(\left\|\frac{1}{m}{\cal O}^*(\xi\circ \xi) \right\|_2\ge t_2 \right).
\end{equation}
Recall that $\displaystyle{\frac{1}{m}{\cal O}^*(\xi)=\frac{1}{m}\sum_{l=1}^m\xi_lX_l}$. Denote $Z_l:=\xi_lX_l$, $l=1,\ldots,m$. Since ${\mathbb E}(\xi_l)=0$ and $\xi_l$ and $X_l$ are independent, we have ${\mathbb E}(Z_l)=0$ for all $l$. Also, we have
\[
\|Z_l\|_2\leq \|Z_l\|=|\xi_l|, \quad l=1,\ldots,m,
\]
which implies that $\left\| \|Z_l\|_2\right\|_{\psi_1}\leq \left\| \xi_l\right\|_{\psi_1}$. Since $\xi_l$ is sub-Gaussian, we know that there exists a  constant $M_1>0$ such that  $\left\| \xi_l\right\|_{\psi_1}\leq M_1$, $l=1,\ldots,m$ \citep[see, e.g.,][Section 5.2.3]{Vershynin10}. Meanwhile, for each $l$, it follows from ${\mathbb E}(\xi^2_l)=1$, \eqref{eq:E2-bounded} and $|\Omega|=n(n-1)/2$ that 
\[
\|{\mathbb E}(Z_l^2)\|_2=\|{\mathbb E}(\xi_{l}^2X_l^2)\|_2=\|{\mathbb E}(X_l^2)\|_2=\frac{1}{4|\Omega|}\big\|\sum_{1\le i<j\le n}(e_ie_j^T+e_je_i^T)^2\big\|_2= \frac{1}{4|\Omega|}(n-1)=\frac{1}{2n}.
\]
For $\displaystyle{\frac{1}{m}{\cal O}^*(\xi\circ\xi)=\frac{1}{m}\sum_{l=1}^m\xi_l^2X_l}$, denote $Y_l:=\xi_l^2X_l-{\mathbb E}(X_l)$, $l=1,\ldots,m$, where
\[
{\mathbb E}(X_l)=\frac{1}{2|\Omega|}\sum_{1\le i<j\le n}(e_ie_j^T+e_je_i^T)=\frac{1}{2|\Omega|}({\bf 1}{\bf 1}^{T}-I).
\] It is clear that for each $l$, $\|{\mathbb E}(X_l)\|=1$ and $\|{\mathbb E}(X_l)\|_{2}=1/n$. Therefore, since ${\mathbb E}(\xi^2_l)=1$, we know that ${\mathbb E}(Y_l)=0$ for all $l$. Moreover, we have
\[
\|Y_l\|_2=\|\xi_l^2X_l-{\mathbb E}(X_l)\|_2\leq \|\xi_l^2X_l-{\mathbb E}(X_l)\|\leq \xi_l^2+\|{\mathbb E}(X_l)\|< \xi_l^2+1.
\]
Thus, we have
\[
\left\|\|Y_l\|_2\right\|_{\psi_1}\leq \left\|\xi_l^2\right\|_{\psi_1}+1.
\]
From \cite[Lemma 5.14]{Vershynin10}, we know that the random variable $\xi_{l}$ is sub-Gaussian if and only if $\xi_{l}^{2}$ is sub-exponential, which implies there exists $M_2>0$ such that $\|\xi_l^2\|_{\psi_1}\leq M_2$ \citep[see e.g.,][Section 5.2.3 and 5.2.4]{Vershynin10}. Therefore, $\left\|\|Y_l\|_2\right\|_{\psi_1}\leq M_2+1$. Meanwhile, we have
\begin{eqnarray*}
\|{\mathbb E}(Y_l^2)\|_2&=&\left\|{\mathbb E}\left((\xi_l^2X_l-{\mathbb E}(X_l))(\xi_l^2X_l-{\mathbb E}(X_l))\right)\right\|_2=\left\|{\mathbb E}\left(\xi_l^4X_l^2\right)-{\mathbb E}(X_l){\mathbb E}(X_l)\right\|_2\\ [3pt]
&\leq & \left\|{\mathbb E}\left(\xi_l^4X_l^2\right)\right\|_2+\|{\mathbb E}(X_l){\mathbb E}(X_l)\|_2=\left\|{\mathbb E}\left(\xi_l^4X_l^2\right)\right\|_2+\|{\mathbb E}(X_l)\|_2^2=\frac{\gamma}{2n}+\frac{1}{n^2}.
\end{eqnarray*}
Therefore, for the sufficiently large $n$, we always have $\|{\mathbb E}(Y_l^2)\|_2\leq \gamma/n$. Denote $M_3=\max\{M_1,M_2+1\}$ and $C_3'=\max\{1/2,\gamma\}$. We know from Lemma \ref{lem:Bernstein-ineq-sym} that for any given $t_1$, $t_2>0$
\begin{equation}\label{eq:propability-1}
{\mathbb P}\left(\left\|\frac{1}{m}{\cal O}^*(\xi) \right\|_2\ge t_1 \right)\leq 2n\max\left\{ {\rm exp}\left(-\frac{nmt_1^2}{4C_3'}\right), {\rm exp}\left(-\frac{mt_1}{2M_3}\right) \right\}
\end{equation}
and
\begin{equation}\label{eq:propability-2}
{\mathbb P}\left(\left\|\frac{1}{m}{\cal O}^*(\xi\circ\xi) \right\|_2\ge t_2 \right)\leq 2n\max\left\{ {\rm exp}\left(-\frac{nmt_2^2}{4C_3'}\right), {\rm exp}\left(-\frac{mt_2}{2M_3}\right) \right\}.
\end{equation}
By choosing $t_1=\displaystyle{2\sqrt{2}\sqrt{\frac{C_3'\log(2n)}{nm}}}$ and $t_2={\omega t_1}/{\eta}$, we know from $m>C_1n\log(2n)$ (for the sufficiently large $C_1$) that the first terms of the right hand sides of \eqref{eq:propability-1} and \eqref{eq:propability-2} both dominate the second terms, respectively. Thus, since $\eta<\omega$, we have
\[
{\mathbb P}\left(\left\|\frac{1}{m}{\cal O}^*(\xi) \right\|_2\ge t_1 \right)\leq \frac{1}{2n} \quad {\rm and} \quad {\mathbb P}\left(\left\|\frac{1}{m}{\cal O}^*(\xi\circ\xi) \right\|_2\ge t_2 \right)\leq \frac{1}{2n}.
\]
Finally, it follows from \eqref{eq:propability_ineq_0} that
\[
{\mathbb P}\left( \left\|\frac{1}{m}{\cal O}^*(\zeta) \right\|_2\ge 12\omega\sqrt{\frac{C_3'\log(2n)}{nm}}\right)\leq \frac{1}{n}.
\]
The proof is completed. 

\section{Proof of Proposition \ref{prop:better-choice-rho2}}
By Ky Fan's inequality \citep{Fan49}, we know that $\langle\overline{P}_1\overline{P}_1^T,\widetilde{P}_1\widetilde{P}_1^T \rangle\leq r$. From \eqref{eq:notation_alpha_beta2}, we have
\[
\alpha^2(2)=\frac{1}{2r}(5r-4\langle\overline{P}_1\overline{P}_1^T,\widetilde{P}_1\widetilde{P}_1^T \rangle)\geq \frac{1}{2r}(5r-4r)=\frac{1}{2}=\alpha^2(0).
\]
Therefore, we only need to show that
\[
\alpha(1)=\frac{1}{\sqrt{2r}}\|\widetilde{P}_1\widetilde{P}_1^T-\overline{P}_1\overline{P}_1^T\|<\frac{1}{\sqrt{2}}=\alpha(0).
\]
The rest of the proof is similar to that of \cite[Theorem 3]{MPSun12}. Let ${\cal N}_{\delta}:=\{D\in{\cal S}^n\mid \|D-\overline{D}\|\leq \delta \}$, where $\delta=\|\widetilde{D}-\overline{D}\|$. For any $D\in{\cal N}_{\delta}$, we have
\[
|\lambda_i(-JDJ)-\lambda_i(-J\overline{D}J)|=|\lambda_i(-JDJ)-\overline{\lambda}_i|\leq \|-JDJ+J\overline{D}J\|\leq \|D-\overline{D}\|\leq \delta,\quad i=1,\ldots,n.
\]
Moreover, it follows from $\delta<\overline{\lambda}_r/2$ that for any $D\in{\cal N}_{\delta}$,
\[
 \lambda_r(-JDJ)\ge \overline{\lambda}_r-\delta> \overline{\lambda}_r/2>\delta\ge \lambda_{r+1}(-JDJ).
\]
Therefore, for any $D\in {\cal N}_{\delta}$, we have $\Phi(-JDJ)=P_1P_1^T$, where $P=[P_1\ \ P_2]\in{\mathbb O}^n$ satisfies $-JDJ=P{\rm Diag}(\lambda(-JDJ))P^T$ with $P_1\in\Re^{n\times r}$ and $P_2\in\Re^{n\times (n-r)}$. Moreover, $\Phi$ defined by \eqref{eq:def-Phi-LOP} is continuously differentiable over ${\cal N}_{\delta}$. Thus, we know from the mean value theorem that
\begin{equation}\label{eq:mean-v-t}
\widetilde{P}_1\widetilde{P}_1^T-\overline{P}_1\overline{P}_1^T=\Phi(-J\widetilde{D}J)-\Phi(-J\overline{D}J)=\int_{0}^{1}\Phi'(-JD_tJ)(-J\widetilde{D}J+J\overline{D}J){\rm d}t,
\end{equation}
where $D_t:=\overline{D}+t(\widetilde{D}-\overline{D})$.

For any $D\in{\cal N}_{\delta}$, we know from the derivative formula of the L\"{o}wner operator  that for any $H\in{\cal S}^n$,
\[
\Phi'(-JDJ)H=P[\Omega \circ (P^THP)]P^T,
\]
where $\Omega\in{\cal S}^n$ is given by
\[
(\Omega)_{ij}:=\left\{ \begin{array}{ll}
\displaystyle{\frac{1}{\lambda_i(-JDJ)-\lambda_j(-JDJ)}} & \mbox{if $1\le i \le r$ and $r+1\le j \le n$,}\\ [3pt]
\displaystyle{\frac{-1}{\lambda_i(-JDJ)-\lambda_j(-JDJ)}} & \mbox{if $r+1\le i \le n$ and $1\le j \le r$,}\\ [3pt]
0 & \mbox{otherwise},
\end{array} \right. \quad i,j\in\{1,\ldots,n\},
\]
which implies that
\[
\|\Phi'(-JDJ)H\|\leq \frac{\|H\|}{\lambda_r(-JDJ)-\lambda_{r+1}(-JDJ)}.
\]
This, together with \eqref{eq:mean-v-t} yields
\begin{eqnarray*}
\|\widetilde{P}_1\widetilde{P}_1^T-\overline{P}_1\overline{P}_1^T\|&\leq&\int_{0}^{1}\|\Phi'(-JD_tJ)(-J\widetilde{D}J+J\overline{D}J)\|{\rm d}t\\ [3pt]
&\leq&\int_{0}^{1}\frac{\|J(\widetilde{D}-\overline{D})J\|}{\lambda_r(-JD_tJ)-\lambda_{r+1}(-JD_tJ)}{\rm d}t\\ [3pt]
&\leq&\int_{0}^{1}\frac{\|\widetilde{D}-\overline{D}\|}{\lambda_r(-JD_tJ)-\lambda_{r+1}(-JD_tJ)}{\rm d}t.
\end{eqnarray*}
By Ky Fan's inequality, we know that
\begin{eqnarray*}
&&(\lambda_r(-JD_tJ)-\overline{\lambda}_r)^2+\lambda_{r+1}^2(-JD_tJ)\\ [3pt]
&\leq& \|\lambda(-JD_tJ)-\lambda(-J\overline{D}J)\|^2\leq \|-JD_tJ+J\overline{D}J\|^2\leq \|D_t-\overline{D}\|^2=t^2\delta^2.
\end{eqnarray*}
It can be checked directly that $\lambda_r(-JD_tJ)-\overline{\lambda}_r-\lambda_{r+1}(-JD_tJ)\geq -\sqrt{2}t\delta$, which implies that
\[
\lambda_r(-JD_tJ)-\lambda_{r+1}(-JD_tJ)\geq \overline{\lambda}_r+\lambda_r(-JD_tJ)-\overline{\lambda}_r-\lambda_{r+1}(-JD_tJ)\geq \overline{\lambda}_r-\sqrt{2}t\delta.
\]
Thus,
\[
\|\widetilde{P}_1\widetilde{P}_1^T-\overline{P}_1\overline{P}_1^T\| \leq \int_{0}^{1}\frac{\delta}{\overline{\lambda}_r-\sqrt{2}t\delta}{\rm d}t=-\frac{1}{\sqrt{2}}\log\left(1-\frac{\sqrt{2}\delta}{\overline{\lambda}_r}\right).
\]
Since $r\ge 1$, we know that
\[
\delta/\overline{\lambda}_r<1/2<0.5351<\frac{1}{\sqrt{2}}\left(1-{\rm exp}\left(-{\sqrt{2r}}\right)\right),
\]
which implies that
\[
\frac{1}{\sqrt{r}}\|\widetilde{P}_1\widetilde{P}_1^T-\overline{P}_1\overline{P}_1^T \|<1.
\]
Therefore, the proof is completed.

\vskip 0.2in
\bibliography{EDMEbib}

\end{document}